\documentclass{article}

\usepackage{microtype}
\usepackage{subfig}
\usepackage{graphicx}

\usepackage{booktabs} 

\usepackage{hyperref}

\usepackage{amsmath,amsfonts,bm}

\def\Figref#1{Figure~\ref{#1}}

\def\Secref#1{Section~\ref{#1}}

\def\eqref#1{equation~\ref{#1}}
\def\Eqref#1{Equation~\ref{#1}}

\def\1{\bm{1}}

\def\va{{\bm{a}}}
\def\vb{{\bm{b}}}

\def\vu{{\bm{u}}}
\def\vv{{\bm{v}}}

\def\vy{{\bm{y}}}

\DeclareMathAlphabet{\mathsfit}{\encodingdefault}{\sfdefault}{m}{sl}
\SetMathAlphabet{\mathsfit}{bold}{\encodingdefault}{\sfdefault}{bx}{n}

\def\gA{{\mathcal{A}}}

\def\gD{{\mathcal{D}}}

\def\gG{{\mathcal{G}}}
\def\gH{{\mathcal{H}}}

\def\gL{{\mathcal{L}}}

\def\gO{{\mathcal{O}}}
\def\gP{{\mathcal{P}}}

\def\gS{{\mathcal{S}}}

\def\gU{{\mathcal{U}}}

\def\gW{{\mathcal{W}}}

\def\sN{{\mathbb{N}}}

\newcommand{\E}{\mathbb{E}}

\newcommand{\reg}{\lambda}

\newcommand{\normltwo}{L^2}

\newcommand{\D}{\textsc{D}}
\newcommand{\ipm}{\mathrm{IPM}}
\newcommand{\mmd}{\mathrm{MMD}}
\newcommand{\sd}{\mathrm{SD}}
\newcommand{\sg}{\mathtt{sg}}

\usepackage[accepted]{icml2024}

\usepackage{amsmath}
\usepackage{amssymb}
\usepackage{mathtools}
\usepackage{amsthm}

\usepackage[capitalize,noabbrev]{cleveref}

\usepackage{array}
\usepackage{multirow}

\theoremstyle{plain}

\theoremstyle{definition}

\theoremstyle{remark}

\usepackage[textsize=tiny]{todonotes}

\icmltitlerunning{Dynamics Stable Learning by Invariant Measure for Chaotic Systems}

\begin{document}

\twocolumn[
\icmltitle{DySLIM: Dynamics Stable Learning by Invariant Measure for Chaotic Systems}



\icmlsetsymbol{equal}{*}

\begin{icmlauthorlist}
\icmlauthor{Yair Schiff}{cornell}
\icmlauthor{Zhong Yi Wan}{google}
\icmlauthor{Jeffrey B. Parker}{google}
\icmlauthor{Stephan Hoyer}{google}
\icmlauthor{Volodymyr Kuleshov}{cornell}
\icmlauthor{Fei Sha}{google}
\icmlauthor{Leonardo Zepeda-N\'u\~nez}{google,uwmadison}
\end{icmlauthorlist}

\icmlaffiliation{cornell}{Department of Computer Sciences, Cornell Tech, New York, NY, USA}
\icmlaffiliation{google}{Google Research, Mountain View, CA, USA}
\icmlaffiliation{uwmadison}{Department of Mathematics, University of Wisconsin-Madison, WI, USA}

\icmlcorrespondingauthor{Yair Schiff}{yairschiff@cs.cornell.edu}
\icmlcorrespondingauthor{Leonardo Zepeda-N\'u\~nez}{lzepedanunez@google.com}

\icmlkeywords{Machine Learning, ICML}

\vskip 0.3in
]



\printAffiliationsAndNotice{}

\begin{abstract}
Learning dynamics from dissipative chaotic systems is notoriously difficult due to their inherent instability, as formalized by their positive Lyapunov exponents, which exponentially amplify errors in the learned dynamics. However, many of these systems exhibit ergodicity and an attractor: a compact and highly complex manifold, to which trajectories converge in finite-time, that supports an invariant measure, i.e., a probability distribution that is invariant under the action of the dynamics, which dictates the long-term statistical behavior of the system. In this work, we leverage this structure to propose a new framework that targets learning the invariant measure as well as the dynamics, in contrast with typical methods that only target the misfit between trajectories, which often leads to divergence as the trajectories' length increases. We use our framework to propose a tractable and sample efficient objective that can be used with any existing learning objectives. Our \textbf{Dy}namics \textbf{S}table \textbf{L}earning by \textbf{I}nvariant \textbf{M}easure (DySLIM) objective enables model training that achieves better point-wise tracking and long-term statistical accuracy relative to other learning objectives. By targeting the distribution with a scalable regularization term, we hope that this approach can be extended to more complex systems exhibiting slowly-variant distributions, such as weather and climate models. Code to reproduce our experiments is available \href{https://github.com/google-research/swirl-dynamics/tree/main/swirl_dynamics/projects/ergodic}{here}.
\end{abstract}

\section{Introduction}\label{sec:intro}
Building data-driven surrogate models to emulate the dynamics of complex time-dependent systems is a cornerstone task in scientific machine learning \cite{Farmer1987}, with applications ranging from fluid dynamics \cite{sanchez2020learning}, weather forecasting \cite{lam2022graphcast, pathak2022fourcastnet,bi2023accurate}, climate modeling \cite{kochkov2023neural}, molecular dynamics \cite{jia2020pushing, merchant2023scaling}, quantum chemistry \cite{chen2020deepks,zepeda2021deep}, and plasma physics \cite{mathews2021uncovering, anirudh20222022}.

Historically, various methods based on PCA \citep{PCA_Pearson_1901} and Koopman theory \cite{koopman1931hamiltonian} have been proposed to learn emulators by leveraging large datasets to build a surrogate model during a, typically expensive, offline phase \cite{schmid_dmd_2010, alexander2020operator, kaiser2021data,schmid2022dynamic}.
The learned emulator provides fast and inexpensive inference, which is then used to accelerate downstream tasks such as design, control, optimization, data assimilation, and uncertainty quantification. Alas, many of these techniques are inherently linear, which renders them inadequate for problems with highly non-linear dynamics.


Indeed, many of the underlying physical processes driving target applications are described by non-linear and chaotic systems, which are characterized by strong instabilities, particularly with respect to initial conditions: trajectories with close initial conditions diverge quickly due to the positive Lyapunov exponents \cite{medio2001nonlinear,strogatz2018nonlinear}.
Fortunately, many of these systems are dissipative, which implies the existence of a compact set, often called an \textbf{attractor}, towards which all bounded sets of initial conditions converge in finite-time \citep{temam2012infinite}.
In addition, many of these systems also empirically exhibit \textbf{ergodicity}, whose main consequence translates to the existence of an attractor-induced \textbf{invariant measure}, 
a measure that is unchanged by the dynamics of the system,
which captures the \textit{long-term behavior} of the system \citep{stuart1998dynamical}.

\begin{figure*}[h]
    \centering

    \begin{minipage}[c]{0.29\textwidth} 
    \subfloat[]{\includegraphics[width=0.95\textwidth, trim ={0mm 00mm 0mm 0mm}, clip]{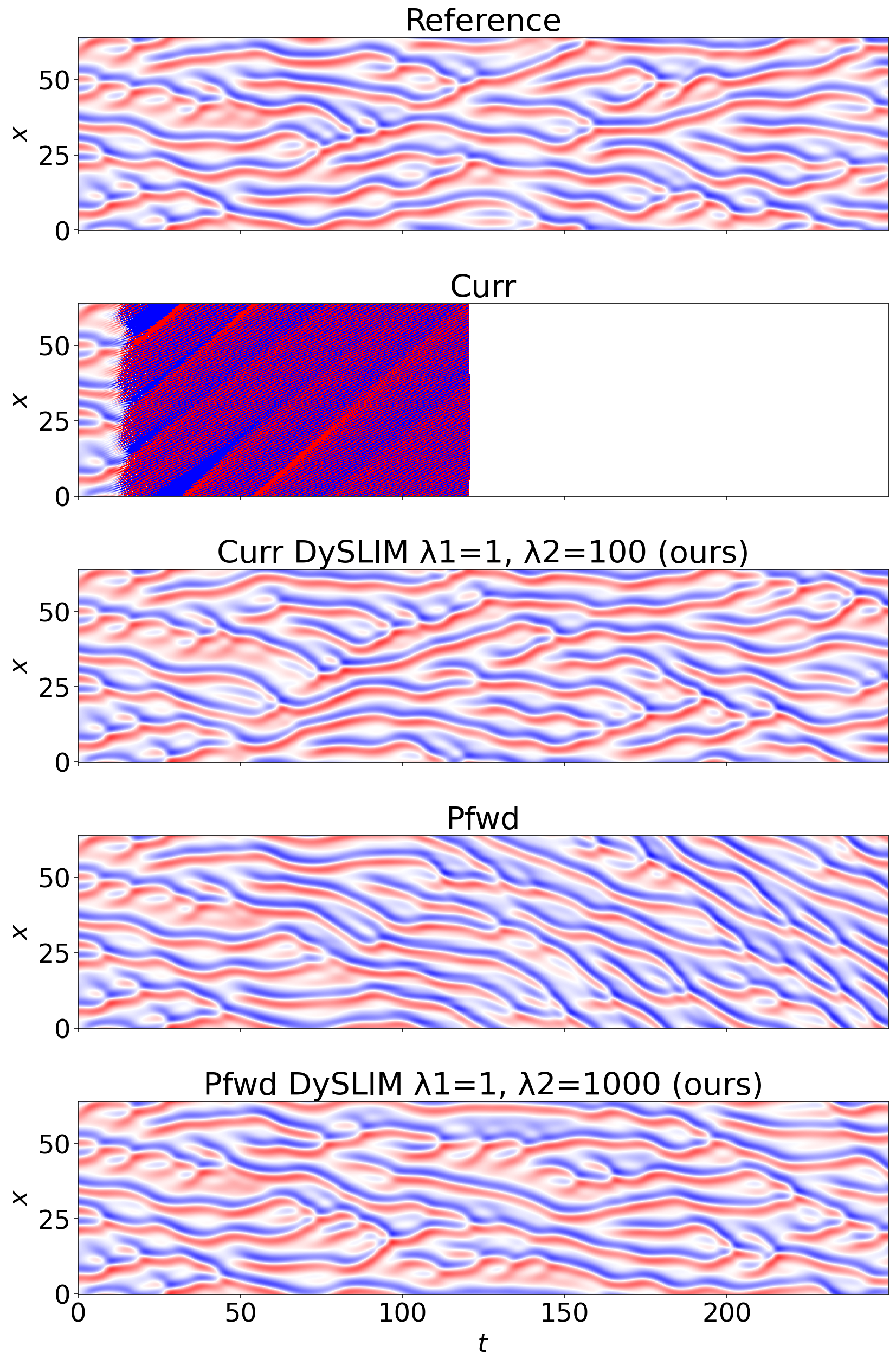}}
    \label{fig:ks_traj}
    \end{minipage}
    \hspace{0.3cm}
    \begin{minipage}[c]{0.67\textwidth}  
        \vfill
        \subfloat[]
        {\includegraphics[width=0.95\textwidth, 
        ]{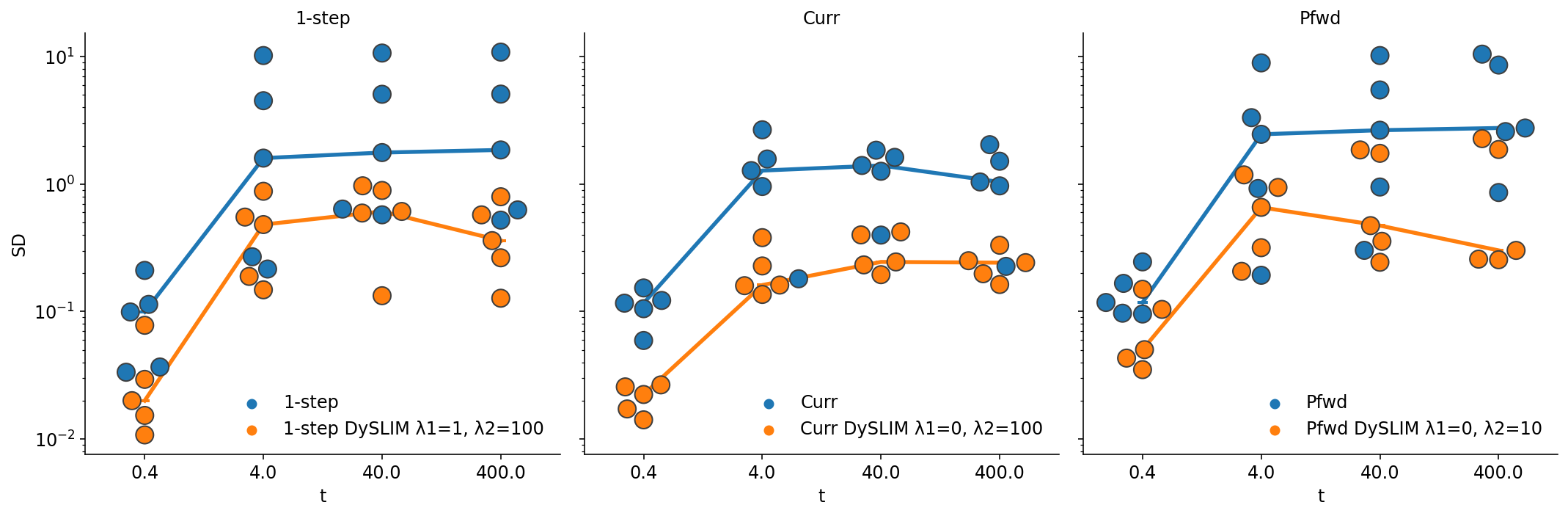}}
         \vfill
        \vspace{-0.3cm}
        \subfloat[]{
        \includegraphics[
        width=0.95\textwidth
        ]{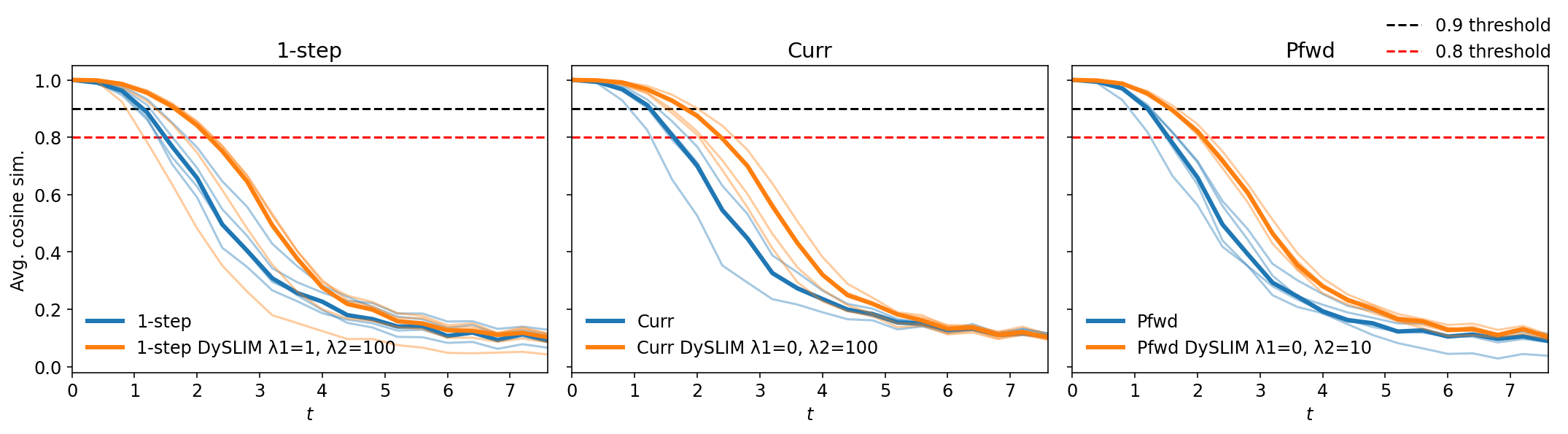}
        }

    \end{minipage}
    \caption{Improved stability with regularized DySLIM objectives in the Kuramoto–Sivashinsky (KS) and Lorenz 63 systems.
    (a) Sample ground truth and predicted trajectory across models trained on the KS system using Curriculum training (Curr) and the Pushforward trick (Pfwd) with/without regularization. The base versions showcase the blow-up issue (Curr) and wrong long-time dynamics (Pfwd).
    (b) Sinkhorn Divergence (SD; $\downarrow$) between trajectories at various rollout times of the Lorenz 63 system.
    Each point represents a random training seed, with the solid line indicating median values.
    (c) Cosine similarity ($\uparrow$) over time for the Lorenz 63 system.
    Each line corresponds to one of five random training seeds with bolded lines indicating median values.
    }
    \label{fig:ks_lz63_traj}
\end{figure*}

Recent advances in machine learning (ML) have driven the development of several techniques for learning data-driven surrogates for non-linear dynamics \cite{rajendra2020modeling, roy2021machine, brunton2022data,ghadami2022data,nghiem2023physics}.
In the context of data-driven learning, autoregressive models (the focus of this paper) are a prevalent approach due to their ability to infer trajectories of arbitrary length.
These autoregressive models predict a system's state at time $t+\Delta t$ based on its state at time $t$. Iterative application (unrolling) allows for trajectory forecasting far beyond training time horizons.

However, learning chaotic dynamics using autoregressive models in a stable manner has proven elusive.
Due to memory and computational constraints, traditional ML-based approaches focus on learning short-term dynamics by minimizing the mean square error (MSE) between reference trajectories and those generated by unrolling a learned model; commonly using recurrent neural networks \cite{vlachas2018data, fan2020long} or learning a projection from a stochastic trajectory using reservoir computing \cite{pathak2017using, bollt2021explaining, hara2022learning}.
Unfortunately, these models usually overfit to the short-term dynamics to the detriment of accurately predicting long-term behavior \citep{bonev2023spherical}.
This manifests as trajectory blow-up, the values of the state variables diverging to infinity, or inaccurate long-term statistics during inference with large time-horizon, as depicted in \Figref{fig:ks_lz63_traj} (a).
Recent works have focused on minimizing the misfit between increasingly longer trajectories \cite{keisler2022forecasting}.
Although these methods have been shown to attenuate the instability, the underlying difficulty remains: due to chaotic divergence, the losses become increasingly uninformative, which causes their gradients\footnote{Gradients are computed by backpropagation through the unrolled steps and are prone to exacerbate instabilities in the system.
This is related to the well-known issue of exploding/vanishing gradients \cite{pascanu2013difficulty}.} to diverge, as shown by \citet{mikhaeil2022difficulty}.

A prime example of this challenge is weather and climate. While state-of-the-art ML models can learn short-term weather patterns \cite{lam2022graphcast}, learning long-term climate behavior remains a very challenging open problem \cite{kochkov2023neural,watt2023ace}.
Thus the question we seek to answer becomes: 
\textit{How do we incorporate knowledge of a system's long-term behavior into the learning stage, so that models remain point-wise accurate for short-term predictions and statistically accurate for long-term ones.}

This work addresses this challenge by leveraging the presence of attractors and their invariant measures.
We propose a framework that directly targets the learning of long-term statistics by a measure-matching regularization loss.

\textbf{Contributions} \hspace{0.1cm}
The contributions of this paper are three-fold: first, we propose a probabilistic and scalable framework for learning chaotic dynamics using data-driven, ML-based methods.
Our framework introduces a system-agnostic\footnote{We assume no knowledge of the systems, such as the expression of the equations driving the dynamics.} measure-matching regularization term into the loss that induces stable and accurate trajectories satisfying the long-term statistics while enhancing short-term predictive power.
Second, we use our framework to propose a tractable, sample and computationally-efficient\footnote{Our regularization loss incurs an extra cost depending only quadratically on the batch size.}
objective that we dub \textbf{Dy}namics \textbf{S}table \textbf{L}earning by \textbf{I}nvariant \textbf{M}easure (DySLIM) that can be used in conjunction with any existing dynamical system learning objective.
Third, we demonstrate that DySLIM is capable of tackling larger and more complex systems than competing probabilistic methods, up to a state-dimension of 4,096 with complex 2D dynamics.
Namely, we show competitive results in three increasingly complex and higher dimensional problems: the Lorenz 63 system \citep{lorenz1963deterministic, tucker1999lorenz}, a prototypical chaotic systems with the well known ``butterfly'' attractor, the Kuramoto-Sivashinsky (KS) equation, a 1D chaotic PDE, and the Kolmogorov-Flow \citep{obukhov1983kolmogorov}, a 2D chaotic system that is routinely used as a benchmark for turbulent fluid dynamics \cite{kochkov_machine_2021}.
For the largest systems, we show that DySLIM performance remains stable even for large batch sizes and learning rates, regimes in which the performance of other methods deteriorates rapidly.
These capabilities are potentially useful for accelerating the training stage by leveraging data parallelism and large learning rates.


\section{Background}\label{sec:background}
We consider 
autonomous systems of the form
\begin{align} \label{eq:dyn_system}
    \partial_t\vu = \mathcal{F}[\vu(t)],
\end{align}
where $\vu(t)$ is the state of the system at time $t$.

For a given fixed time-step $\Delta t$, we discretize \Eqref{eq:dyn_system} in space and time, and we define $\vu_k = \vu(k \Delta t ) \in \mathbb{R}^D $ together with the discrete dynamical system  
\begin{align} \label{eq:discrete_propagator}
  \vu_{k+1} = \gS_{\Delta t}(\vu_{k}),  
\end{align}
where $\gS_{\Delta t}$ is the map obtained by integrating \Eqref{eq:dyn_system} in time by a period $\Delta t$.
In what follows, for brevity, we drop the subscript $\Delta t.$
We can use the operator $\gS$ to unroll, or advance in time, the solution of the dynamical system, 
\begin{align}\label{eq:unroll}
    \vu_{k} = \gS(\vu_{k-1}) = \gS \circ \gS(\vu_{k-2}) =  ... =  \gS^k (\vu_0). 
\end{align} 

\textbf{Chaotic Systems} \hspace{0.1cm}
A chaotic system can be loosely defined as one whose trajectories are highly \textit{sensitive to initial conditions.}
Let $(\gU, \mathrm{d})$ be an Euclidean metric space.
We say that the system given by $\gS$ is chaotic if there exists $\varepsilon > 0$ such that for all $\vu_0 \in \gU$ and $\delta > 0,$ there exists $\vv_0 \in B_\delta(\vu_0)$ and $k \in \sN,$ such that:
\begin{align} \label{eq:chaotic_divergence}
    \mathrm{d}(\gS^{k}(\vu_0), \gS^{k}(\vv_0)) \geq \varepsilon,
\end{align}
where $B_\delta(\vu_0) = \{\vy \mid \mathrm{d}(\vu_0, \vy) < \delta\}$ is a ball of radius $\delta$ centered at $\vu_0.$
Chaotic systems are also characterized by having a positive Lyapunov exponent: small discrepancies in the initial conditions are exaggerated exponentially over time \citep{strogatz2018nonlinear}.

\textbf{Invariant Measures and Attractors} \hspace{0.1cm}
We assume that the state space is measurable $(\gU, \gA),$ where $\gA$ is the Borel $\sigma$-algebra on $(\gU, \mathrm{d}),$ and we have a probability measure $\mu: \gA \rightarrow [0, 1].$
If the discrete-time dynamical system map $\gS$ is measurable,
then it also defines a probability distribution $\gS_{\#}\mu: \gA \rightarrow [0, 1]$ which is called the pushforward of $\mu$ by $\gS$, with $\gS_{\#}\mu(A) = \mu(\gS^{-1}(A)),$ for all $A \in \gA.$
We say $\gS$ preserves a measure $\mu$, also denoted as $\mu$ is invariant\footnote{We note that invariant measures may not be unique. For example, transformations with high degree of symmetries, such as a rigid-body transformation (e.g., translation and rotation), can have an infinite number of invariant measures.} under/to $\gS,$ if:
\begin{align*}
    \mu(\gS^{-1}(A)) = \mu(A), \forall A \in \gA, ~~ \text{or equivalently}
    ~~ \gS_{\#}\mu = \mu.
\end{align*}
Intuitively, an attractor is a subset of the state space that characterizes the `long-run' or `typical' condition of the system.
Formally, $A^* \subseteq \gU $ is an attractor if it is a minimal set that satisfies the following properties: \textit{i}) for all $\va \in A^*$ and $k \geq 0, \, \, \gS^k(\va) \in A^*$ (i.e., $A^*$ is invariant under $\gS$) and \textit{ii}) there exists $B \subseteq \gU,$ known as the basin of attraction, such that for all $\vb \in B$ and $\varepsilon > 0$ there exists some $k^{\star} > 0$ such that $\gS^k(\vb)$ is in an $\varepsilon$-neighborhood of $A^*$, for all $k \geq k^{\star}.$
If the basin of attraction consists of the entire state space, then $A^*$ is said to be a global attractor \cite{stuart1994numerical}. As an example, \Figref{fig:app_l63} depicts the Lorenz 63 attractor.

\section{Learning Dynamical Systems}\label{subsec:learn_ds}

Our goal is to find a Markovian parametric model $\gS_\theta$ that governs our system in a manner consistent with the true dynamics defined by $\gS$.
To do so, we leverage previously collected data, which consists of $n$ trajectories: $\gD = \{(\vu^{(i)}_j)_{j=0}^{\ell^{(i)}}\}_{i=1}^n,$ where $\ell^{(i)}$ is the length of the $i$-th trajectory, whose initial conditions are sampled from an invariant measure supported on the attractor, i.e., $\{\vu^{(i)}_0\}_{i=1}^n \stackrel{\mathrm{iid}}{\sim} \mu_0 = \mu^*$.
Letting $\mu_j$ be the distribution over states $\vu_j$, i.e., states after $j$ time-steps, for ergodic dissipative systems, we have that $\mu_j := \gS^j_{\sharp} \mu_0 = \gS^j_{\sharp} \mu^{*} = \mu^{*}$.

$\gS_\theta$ is trained by minimizing an empirical estimate of the mismatch between predicted and observed trajectories.
Most of these estimates are based on MSE, e.g., the {\it one-step objective}:
\begin{align}\label{eq:obj_one_step}
    \gL^{\text{1-step}}(\theta) = \E_j\E_{\vu_j\sim\mu_j}\left[ \left \|\gS_\theta(\vu_j) - \gS(\vu_{j}) \right \|^2\right],
\end{align}
for a norm $||\cdot||$ induced by $\mathrm{d}$ in \Eqref{eq:chaotic_divergence}, and where the outer expectation $\E_j$ represents averages along trajectories.

At inference, learned models generate trajectories by autoregressively unrolling predictions, as in \Eqref{eq:unroll}: starting from a given $\vu_0,$ we generate $\tilde{\vu}_{k} =  \gS_\theta^k(\vu_0).$ 
As we unroll for large $k$, the learning dynamics can become unstable, by either diverging or converging to a different attractor.

\paragraph{Multi-step Objectives} To attenuate this issue,  two popular objectives have been introduced recently, which have been used to train state-of-the-art models \citep{brandstetter2022message,lam2022graphcast,kochkov2023neural}.
Specifically, we examine a generalization of $\gL^{\text{1-step}}$, the $\ell$-step objective: 
\begin{align}\label{eq:obj_ell_step}
    \gL^{\ell\text{-step}}(\theta) = \E_j\E_{\vu_j\sim\mu_j} \sum_{k=1}^{\ell} \omega(k) \left \|\gS^k_\theta(\vu_j) - \gS^k(\vu_{j}) \right \|^2,
\end{align}
where 
$\omega(k)$ is a \textit{discount factor} used to stabilize training\footnote{Since matching further rolled-out steps increases in difficulty with $k,$ especially for chaotic systems, we consider a monotonically decreasing 
discount factor of the form $\omega(k) = r^{k-1},$ $0 < r < 1$, inspired by \citet{kochkov2023neural}.}.
Training paradigms where $\ell$ starts at one and is gradually increased are known as \textit{curriculum training} (Curr; \citet{krishnapriyan2021characterizing, keisler2022forecasting}), and we denote them as $\gL^{\text{Curr}}$.

Alas, $\gL^{\ell\text{-step}}$ introduces several difficulties.
By the chain rule, computing the gradient of \Eqref{eq:obj_ell_step} requires the storage of $k$ intermediate evaluations 
for each term in the inner sum in order to calculate the Jacobian $\nabla_\theta \gS_\theta^k(\vu_0),$ which can be prohibitive unless gradient checkpointing is used \cite{chen2016training}.
Crucially, for chaotic systems, \citet{mikhaeil2022difficulty} proved that these gradients necessarily `explode' as the length of the trajectory grows.

To reduce computational cost and further induce stability, one can use the \textit{pushforward trick} (Pfwd), introduced in \citet{brandstetter2022message}.
The pushforward trick replaces inputs $\vu_j$ to the parametric model with noised states $\tilde{\vu}_{j}$ drawn from an adversarial distribution induced by the model, e.g., $\tilde{\vu}_j = \sg(\gS_\theta(\vu_{j-1})),$ where $\sg(\cdot)$ represents the stop-gradient operation.
The noise can be also generated by the repetitive application of the to-be-learned model\footnote{The pushforward trick can be re-framed in our measure-matching framework although using a discrete Wasserstein 2 metric. See Appendix \ref{sec:app_pfwd_trick} for more details.} $\gS_\theta$, e.g. $\tilde{\vu}_{j+ k}= \sg(\gS_{\theta}^{k}(\vu_{j})).$
In such cases, the pushforward objective can be written in general form as:
\begin{align}\label{eq:obj_pfwd}
    &\gL^{\text{Pfwd}, \ell
    }(\theta) = \\ \nonumber
    &\E_j\E_{\vu_{j} \sim \mu_{j}}[\omega(\ell)||\gS_\theta(\sg(\gS^{\ell-1}_{\theta}(\vu_{j}))) - \gS^{\ell}(\vu_j)||^{2}].
\end{align}
This objective can either be used to replace or in addition to those defined in Equations \ref{eq:obj_one_step} and \ref{eq:obj_ell_step}.

\begin{figure*}
    \centering
    \begin{minipage}[c]{0.99\textwidth}
        \centering
        \vfill
        \subfloat[]{\includegraphics[height=3.3cm, trim={0mm 0mm 0mm 4mm}, clip]{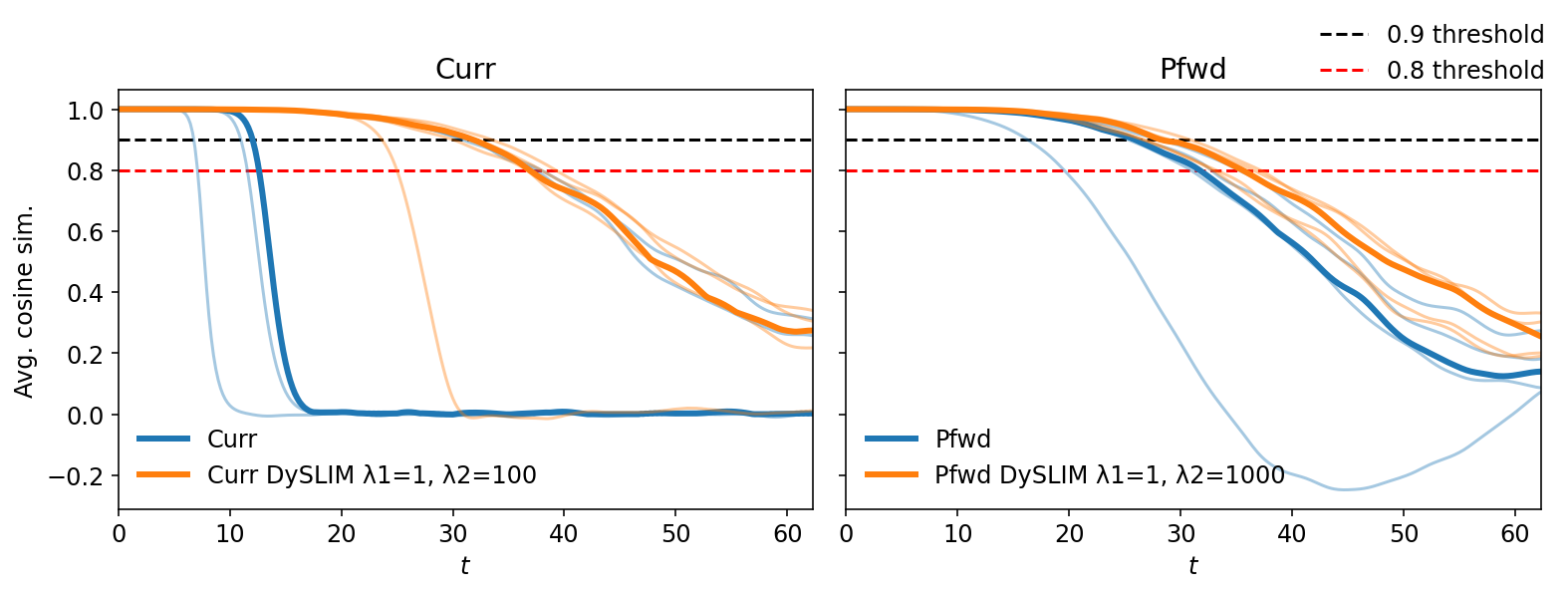}}
        \subfloat[]{\includegraphics[height=3.3cm, trim={0mm 0mm 0mm 0mm}, clip]{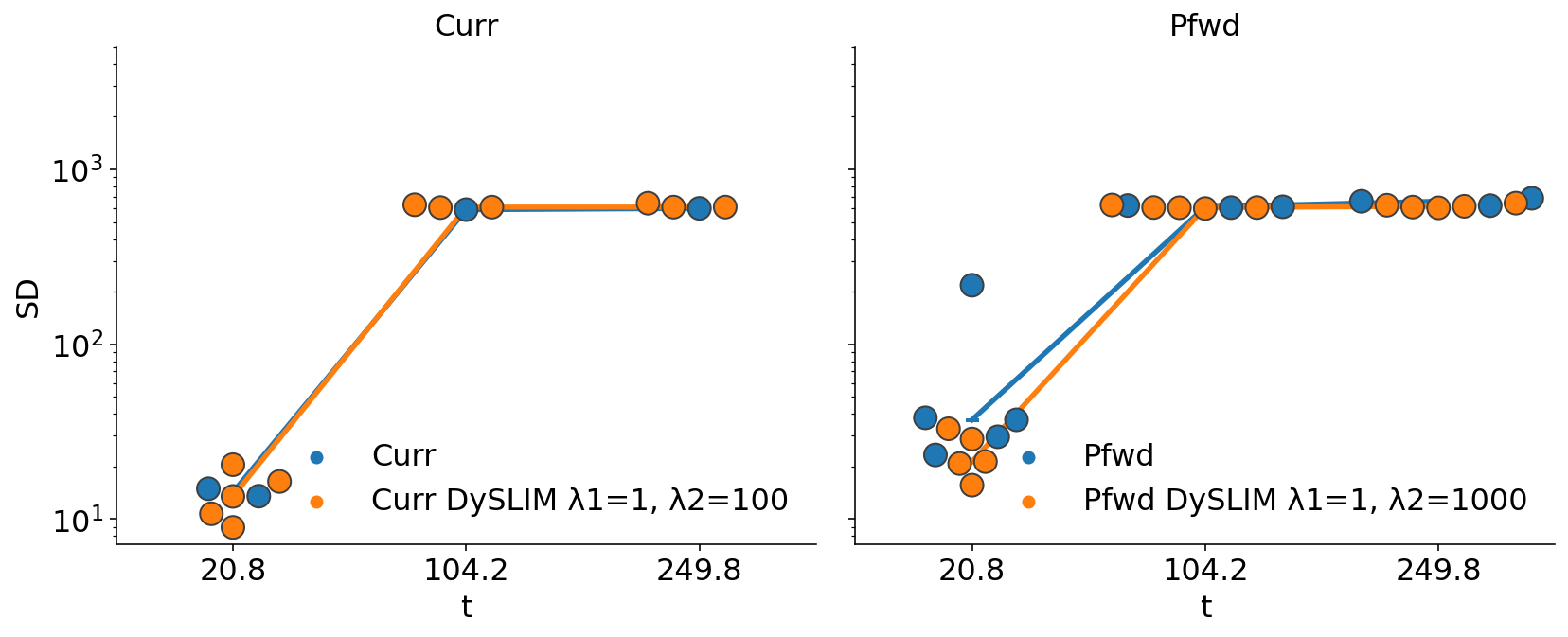}}
    \end{minipage}
    \caption{Regularized DySLIM objectives outperform baselines for the KS system.
    (a) Cosine similarity ($\uparrow$) over time.
    Each line corresponds to the mean over trajectories of each of five random training seeds, with bold lines indicating median values.
    (b) Sinkhorn Divergence (SD; $\downarrow$) between trajectories at various rollout times.
    Each point represents a random training seed that remains stable, with the solid line indicating median values.
    }
    \label{fig:ks_lz63_traj_stats}
\end{figure*}

\textbf{Sources of Instability} \hspace{0.1cm}
We recast the instability of learned dynamical models as short-term \textit{overfitting} and long-term \textit{distribution shift}:   
parameters $\theta$ that minimize $\gL^{\text{1-step}}$ on training data often overfit to this data and lead to $\gS_{\theta\#}\mu_j \neq \mu_{j+1}.$
When deployed, the learned dynamical model will accumulate errors along a predicted trajectory as the distribution of predicted states veers further away from that of the actual system.
Recent techniques (including the Curr and Pfwd training) attempt to mitigate this issue by encouraging the model to recover from deviations caused by pushing forward by $\gS_\theta$.
However, these objectives are still prone to instabilities.
For example, \Figref{fig:ks_lz63_traj} (a) depicts issues for the chaotic KS equation.
The model trained with Curr training fails to generalize beyond the first $\ell$ steps on which it was trained: the trajectory quickly enters an unstable attractor, from which it blows up.
Similarly, the model trained using the Pfwd training
is able to learn the short-term dynamics, however, as time increases, the trajectories enter a different attractor, one in which the dynamics are biased towards the right.
In both cases, by introducing our proposed regularization, we are able to correct the long-term behavior.

\section{Main Idea and Methods}\label{sec:methods}
To tackle the issue of distribution shift, we propose to focus on systems' invariant measure preservation.
Specifically, many systems of interest have some measure $\mu^*$ supported on an attractor that is invariant to the transformation $\gS$ \cite{tucker2002rigorous,weinan2002gibbsian, luzzatto2005lorenz, ferrario2008invariant, hawkins2021attractors}.
We cast our learning problem as finding parameters $\theta$ such that a surrogate $\gS_\theta$ preserves $\mu^*$ while approximating $\gS$ locally, which defines the following constrained optimization:
\begin{align}\label{eq:obj_constrained}
        \min_\theta \gL(\theta) ~~~~ \text{s.t.} ~~\mu_{\theta}^* = \mu^*,
\end{align}
where $\gL(\theta)$ is the short-term loss, and $\mu_{\theta}^*$ is the invariant measure of $\gS_\theta$, i.e., $(\gS_\theta)_{\sharp} \mu_{\theta}^*=\mu_{\theta}^*$, which we assume exists\footnote{This is a key hypothesis in our methodology.
Otherwise, this assumption can be enforced in $\gS_{\theta}$ by adding a potential term, as done in \citet{li2022learning}.}.
Solving this constrained optimization inherently alleviates the distribution shift problem.
Since trivial solutions exist for measure preservation, e.g., if $\gS_\theta$ is the identity, 
the trajectory matching component $\gL(\theta)$ of this constrained objective is necessary for producing useful surrogates.

The intractability of the constrained problem in  \Eqref{eq:obj_constrained} leads us to consider a relaxed version by turning the problem into a regularized objective of the form:
\begin{align}\label{eq:obj_dist_reg}
    \gL_\reg^{\D}(\theta) = \gL(\theta) + \reg \D(\mu^*, \mu_{\theta}^{*} ),
\end{align}
where the hyperparameter $\reg$ controls the strength of regularization, and $\D$ is a measure distance / divergence.

This formulation raises three additional questions: \textit{i}) which metric to use for measuring distance between distributions, \textit{ii}) how to sample from $\mu_{\theta}^{*}$, which is unknown, and \textit{iii}) how to estimate the regularization with a finite (and potentially small) number of samples, which is crucial for solving \Eqref{eq:obj_dist_reg} using stochastic optimization pipelines.

\textbf{Measure Distance} \hspace{0.1cm}
Our choice of measure distance needs to satisfy several desiderata, namely it should: \textit{i}) respect the underlying geometry of $\gU$ and support comparison between measures with non-overlapping supports, \textit{ii}) admit an unbiased, sampled-based estimator, \textit{iii}) have low computational complexity with respect to the system dimension and number of samples, \textit{iv)} entail convergence properties on the space of measures defined on $\gU$ (informally, $\D(\mu_{\theta}^*, \mu^*) \rightarrow 0 \implies \mu_{\theta}^* \rightarrow \mu^*$), and \textit{v)} enjoy parametric rates of estimation (i.e., sampling error $|\D - \widehat{\D}|$ is independent of system dimension). 

Some popular notions of distance / divergence from statistical learning theory include the Kullback-Leibler and Hellinger.
However, these do not take into account the distance metric of the space on which the distributions are defined \citep{genevay2018learning, feydy2019interpolating} and in some cases are undefined for non-overlapping supports.

In contrast, Integral Probability Metrics (IPMs; \citet{muller1997integral}) represent a general purpose tool for comparing two distributions.
Among the class of IPMs, the Maximum Mean Discrepancy (MMD; \citet{gretton2012kernel}) stands out as it has a closed form expression and satisfies all our requirements described above.
Deferring several details about the MMD to Appendix \ref{appsec:mmd}, we define it here as,
\begin{align}\label{eq:mmd}
    &\mmd^2(\mu^*, \mu_{\theta}^*) = \E_{\vu, \vu' \sim \mu^*}[\kappa(\vu, \vu')]\\
    &+ \E_{\vv, \vv' \sim \mu_{\theta}^*}[\kappa(\vv, \vv')]
    - 2\E_{\vu \sim \mu^*, \vv \sim \mu_{\theta}^*}[\kappa(\vu, \vv)], \nonumber
\end{align}
where $\kappa: \gU \times \gU \rightarrow \mathbb{R}$ is a kernel.\footnote{The choice of kernel has important practical implications \cite{Feng2020:pmlr-v119-liu20m, Schrab2023:JMLR:v24:21-1289}, and many kernels $\kappa_\sigma$ are controlled by a bandwidth hyperparameter $\sigma$ that should be tuned.
}
For two sets of $n$ samples $\{\vu^{(i)}\}_{i=1}^n \sim \mu^*$ and $\{\vv^{(i)}\}_{i=1}^n \sim \mu_\theta^*$,
\Eqref{eq:mmd} admits an unbiased estimator \cite{gretton2012kernel}:
\begin{align*}
    &\widehat{\mmd}^2(\mu^*, \mu_\theta^*) =
    \frac{1}{n(n-1)}\Bigg [ \sum_{i=1}^{n} \sum_{j\neq i}^{n}[\kappa(\vu^{(i)}, \vu^{(j)})]\\
    &+ \sum_{i=1}^{n}\sum_{j\neq i}^{n}[\kappa(\vv^{(i)}, \vv^{(j)})] \Bigg ]
    - \frac{2}{n^2}\sum_{i=1}^{n}\sum_{j=1}^{n}[\kappa(\vu^{(i)}, \vv^{(j)})],\nonumber
\end{align*}
This estimator can be easily computed in $\gO(n^2)$ operations.
Additionally, the MMD entails convergence properties on the space of measures defined on $\gU$.
That is, it metrizes weak convergence \cite{simon2023metrizing}, and, for characteristic kernels, we have $\mmd(\mu^*, \mu_\theta^*) = 0 \iff \mu^* = \mu_\theta^*$ \cite{sriperumbudur2010hilbert}. The MMD also enjoys parametric rates of estimation, with $\gO(1 / \sqrt{n})$ sampling error \cite{gretton2006kernel,NIPS2016_Tolstikhin}. 
We point out that, in practice, $n$ is the batch size, since we employ stochastic optimization methods.

\paragraph{Approximate Sampling from the Invariant Measure by Time-stepping}
Even though the metric above satisfies several desirable properties, we do not have access to samples of $\mu_{\theta}^*$.
Fortunately, if we assume that $\gS_{\theta}$ has an attractor, then $\gS_{\theta}^k(\vu_0)$ will become a sample of $\mu_{\theta}^*$ for sufficiently large $k$ and $\vu_0$ in the basin of attraction.
Analogous to \Eqref{eq:discrete_propagator}, we have that applying $\gS_{\theta}$ to $\vu_0$ is equivalent to stepping forward in time according to the learned dynamics, i.e., sampling from $\mu_{\theta}^*$ is equivalent to unrolling the trajectory in time.
Then we approximate the invariant measure, $\mu_{\theta}^{*}$, associated with $\gS_{\theta}$ by time-unrolling samples of $\mu^{*}$, i.e., $\left ( \gS_{\theta}^k\right)_{\sharp} \mu^* \approx \mu_{\theta}^*$ for a large 
$k$.
This observation allows us to approximate the regularization term in \Eqref{eq:obj_dist_reg} by 
\begin{align} \label{eq:sampling_approx}
  \D(\mu^*, \mu_{\theta}^*) \approx \D(\mu^*, \left ( \gS_{\theta}^k\right)_{\sharp} \mu^*).
\end{align}

\paragraph{Conditional and Unconditional Regularization}

Using the approximation in \Eqref{eq:sampling_approx}
in the context of a stochastic optimization pipeline requires that we estimate this term with a potentially small batch size.
In fact, besides some simple systems, one typically can only afford small batch sizes, which means we may not be fully capturing both $\mu^*$ and $\mu_{\theta}^*$. 
Although our choice of regularization loss comes with an unbiased estimator with parametric rates of error, in small regimes, estimation error can still diminish its effectiveness at providing a meaningful signal.

We therefore manipulate the expression in \Eqref{eq:obj_dist_reg} to obtain a different yet equivalent loss.
Using the fact that $\mu^*=\gS_{\sharp}\mu^{*}=(\gS^k)_{\sharp}\mu^{*},$ we obtain the equivalent expressions
$\D(\mu^*, \left ( \gS_{\theta}^k\right)_{\sharp} \mu^*)$ and $\D( (\gS^k)_{\sharp} \mu^*,( \gS_{\theta}^k)_{\sharp} \mu^*)$, which we respectively dub as the \textbf{unconditional} and \textbf{conditional regularization}.
We can easily manipulate the latter expression using \Eqref{eq:mmd} to yield
\begin{align} \label{eq:mmd_conditional}
    &\mmd^2((\gS^k)_{\sharp} \mu^*, ( \gS_{\theta}^k)_{\sharp} \mu^*) = \E_{\vu, \vv \sim \mu^*}[\kappa(\gS^k(\vu), \gS^k(\vv))  \nonumber \\
    &\quad+ \kappa(\gS_{\theta}^k(\vu), \gS_{\theta}^k(\vv'))- 2 \kappa(\gS^k(\vu), \gS_{\theta}^k(\vv)],
\end{align}
which can be estimated by extracting a subset of initial conditions, time-evolving them $k$ steps, and then computing the estimator on the time-evolved samples.
We point out that this is equivalent to conditioning the loss on the initial conditions, hence the name conditional regularization.
When using samples to estimate MMD, we use a collection of initial conditions $\{\vu_0^{(i)}\}_{i=1}^n$ for unconditional regularization and a collection of samples time-evolved by the true system, i.e., ones that come from later time steps in training trajectories, $\{\vu_k^{(i)} = \gS^k(\vu_0^{(i)}))\}_{i=1}^n$ for conditional regularization.

Although the expressions for both regularizations are equivalent, they lead to different finite-sample estimators.
The former compares initial samples with ones evolved using $\gS_{\theta},$ while the latter compares samples evolved using both $\gS$ and $\gS_{\theta}$.
In our experiments, we incorporate both terms and explore different weighting schemes.
Empirically, we find that the unconditional regularization is useful when the dynamical system's state dimension is small, and one can afford a large batch; but it becomes uninformative as the dimension of the dynamical system state increases, due to larger distances between samples and sparse coverage of the attractor, which also becomes higher dimensional. 

\paragraph{DySLIM}
Combining the elements above leads us to our proposed objective, DySLIM:
\begin{align}\label{eq:obj_dist_reg_full}
    \widehat{\gL}_\reg^{\D}(\theta) = \widehat{\gL}^{\text{obj}}(\theta) +&  \reg_1 \widehat{\D}(\mu^*, (\gS_\theta^\ell)_{\sharp}\mu^{*})\\ +& \reg_2 \widehat{\D}((\gS^\ell)_{\sharp}\mu^*, (\gS_\theta^\ell)_{\sharp}\mu^{*}), \nonumber
\end{align}
where $\ell$ depends on the type of baseline objective $\widehat{\gL}^{\text{obj}}(\theta)$ used.
The second and third terms in \Eqref{eq:obj_dist_reg_full} correspond to the unconditional and conditional regularization, respectively.
We perform a hyperparameter search over $\reg_1$ and $\reg_2,$ taking $\reg_1 \in \{0, 1\}$ and $
\reg_2 \in \{1, 10, 100, 1000\}.$
Importantly, this objective can be used in conjunction with any of the base losses introduced above.

For the measure distance $\widehat{\D}$ in \Eqref{eq:obj_dist_reg_full}, we use $\widehat{\mmd}^2$ and define $\kappa_{\boldsymbol\sigma}$ as a mixture of rational quadratic kernels \cite{rasmussen2006gaussian, li2015generative}:
\begin{align*}
    \kappa_{\boldsymbol\sigma}(\vu, \vv)
    = \sum_{\sigma_q \in \boldsymbol\sigma}\kappa_{\sigma_q}(\vu, \vv)
    = \sum_{\sigma_q \in \boldsymbol\sigma}\frac{\sigma_q^2}{\sigma_q^2 + ||\vu - \vv||_2^2}, 
\end{align*}
where we select the set $\boldsymbol\sigma$ depending on the dynamical system, see Appendix \ref{appsec:exp_setup} for details.


\section{Experiments}\label{sec:exp}
\textbf{Baselines}\hspace{0.1cm}
Our baseline models are trained with $\widehat{\gL}^{\text{obj}}$, where $\text{obj} \in \{\text{1-step, Curr, Pfwd}\}$ and where $||\cdot||$ is the $\normltwo$ norm.
For each system and objective, the same model architecture, learning rate, and optimizer hyperparameters are used. All experiments are repeated with five different random seeds.

\textbf{Evaluation} \hspace{0.1cm}
We evaluate models both for their `short-term' predictive ability and `long-run' stability.
The former is measured by a cosine similarity statistic between true and predicted trajectories.
The latter is measured by system-specific metrics (see \citet{wan2023debias}) capturing the distributional similarity between true and generated trajectories, along with their visual inspection.
In particular, we use the Sinkhorn Divergence (\textbf{SD}) \cite{genevay2018learning} between the empirical distributions of ground truth and predicted trajectories at various time steps to quantify distributional overlap.
We also use the mean energy log ratio (\textbf{MELR}), which measures the average deviation of the energy at each Fourier mode of the generated snapshots when compared to the ground truth. We also consider its  weighted variant  (\textbf{MELRw}), which up-weights modes with higher energy, in particular the low-frequency modes. We also use the mean of the Frobenius norm of covariance matrix  (\textbf{covRMSE}) that measures the spatial statistical properties of the generated samples. We additionally compute point-wise Wasserstein metrics.
Finally, we consider a time correlation metric (\textbf{TCM}) which provides a measure of temporal behavior, in contrast to most of the metrics above, which are snapshot-based. For more detail on and precise definitions of the evaluation criteria, see Appendix \ref{appsec:eval}.


\subsection{Lorenz 63}\label{subsec:l63}
The first system we examine is the Lorenz 63 model \cite{lorenz1963deterministic}, which is a simplified model of atmospheric convection and is defined by a non-linear ordinary differential equation.
Our models use a simple MLP network, due to the low-dimensional nature of the problem.
For more details about this differential equation and the experimental setup for this system, see Appendix \ref{appsubsec:lorenz63}. 

\Figref{fig:ks_lz63_traj} (b) and (c) demonstrate the improved stability from adding invariant measure regularization to the different training objectives considered in this paper.
In particular, \Figref{fig:ks_lz63_traj} (b) demonstrates improved long-term statistics, as the distribution of points for models trained with DySLIM are closer to that of ground truth trajectories compared to those from models trained with unregularized objectives.
In addition, \Figref{fig:ks_lz63_traj} (c) shows that with DySLIM we obtain the added benefits of improving the short-term model prediction, with longer de-correlation times. For further results with different metrics, see Appendix \ref{sec:app_l64_further_results}.

\subsection{Kuramoto-Sivashinsky}\label{subsec:ks}
We next experiment in the more difficult setting of the high-order PDE known as the Kuramoto-Sivashinsky (KS) equation, which is discussed, along with the experimental setup, in Appendix \ref{appsubsec:ks}.
For this experiment, the 1-step objective proved to be too unstable, even when regularization was applied, so we focus only on Curr and Pfwd objectives.

In \Figref{fig:ks_lz63_traj_stats}, we observe better short-term predictions and improved long-term stability, as measured by lower SD between ground truth and predicted trajectory distributions. 
In \Figref{fig:ks_lz63_traj} (a), we see example trajectories that highlight the difference between models trained with and without regularized objectives.
For the Curr objective, models often diverge and produce numerical instability, while for the Pfwd objective, models deviate from the attractor.
In contrast, the regularized versions of these objectives yield more stable models that remain on the correct attractor manifold.

\subsection{Kolmogorov Flow}\label{subsec:ns}
Finally, we study chaotic 2D fluid flow defined by the Navier-Stokes equations with Kolmogorov forcing.
Information about the PDE and experiment setups is available in Appendix \ref{appsubsec:ns}.
For this system, the SD becomes non-discriminative due to the high-dimension of the state space, so we rely on the other metrics outlined above.

\begin{figure*}
    \centering
    \begin{minipage}[b]{0.65\textwidth} 
        \centering
        \includegraphics[width=\textwidth]{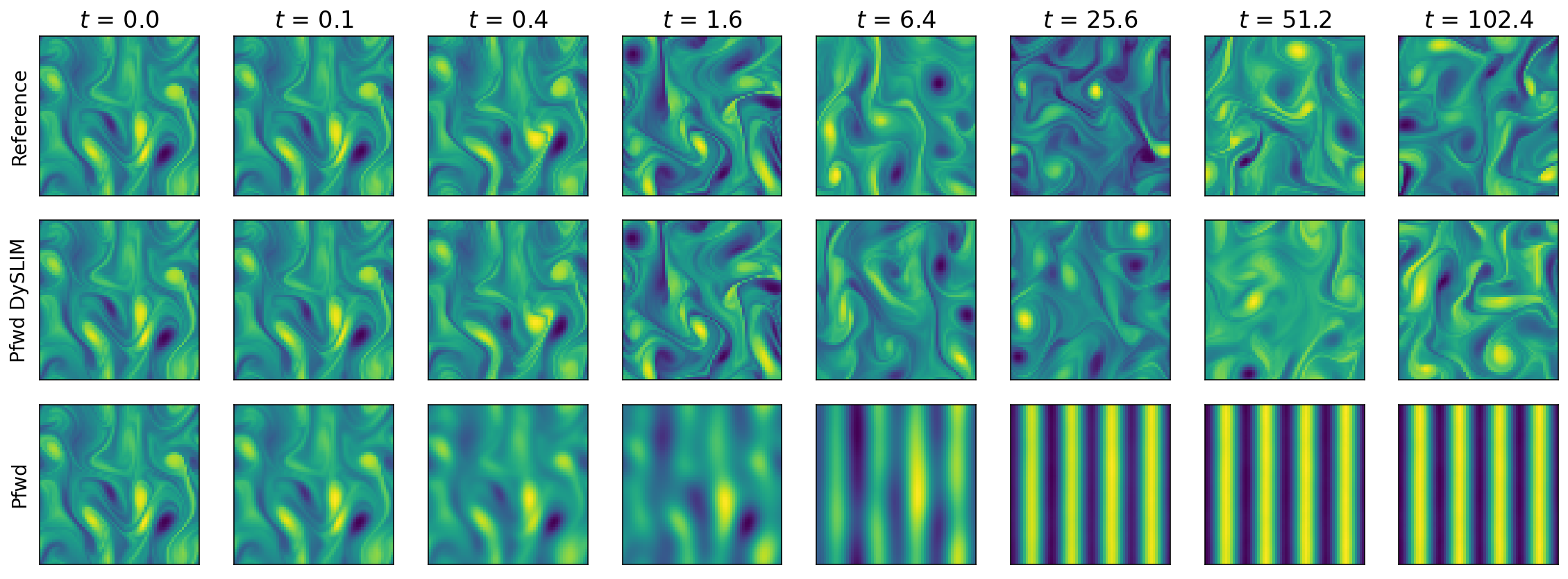} \\ 
        \vfill 
        \includegraphics[width=\textwidth, trim={0mm 0mm 0mm 55mm}, clip]{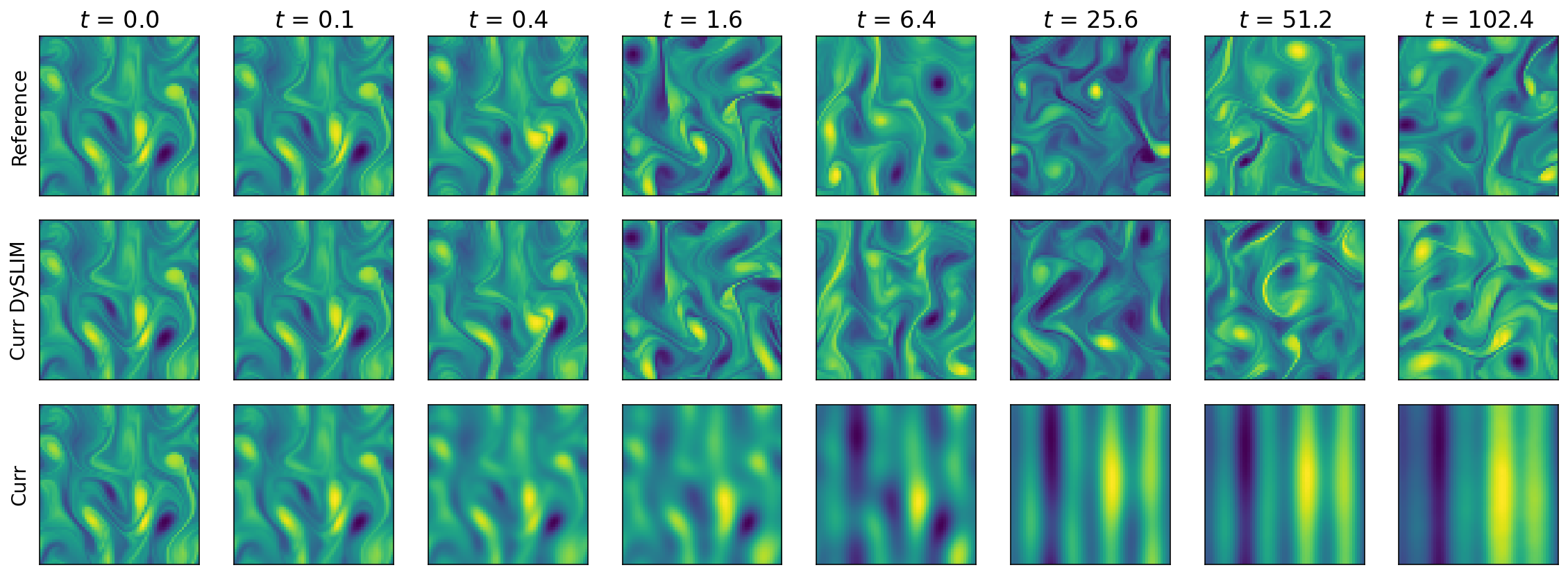} 
    \end{minipage}
    \begin{minipage}[b]{0.26\textwidth} 
        \centering
        \includegraphics[width=\textwidth]{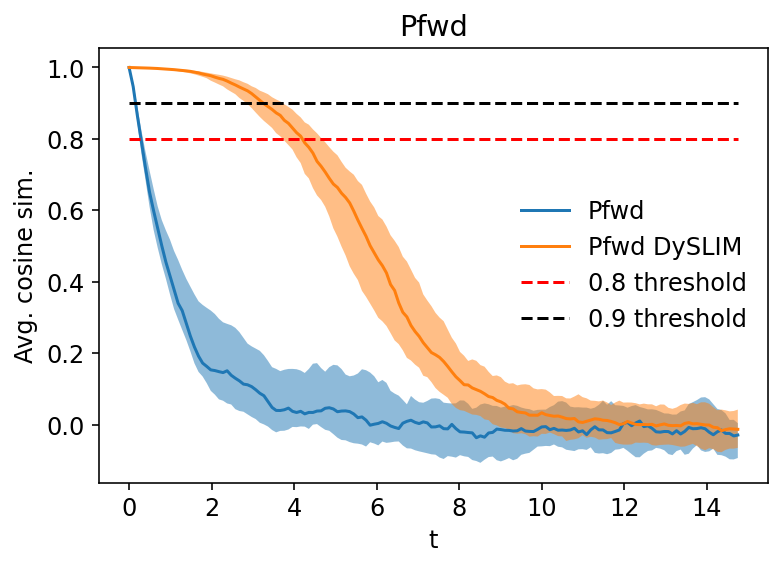} \\ 
        \vfill
        \includegraphics[width=\textwidth]{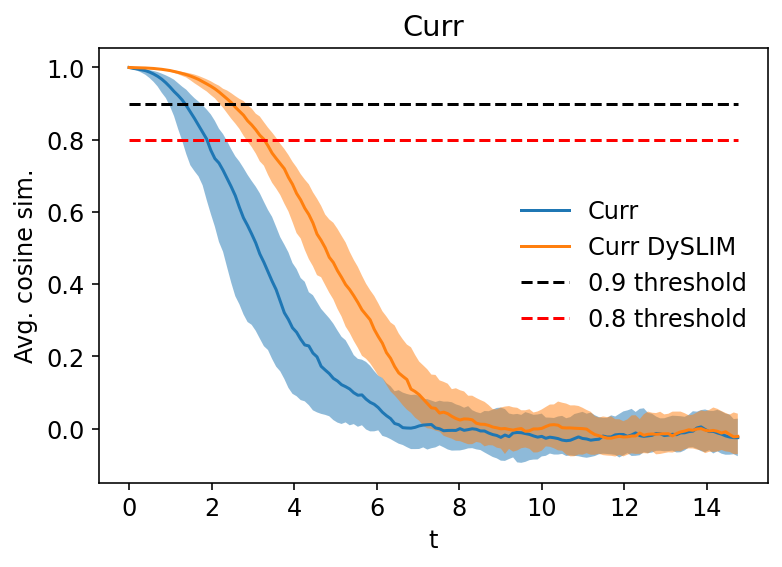} 
    \end{minipage}

    \caption{(\textit{Left}) Sample reference and predicted trajectory across models trained on the Kolmogorov Flow data using the Curr and Pfwd objectives, together with the regularized versions. (\textit{Right}) Evolution of the cosine similarity over time for Curr and Pfwd objectives with and without regularization.
    The solid line is the median among $160$ runs ($32$ trajectories for each of the $5$ random seeds), and the shaded regions correspond to the second and third quartile.
    ($\lambda_1 = 0$, $\lambda_2 = 100$, batch size = 128 and learning rate =  $5\mathrm{e}^{-4})$.}
    \label{fig:ns_evolution}
\end{figure*}

\Figref{fig:ns_evolution} (left) shows typical behavior of the unrolled trained models for a given initial condition: the baselines become highly dissipative and quickly veer towards the mean, whereas DySLIM greatly improves long-term behavior. 
\Figref{fig:ns_evolution} (right) shows a similar results to those in the other experiments: the short-term behavior of the solution is enhanced by the regularization (see \Figref{fig:ns_ablation_cos_sim} for further comparisons).
We find that curriculum training is often worse due to more stringent memory requirements that prevent us from unrolling for longer time-horizons during training.

As an ablation, we 
sweep over different batch sizes and learning rates (see Appendix \ref{appsubsec:ns} for the specific details.)
The results are summarized in Table \ref{tab:ns_stats}, which shows that models trained with DySLIM either have an edge or remain competitive across the spectrum of different learning rates and batch sizes considered.
However, as batch size and learning rate increase, the behavior of the model trained with DySLM remains consistent, whereas the models trained only using the original objective deteriorate quickly.

\begin{table*}[t]
  \centering
  \small
\caption{
Kolmogorov flow: Metrics for 1-step, curriculum, and pushforward objectives without and with regularization ($\lambda_1 = 0$, $\lambda_2 = 100$).
Boldface numbers indicate that the metric is improved by our regularization.
All values displayed are in units of $\times 10^{-2}$.}
\label{tab:ns_stats}
{
\setlength\tabcolsep{4pt} 
{
\setlength{\extrarowheight}{2.5pt}
\vspace{2pt}
\begin{tabular}{lcc|cc|cc|cc|cc|cc}
\toprule
& \multirow{2}{*}{\begin{tabular}[c]{@{}c@{}}Batch size\end{tabular}} & \multirow{2}{*}{\begin{tabular}[c]{@{}c@{}}LR\end{tabular}} & \multicolumn{2}{c|}{\begin{tabular}[c]{@{}c@{}}MELR ($\downarrow$)\end{tabular}} & \multicolumn{2}{c|}{\begin{tabular}[c]{@{}c@{}}MELRw ($\downarrow$)\end{tabular}} & \multicolumn{2}{c|}{\begin{tabular}[c]{@{}c@{}}covRMSE ($\downarrow$)\end{tabular}} & \multicolumn{2}{c|}{\begin{tabular}[c]{@{}c@{}}Wass1 ($\downarrow$)\end{tabular}} & \multicolumn{2}{c}{\begin{tabular}[c]{@{}c@{}}TCM ($\downarrow$)\end{tabular}} \\
                          &                                                                       &                                                                          & Base                                  & DySLIM                                         & Base                                   & DySLIM                                         & Base                                    & DySLIM                                          & Base                                   & DySLIM                                         & Base                                 & DySLIM                                        \\ \midrule
1-step                    & 64                                                                    & 5e-4                                                                     & 2.77                                  & \textbf{1.84}                                  & 0.44                                   & 0.85                                           & 7.93                                    & \textbf{7.30}                                   & 16.2                                   & \textbf{5.55}                                  & 5.39                                 & \textbf{2.45}                                 \\ 
Curr                & 64                                                                    & 5e-4                                                                     & 5.35                                  & \textbf{1.64}                                  & 0.95                                   & \textbf{0.45}                                  & 8.13                                    & \textbf{6.95}                                   & 9.66                                   & \textbf{4.76}                                  & 3.50                                 & \textbf{2.83}                                 \\ 
Pfwd               & 128                                                                   & 1e-4                                                                     & 3.19                                  & \textbf{2.46}                                  & 0.53                                   & 0.53                                           & 6.81                                    & \textbf{6.69}                                   & 4.64                                   & \textbf{4.51}                                  & 3.68                                 & \textbf{0.72}                                 \\ \bottomrule
\end{tabular}
}}
\end{table*}

\section{Related Work} \label{sec:related_work}

\textbf{ROM Methods} \hspace{0.1cm}
Classical reduced order model (ROM) methods build surrogates by identifying low-dimensional linear approximation spaces tailored to representing target system states.
Such spaces are usually derived from data samples \citep{Aubry_FirstPOD, EIM,PGD_shortreview,Farhat_localbases}, and ROMs are obtained by projecting the system equations onto the approximation space \citep{Galerkin}.
Although these methods inherently leverage the linear behavior of underlying dynamics, they have been recently extended to handle mildly non-linear dynamics \citep{GappyPOD, MissPtEst,DEIM,Ayed_Gallinari:2019,Geelen_Wilcox:2022}.
However, their performance deteriorates rapidly for highly non-linear advection-dominated systems, such as KS and Kolmogorov flow \cite{peherstorfer2022breaking}

\textbf{Hybrid Physics-ML} \hspace{0.1cm}
More recent methods hybridize classical numerical methods with contemporary data-driven deep learning techniques \citep{mishra_machine_2019,bar-sinai_learning_2019, bruno_fc-based_2021,kochkov_machine_2021,list_learned_2022,Frezat2022-fs,dresdner2022learning,boral2023neural}.
These approaches \emph{learn} corrections to numerical schemes from high-resolution simulation data, resulting in fast, low-resolution methods with high-accuracy.
However, they require knowledge of the underlying PDE.

\textbf{Pure ML-surrogates and Stabilization Techniques} \hspace{0.1cm}
Recent works have focused on short-term training trajectories using recurrent networks \cite{vlachas2018data}
and reservoir computing techniques \cite{vlachas2020backpropagation, platt2021robust}.
Other approaches seek to regularize the training stage by leveraging properties of the systems.
Such stabilization can be achieved by incorporating noise \cite{sanchez2020learning}, which can be induced by the learned model \cite{brandstetter2022message}; by back-propagating the gradient along many time steps \cite{Kiwon_2020:NIPS}, and learning the dynamics on a latent space \cite{stachenfeld_learned_2022,serrano2023operator}, while promoting smoothness in the latent space \cite{wan2023evolve}. Or by using a generative teacher network \cite{lamb2016professor}, or leveraging an approximate inertial form \cite{lu2017data}.
Related to the Curriculum training baseline, \citet{hess2023generalized} use states that interpolate between model predicted and ground truth states to mitigate gradient explosion.
Finally, a somewhat related stabilization method, introduced in \citet{wang2014least,blonigan2018least}, develops a shadowing technique for sensitivity analysis of long-term averaged gradients.

\textbf{Operator Learning} \hspace{0.1cm}
Neural operators seek to learn the integro-differential operators directly, without explicit PDE-informed components.
These methods often leverage classical fast-methods \citep{FanYing:mnnh2019,graph_fmm:2020,li_fourier_2021,tran_factorized_2021}, or approximation-theoretic structures \citep{deeponet:2021} to achieve computational efficiency.
Some of these techniques have been extended to handle dissipative systems \cite{li2022learning} by hard-coding a dissipative term at both training and inference time.

\begin{table}[]
  \centering
  \small
\caption{
Complexities for each objective, with $d$ denoting state dimension,
$|\theta|$ number of parameters, which is implementation dependent, $NN$ complexity of the neural network for one application, $n_b$ batch size, and $n_t$ number of maximum rollout steps. 
}
\label{tab:complexity}
\begin{tabular}{lll}
\toprule
Objective & Cost $\mathcal{O}(\cdot)$  & Memory footprint $\mathcal{O}(\cdot)$ \\
\midrule
1-step  & $n_b d +  n_b NN$      & $n_b d + n_b |\theta| $   \\
~+ DySLIM & $n_b^2 d + n_b NN$      & $n_b^2 d  + n_b |\theta| $  \\
Curr & $n_t n_b d +  n_t n_b NN $  & $n_t n_b d + n_t n_b |\theta|$ \\
~ + DySLIM     & $n_t n_b^2 d + n_b NN$      & $(n_b^2 + n_b) n_t d +  n_t n_b |\theta|$\\
Pfwd & $n_b d +  n_b NN$      & $n_b d + n_b |\theta|$   \\
~+ DySLIM  & $n_b^2 d + n_b NN$      & $n_b^2 d  + n_b |\theta|$  \\
\bottomrule
\end{tabular}
\end{table}

\textbf{Learning Invariant Measures} \hspace{0.1cm}
\citet{botvinick2023learning} use an Eulerian approach to learn dynamics on invariant measures for low-dimension ODEs using the Feynman-Kac formula coupled with PDE-constrained optimization.
In recent work, \citet{jiang2023training} use neural operators and optimal transport to match the distribution of system-specific summary statistics, which are built using knowledge of the underlying equation driving the system dynamics. 
However, the approach was only applied to small systems with low-dimensional attractors, and it is not clear how well it scales to high-dimensional problems, such as the Kolmogorov flow. 
Similarly, \citet{platt2023constraining} regularize using other invariants, such as the Lyapunov spectrum.

Our proposed method sits between stabilization techniques and learning invariant measures.
In particular, we stabilize the training by implicitly learning the invariant measure of the system along the short-term dynamics.
While our work elects to use the MMD, we note that using other Optimal Transport-based metrics for measure matching, such as the Sinkhorn Divergence, as in \cite{jiang2023training}, is a reasonable choice when the system state is small, and batch size is large, as shown in Appendix \ref{app:SD_reg}.
However, as system size increases and batch sizes decrease (due to memory constraints), we find that the models trained using SD in the regularization perform worse compared to those trained using MMD.

\textbf{MMD in Generative Modeling} 
\hspace{0.1cm}
MMD-based regularization has been used in the context of generative modeling, e.g., the MMD has been used to distinguish between samples of the generated and true distributions \cite{li2015generative,dziugaite2015training,li2017mmd,bińkowski2018demystifying} within the framework of generative adversarial network \cite{goodfellow2020generative}. Additionally, drawing on the close connection between the MMD and a related proper scoring rule \cite{gneiting2007strictly,ramdas2017wasserstein}, \citet{si2021autoregressive, si2023semi} use the energy distance \cite{baringhaus2004new,szekely2004testing}, a special case of the MMD \cite{Sejdinovic:2013_energy_MMD}, to train normalizing flow generative models \cite{rezende2015variational, papamakarios2021normalizing}.

\textbf{Complexity} \hspace{0.1cm}
Our methodology incurs relatively small overhead compared to the baselines.
Table \ref{tab:complexity} shows that our methodology adds an extra cost depending only quadratically on the batch size and linearly on state dimension.
We see this empirically in Table \ref{tab:timing_NS}: wall-clock times are roughly equal with and without our regularization. 

\begin{table}[]
  \centering
  \small
\caption{
Median execution times (rounded to the hour) for 5 training runs (720k steps) on Kolmogorov flow using the PushFwd (max rollout of 10) and Curr (max rollout of 5).
}
\label{tab:timing_NS}
\begin{tabular}{l|cc|c c}
\toprule
\multirow{2}{*}{Batch size} & \multicolumn{2}{c|}{\begin{tabular}[c]{@{}c@{}} Pfwd \end{tabular}} & \multicolumn{2}{c}{\begin{tabular}[c]{@{}c@{}} Curr\end{tabular}}\\
        & Base  & DYSLIM & Base  & DYSLIM \\
\midrule
32       & 40    &  40  &   48  &  48  \\
64       & 74    &  76  &  161  & 163 \\
128      & 145   &  146 &  OOM  & OOM \\
\bottomrule
\end{tabular}
\end{table}

\section{Conclusion}\label{sec:conclusion}
In this work, we have presented a tractable, scalable, and system-agnostic regularized training objective, DySLIM, that leverages a key property of many dynamical systems of interest in order to produce more stable learned models.
Specifically, by pushing learned system models to preserve the invariant measure of an underlying dynamical system, we demonstrated that both short-term predictive capabilities and long-term stability can be improved across a range of well-studied systems, e.g., Lorenz 63, KS, and Kolmogorov Flows.
We hope that the principles of invariant measure preservation introduced in our work, coupled with a tractable and scalable formulation, can serve to stabilize real-world dynamical system models with slowly varying measures, such as those used in global weather prediction.

\section*{Impact Statement}
This paper presents work whose goal is to advance the field of Machine Learning. There are many potential societal consequences of our work, none of which we feel must be specifically highlighted here.
\bibliography{pnas-sample}

\begin{thebibliography}{140}
\providecommand{\natexlab}[1]{#1}
\providecommand{\url}[1]{\texttt{#1}}
\expandafter\ifx\csname urlstyle\endcsname\relax
  \providecommand{\doi}[1]{doi: #1}\else
  \providecommand{\doi}{doi: \begingroup \urlstyle{rm}\Url}\fi

\bibitem[Abadi et~al.(2015)Abadi, Agarwal, Barham, Brevdo, Chen, Citro,
  Corrado, Davis, Dean, Devin, Ghemawat, Goodfellow, Harp, Irving, Isard, Jia,
  Jozefowicz, Kaiser, Kudlur, Levenberg, Man\'{e}, Monga, Moore, Murray, Olah,
  Schuster, Shlens, Steiner, Sutskever, Talwar, Tucker, Vanhoucke, Vasudevan,
  Vi\'{e}gas, Vinyals, Warden, Wattenberg, Wicke, Yu, and
  Zheng]{tensorflow2015-whitepaper}
Abadi, M., Agarwal, A., Barham, P., Brevdo, E., Chen, Z., Citro, C., Corrado,
  G.~S., Davis, A., Dean, J., Devin, M., Ghemawat, S., Goodfellow, I., Harp,
  A., Irving, G., Isard, M., Jia, Y., Jozefowicz, R., Kaiser, L., Kudlur, M.,
  Levenberg, J., Man\'{e}, D., Monga, R., Moore, S., Murray, D., Olah, C.,
  Schuster, M., Shlens, J., Steiner, B., Sutskever, I., Talwar, K., Tucker, P.,
  Vanhoucke, V., Vasudevan, V., Vi\'{e}gas, F., Vinyals, O., Warden, P.,
  Wattenberg, M., Wicke, M., Yu, Y., and Zheng, X.
\newblock {TensorFlow}: Large-scale machine learning on heterogeneous systems,
  2015.
\newblock URL \url{https://www.tensorflow.org/}.
\newblock Software available from tensorflow.org.

\bibitem[Alexander \& Giannakis(2020)Alexander and
  Giannakis]{alexander2020operator}
Alexander, R. and Giannakis, D.
\newblock Operator-theoretic framework for forecasting nonlinear time series
  with kernel analog techniques.
\newblock \emph{Physica D: Nonlinear Phenomena}, 409:\penalty0 132520, 2020.

\bibitem[Amsallem et~al.(2012)Amsallem, Zahr, and Farhat]{Farhat_localbases}
Amsallem, D., Zahr, M.~J., and Farhat, C.
\newblock Nonlinear model order reduction based on local reduced-order bases.
\newblock \emph{International Journal for Numerical Methods in Engineering},
  92\penalty0 (10):\penalty0 891--916, 2012.

\bibitem[Anirudh et~al.(2022)Anirudh, Archibald, Asif, Becker, Benkadda,
  Bremer, Bud{\'e}, Chang, Chen, Churchill, et~al.]{anirudh20222022}
Anirudh, R., Archibald, R., Asif, M.~S., Becker, M.~M., Benkadda, S., Bremer,
  P.-T., Bud{\'e}, R.~H., Chang, C.-S., Chen, L., Churchill, R., et~al.
\newblock 2022 review of data-driven plasma science.
\newblock \emph{arXiv preprint arXiv:2205.15832}, 2022.

\bibitem[Astrid et~al.(2008)Astrid, Weiland, Willcox, and Backx]{MissPtEst}
Astrid, P., Weiland, S., Willcox, K., and Backx, T.
\newblock Missing point estimation in models described by proper orthogonal
  decomposition.
\newblock \emph{IEEE Transactions on Automatic Control}, 53\penalty0
  (10):\penalty0 2237--2251, 2008.
\newblock \doi{10.1109/TAC.2008.2006102}.

\bibitem[Aubry et~al.(1988)Aubry, Holmes, Lumley, and Stone]{Aubry_FirstPOD}
Aubry, N., Holmes, P., Lumley, J.~L., and Stone, E.
\newblock The dynamics of coherent structures in the wall region of a turbulent
  boundary layer.
\newblock \emph{Journal of Fluid Mechanics}, 192:\penalty0 115–173, 1988.
\newblock \doi{10.1017/S0022112088001818}.

\bibitem[Ayed et~al.(2019)Ayed, de~Bezenac, Pajot, Brajard, and
  Gallinari]{Ayed_Gallinari:2019}
Ayed, I., de~Bezenac, E., Pajot, A., Brajard, J., and Gallinari, P.
\newblock Learning dynamical systems from partial observations, 2019.

\bibitem[Bar-Sinai et~al.(2019)Bar-Sinai, Hoyer, Hickey, and
  Brenner]{bar-sinai_learning_2019}
Bar-Sinai, Y., Hoyer, S., Hickey, J., and Brenner, M.~P.
\newblock Learning data-driven discretizations for partial differential
  equations.
\newblock \emph{Proceedings of the National Academy of Sciences}, 116\penalty0
  (31):\penalty0 15344--15349, July 2019.
\newblock ISSN 0027-8424, 1091-6490.
\newblock \doi{10.1073/pnas.1814058116}.
\newblock URL \url{https://pnas.org/doi/full/10.1073/pnas.1814058116}.

\bibitem[Baringhaus \& Franz(2004)Baringhaus and Franz]{baringhaus2004new}
Baringhaus, L. and Franz, C.
\newblock On a new multivariate two-sample test.
\newblock \emph{Journal of multivariate analysis}, 88\penalty0 (1):\penalty0
  190--206, 2004.

\bibitem[Barrault et~al.(2004)Barrault, Maday, Nguyen, and Patera]{EIM}
Barrault, M., Maday, Y., Nguyen, N.~C., and Patera, A.~T.
\newblock An ‘empirical interpolation’ method: application to efficient
  reduced-basis discretization of partial differential equations.
\newblock \emph{Comptes Rendus Mathematique}, 339\penalty0 (9):\penalty0
  667--672, 2004.
\newblock ISSN 1631-073X.
\newblock \doi{https://doi.org/10.1016/j.crma.2004.08.006}.
\newblock URL
  \url{https://www.sciencedirect.com/science/article/pii/S1631073X04004248}.

\bibitem[Bi et~al.(2023)Bi, Xie, Zhang, Chen, Gu, and Tian]{bi2023accurate}
Bi, K., Xie, L., Zhang, H., Chen, X., Gu, X., and Tian, Q.
\newblock Accurate medium-range global weather forecasting with 3d neural
  networks.
\newblock \emph{Nature}, 619\penalty0 (7970):\penalty0 533--538, 2023.

\bibitem[Bińkowski et~al.(2018)Bińkowski, Sutherland, Arbel, and
  Gretton]{bińkowski2018demystifying}
Bińkowski, M., Sutherland, D.~J., Arbel, M., and Gretton, A.
\newblock Demystifying {MMD} {GAN}s.
\newblock In \emph{International Conference on Learning Representations}, 2018.
\newblock URL \url{https://openreview.net/forum?id=r1lUOzWCW}.

\bibitem[Blonigan et~al.(2018)Blonigan, Wang, Nielsen, and
  Diskin]{blonigan2018least}
Blonigan, P.~J., Wang, Q., Nielsen, E.~J., and Diskin, B.
\newblock Least-squares shadowing sensitivity analysis of chaotic flow around a
  two-dimensional airfoil.
\newblock \emph{AIAA Journal}, 56\penalty0 (2):\penalty0 658--672, 2018.

\bibitem[Bollt(2021)]{bollt2021explaining}
Bollt, E.
\newblock On explaining the surprising success of reservoir computing
  forecaster of chaos? the universal machine learning dynamical system with
  contrast to var and dmd.
\newblock \emph{Chaos: An Interdisciplinary Journal of Nonlinear Science},
  31\penalty0 (1), 2021.

\bibitem[Bonev et~al.(2023)Bonev, Kurth, Hundt, Pathak, Baust, Kashinath, and
  Anandkumar]{bonev2023spherical}
Bonev, B., Kurth, T., Hundt, C., Pathak, J., Baust, M., Kashinath, K., and
  Anandkumar, A.
\newblock Spherical {F}ourier neural operators: Learning stable dynamics on the
  sphere.
\newblock \emph{arXiv preprint arXiv:2306.03838}, 2023.

\bibitem[Boral et~al.(2023)Boral, Wan, Zepeda-N{\'u}{\~n}ez, Lottes, Wang,
  Chen, Anderson, and Sha]{boral2023neural}
Boral, A., Wan, Z.~Y., Zepeda-N{\'u}{\~n}ez, L., Lottes, J., Wang, Q., Chen,
  Y.-F., Anderson, J.~R., and Sha, F.
\newblock Neural ideal large eddy simulation: Modeling turbulence with neural
  stochastic differential equations.
\newblock In \emph{Thirty-seventh Conference on Neural Information Processing
  Systems}, 2023.
\newblock URL \url{https://openreview.net/forum?id=x6cOcxRnxG}.

\bibitem[Botvinick-Greenhouse et~al.(2023)Botvinick-Greenhouse, Martin, and
  Yang]{botvinick2023learning}
Botvinick-Greenhouse, J., Martin, R., and Yang, Y.
\newblock Learning dynamics on invariant measures using {PDE}-constrained
  optimization.
\newblock \emph{Chaos: An Interdisciplinary Journal of Nonlinear Science},
  33\penalty0 (6), 2023.

\bibitem[Bradbury et~al.(2018)Bradbury, Frostig, Hawkins, Johnson, Leary,
  Maclaurin, Necula, Paszke, Vander{P}las, Wanderman-{M}ilne, and
  Zhang]{jax2018github}
Bradbury, J., Frostig, R., Hawkins, P., Johnson, M.~J., Leary, C., Maclaurin,
  D., Necula, G., Paszke, A., Vander{P}las, J., Wanderman-{M}ilne, S., and
  Zhang, Q.
\newblock {JAX}: composable transformations of {P}ython+{N}um{P}y programs,
  2018.
\newblock URL \url{http://github.com/google/jax}.

\bibitem[Brandstetter et~al.(2022)Brandstetter, Worrall, and
  Welling]{brandstetter2022message}
Brandstetter, J., Worrall, D., and Welling, M.
\newblock Message passing neural {PDE} solvers.
\newblock \emph{arXiv preprint arXiv:2202.03376}, 2022.

\bibitem[Brezis(2018)]{brezis2018remarks}
Brezis, H.
\newblock Remarks on the {M}onge--{K}antorovich problem in the discrete
  setting.
\newblock \emph{Comptes Rendus. Math{\'e}matique}, 356\penalty0 (2):\penalty0
  207--213, 2018.

\bibitem[Bruno et~al.(2021)Bruno, Hesthaven, and
  Leibovici]{bruno_fc-based_2021}
Bruno, O.~P., Hesthaven, J.~S., and Leibovici, D.~V.
\newblock \href{http://arxiv.org/abs/2111.01315}{{FC}-based shock-dynamics
  solver with neural-network localized artificial-viscosity assignment}.
\newblock \emph{arXiv:2111.01315 [cs, math]}, November 2021.
\newblock arXiv: 2111.01315.

\bibitem[Brunton \& Kutz(2022)Brunton and Kutz]{brunton2022data}
Brunton, S.~L. and Kutz, J.~N.
\newblock \emph{Data-driven science and engineering: Machine learning,
  dynamical systems, and control}.
\newblock Cambridge University Press, 2022.

\bibitem[Canuto et~al.(2007)Canuto, Hussaini, Quarteroni, and
  Zang]{Canuto:2007_spectral_methods}
Canuto, C.~G., Hussaini, M.~Y., Quarteroni, A.~M., and Zang, T.~A.
\newblock \emph{Spectral Methods: Evolution to Complex Geometries and
  Applications to Fluid Dynamics (Scientific Computation)}.
\newblock Springer-Verlag, Berlin, Heidelberg, 2007.
\newblock ISBN 3540307273.

\bibitem[Chaturantabut \& Sorensen(2010)Chaturantabut and Sorensen]{DEIM}
Chaturantabut, S. and Sorensen, D.~C.
\newblock Nonlinear model reduction via discrete empirical interpolation.
\newblock \emph{SIAM Journal on Scientific Computing}, 32\penalty0
  (5):\penalty0 2737--2764, 2010.
\newblock \doi{10.1137/090766498}.
\newblock URL \url{https://doi.org/10.1137/090766498}.

\bibitem[Chen et~al.(2016)Chen, Xu, Zhang, and Guestrin]{chen2016training}
Chen, T., Xu, B., Zhang, C., and Guestrin, C.
\newblock Training deep nets with sublinear memory cost.
\newblock \emph{arXiv preprint arXiv:1604.06174}, 2016.

\bibitem[Chen et~al.(2020)Chen, Zhang, Wang, and E]{chen2020deepks}
Chen, Y., Zhang, L., Wang, H., and E, W.
\newblock {DeePKS}: A comprehensive data-driven approach toward chemically
  accurate density functional theory.
\newblock \emph{Journal of Chemical Theory and Computation}, 17\penalty0
  (1):\penalty0 170--181, 2020.

\bibitem[Chinesta et~al.(2011)Chinesta, Ladeveze, and Cueto]{PGD_shortreview}
Chinesta, F., Ladeveze, P., and Cueto, E.
\newblock A short review on model order reduction based on proper generalized
  decomposition.
\newblock \emph{Archives of Computational Methods in Engineering}, 18\penalty0
  (4):\penalty0 395--404, 2011.

\bibitem[Cooley \& Tukey(1965)Cooley and Tukey]{Cooley_Tukey:1965}
Cooley, J.~W. and Tukey, J.~W.
\newblock An algorithm for the machine calculation of complex {F}ourier series.
\newblock \emph{Math. Comput.}, 19\penalty0 (90):\penalty0 297--301, 1965.
\newblock ISSN 00255718, 10886842.
\newblock URL \url{http://www.jstor.org/stable/2003354}.

\bibitem[Cuturi(2013)]{cuturi2013sinkhorn}
Cuturi, M.
\newblock Sinkhorn distances: Lightspeed computation of optimal transport.
\newblock \emph{Advances in neural information processing systems}, 26, 2013.

\bibitem[Cuturi et~al.(2022)Cuturi, Meng-Papaxanthos, Tian, Bunne, Davis, and
  Teboul]{cuturi2022optimal}
Cuturi, M., Meng-Papaxanthos, L., Tian, Y., Bunne, C., Davis, G., and Teboul,
  O.
\newblock Optimal transport tools (ott): A {JAX} toolbox for all things
  {W}asserstein.
\newblock \emph{arXiv preprint arXiv:2201.12324}, 2022.

\bibitem[Dresdner et~al.(2022)Dresdner, Kochkov, Norgaard,
  Zepeda-N{\'u}{\~n}ez, Smith, Brenner, and Hoyer]{dresdner2022learning}
Dresdner, G., Kochkov, D., Norgaard, P., Zepeda-N{\'u}{\~n}ez, L., Smith,
  J.~A., Brenner, M.~P., and Hoyer, S.
\newblock Learning to correct spectral methods for simulating turbulent flows.
\newblock \emph{arXiv preprint arXiv:2207.00556}, 2022.

\bibitem[Dziugaite et~al.(2015)Dziugaite, Roy, and
  Ghahramani]{dziugaite2015training}
Dziugaite, G.~K., Roy, D.~M., and Ghahramani, Z.
\newblock Training generative neural networks via maximum mean discrepancy
  optimization.
\newblock \emph{arXiv preprint arXiv:1505.03906}, 2015.

\bibitem[Fan et~al.(2020)Fan, Jiang, Zhang, Wang, and Lai]{fan2020long}
Fan, H., Jiang, J., Zhang, C., Wang, X., and Lai, Y.-C.
\newblock Long-term prediction of chaotic systems with machine learning.
\newblock \emph{Physical Review Research}, 2\penalty0 (1):\penalty0 012080,
  2020.

\bibitem[Fan et~al.(2019)Fan, Feliu-Fab{\`a}, Lin, Ying, and
  Zepeda-N{\'u}{\~{n}}ez]{FanYing:mnnh2019}
Fan, Y., Feliu-Fab{\`a}, J., Lin, L., Ying, L., and Zepeda-N{\'u}{\~{n}}ez, L.
\newblock A multiscale neural network based on hierarchical nested bases.
\newblock \emph{Research in the Mathematical Sciences}, 6, Mar. 2019.
\newblock ISSN 2197-9847.
\newblock \doi{10.1007/s40687-019-0183-3}.

\bibitem[Farmer \& Sidorowich(1987)Farmer and Sidorowich]{Farmer1987}
Farmer, J.~D. and Sidorowich, J.~J.
\newblock Predicting chaotic time series.
\newblock \emph{Phys. Rev. Lett.}, 59:\penalty0 845--848, Aug 1987.
\newblock \doi{10.1103/PhysRevLett.59.845}.
\newblock URL \url{https://link.aps.org/doi/10.1103/PhysRevLett.59.845}.

\bibitem[Ferrario(2008)]{ferrario2008invariant}
Ferrario, B.
\newblock Invariant measures for a stochastic kuramoto--sivashinsky equation.
\newblock \emph{Stochastic analysis and applications}, 26\penalty0
  (2):\penalty0 379--407, 2008.

\bibitem[Feydy et~al.(2019)Feydy, S{\'e}journ{\'e}, Vialard, Amari, Trouv{\'e},
  and Peyr{\'e}]{feydy2019interpolating}
Feydy, J., S{\'e}journ{\'e}, T., Vialard, F.-X., Amari, S.-i., Trouv{\'e}, A.,
  and Peyr{\'e}, G.
\newblock Interpolating between optimal transport and {MMD} using sinkhorn
  divergences.
\newblock In \emph{The 22nd International Conference on Artificial Intelligence
  and Statistics}, pp.\  2681--2690. PMLR, 2019.

\bibitem[Frezat et~al.(2022)Frezat, Le~Sommer, Fablet, Balarac, and
  Lguensat]{Frezat2022-fs}
Frezat, H., Le~Sommer, J., Fablet, R., Balarac, G., and Lguensat, R.
\newblock \href{http://arxiv.org/abs/2204.03911}{A posteriori learning for
  quasi-geostrophic turbulence parametrization}.
\newblock \emph{arXiv}, April 2022.

\bibitem[Galerkin(1915)]{Galerkin}
Galerkin, B.~G.
\newblock Series occurring in various questions concerning the elastic
  equilibrium of rods and plates.
\newblock \emph{Vestnik Inzhenernov i Tekhnikov}, 19:\penalty0 897--908, 1915.

\bibitem[Geelen et~al.(2022)Geelen, Wright, and Willcox]{Geelen_Wilcox:2022}
Geelen, R., Wright, S., and Willcox, K.
\newblock Operator inference for non-intrusive model reduction with nonlinear
  manifolds.
\newblock \emph{arXiv:2205.02304 [math.NA]}, May 2022.
\newblock URL \url{https://arxiv.org/abs/2205.02304}.
\newblock arXiv: 2205.02304.

\bibitem[Genevay et~al.(2018)Genevay, Peyr{\'e}, and
  Cuturi]{genevay2018learning}
Genevay, A., Peyr{\'e}, G., and Cuturi, M.
\newblock Learning generative models with sinkhorn divergences.
\newblock In \emph{International Conference on Artificial Intelligence and
  Statistics}, pp.\  1608--1617. PMLR, 2018.

\bibitem[Ghadami \& Epureanu(2022)Ghadami and Epureanu]{ghadami2022data}
Ghadami, A. and Epureanu, B.~I.
\newblock Data-driven prediction in dynamical systems: recent developments.
\newblock \emph{Philosophical Transactions of the Royal Society A},
  380\penalty0 (2229):\penalty0 20210213, 2022.

\bibitem[Gneiting \& Raftery(2007)Gneiting and Raftery]{gneiting2007strictly}
Gneiting, T. and Raftery, A.~E.
\newblock Strictly proper scoring rules, prediction, and estimation.
\newblock \emph{Journal of the American statistical Association}, pp.\
  359--378, 2007.

\bibitem[Goodfellow et~al.(2020)Goodfellow, Pouget-Abadie, Mirza, Xu,
  Warde-Farley, Ozair, Courville, and Bengio]{goodfellow2020generative}
Goodfellow, I., Pouget-Abadie, J., Mirza, M., Xu, B., Warde-Farley, D., Ozair,
  S., Courville, A., and Bengio, Y.
\newblock Generative adversarial networks.
\newblock \emph{Communications of the ACM}, 63\penalty0 (11):\penalty0
  139--144, 2020.

\bibitem[Gretton et~al.(2006)Gretton, Borgwardt, Rasch, Sch{\"o}lkopf, and
  Smola]{gretton2006kernel}
Gretton, A., Borgwardt, K., Rasch, M., Sch{\"o}lkopf, B., and Smola, A.
\newblock A kernel method for the two-sample-problem.
\newblock \emph{Advances in neural information processing systems}, 19, 2006.

\bibitem[Gretton et~al.(2012)Gretton, Borgwardt, Rasch, Sch{\"o}lkopf, and
  Smola]{gretton2012kernel}
Gretton, A., Borgwardt, K.~M., Rasch, M.~J., Sch{\"o}lkopf, B., and Smola, A.
\newblock A kernel two-sample test.
\newblock \emph{The Journal of Machine Learning Research}, 13\penalty0
  (1):\penalty0 723--773, 2012.

\bibitem[Hara \& Kokubu(2022)Hara and Kokubu]{hara2022learning}
Hara, M. and Kokubu, H.
\newblock Learning dynamics by reservoir computing (in memory of prof. pavol
  brunovsk{\`y}).
\newblock \emph{Journal of Dynamics and Differential Equations}, pp.\  1--26,
  2022.

\bibitem[Harris et~al.(2020)Harris, Millman, van~der Walt, Gommers, Virtanen,
  Cournapeau, Wieser, Taylor, Berg, Smith, Kern, Picus, Hoyer, van Kerkwijk,
  Brett, Haldane, del R{\'{i}}o, Wiebe, Peterson, G{\'{e}}rard-Marchant,
  Sheppard, Reddy, Weckesser, Abbasi, Gohlke, and Oliphant]{harris2020array}
Harris, C.~R., Millman, K.~J., van~der Walt, S.~J., Gommers, R., Virtanen, P.,
  Cournapeau, D., Wieser, E., Taylor, J., Berg, S., Smith, N.~J., Kern, R.,
  Picus, M., Hoyer, S., van Kerkwijk, M.~H., Brett, M., Haldane, A., del
  R{\'{i}}o, J.~F., Wiebe, M., Peterson, P., G{\'{e}}rard-Marchant, P.,
  Sheppard, K., Reddy, T., Weckesser, W., Abbasi, H., Gohlke, C., and Oliphant,
  T.~E.
\newblock Array programming with {NumPy}.
\newblock \emph{Nature}, 585\penalty0 (7825):\penalty0 357--362, September
  2020.
\newblock \doi{10.1038/s41586-020-2649-2}.
\newblock URL \url{https://doi.org/10.1038/s41586-020-2649-2}.

\bibitem[Hawkins(2021)]{hawkins2021attractors}
Hawkins, J.
\newblock Attractors in dynamical systems.
\newblock \emph{Ergodic dynamics: From basic theory to applications}, pp.\
  27--39, 2021.

\bibitem[Heek et~al.(2023)Heek, Levskaya, Oliver, Ritter, Rondepierre, Steiner,
  and van {Z}ee]{flax2020github}
Heek, J., Levskaya, A., Oliver, A., Ritter, M., Rondepierre, B., Steiner, A.,
  and van {Z}ee, M.
\newblock {F}lax: A neural network library and ecosystem for {JAX}, 2023.
\newblock URL \url{http://github.com/google/flax}.

\bibitem[Hess et~al.(2023)Hess, Monfared, Brenner, and
  Durstewitz]{hess2023generalized}
Hess, F., Monfared, Z., Brenner, M., and Durstewitz, D.
\newblock Generalized teacher forcing for learning chaotic dynamics.
\newblock \emph{arXiv preprint arXiv:2306.04406}, 2023.

\bibitem[Hoyer \& Hamman(2017)Hoyer and Hamman]{hoyer2017xarray}
Hoyer, S. and Hamman, J.
\newblock xarray: {N-D} labeled arrays and datasets in {Python}.
\newblock \emph{Journal of Open Research Software}, 5\penalty0 (1), 2017.
\newblock \doi{10.5334/jors.148}.
\newblock URL \url{https://doi.org/10.5334/jors.148}.

\bibitem[Hunter(2007)]{Hunter:2007}
Hunter, J.~D.
\newblock Matplotlib: A 2d graphics environment.
\newblock \emph{Computing in Science \& Engineering}, 9\penalty0 (3):\penalty0
  90--95, 2007.
\newblock \doi{10.1109/MCSE.2007.55}.

\bibitem[Jia et~al.(2020)Jia, Wang, Chen, Lu, Lin, Car, Weinan, and
  Zhang]{jia2020pushing}
Jia, W., Wang, H., Chen, M., Lu, D., Lin, L., Car, R., Weinan, E., and Zhang,
  L.
\newblock Pushing the limit of molecular dynamics with ab initio accuracy to
  100 million atoms with machine learning.
\newblock In \emph{{SC}20: International conference for high performance
  computing, networking, storage and analysis}, pp.\  1--14. IEEE, 2020.

\bibitem[Jiang et~al.(2023)Jiang, Lu, Orlova, and Willett]{jiang2023training}
Jiang, R., Lu, P.~Y., Orlova, E., and Willett, R.
\newblock Training neural operators to preserve invariant measures of chaotic
  attractors.
\newblock \emph{arXiv preprint arXiv:2306.01187}, 2023.

\bibitem[Kaiser et~al.(2021)Kaiser, Kutz, and Brunton]{kaiser2021data}
Kaiser, E., Kutz, J.~N., and Brunton, S.~L.
\newblock Data-driven discovery of {K}oopman eigenfunctions for control.
\newblock \emph{Machine Learning: Science and Technology}, 2\penalty0
  (3):\penalty0 035023, 2021.

\bibitem[Kantorovich(1942)]{kantorovich1942translocation}
Kantorovich, L.~V.
\newblock On the translocation of masses.
\newblock In \emph{Dokl. Akad. Nauk. USSR (NS)}, volume~37, pp.\  199--201,
  1942.

\bibitem[Keisler(2022)]{keisler2022forecasting}
Keisler, R.
\newblock Forecasting global weather with graph neural networks.
\newblock \emph{arXiv preprint arXiv:2202.07575}, 2022.

\bibitem[Kingma \& Ba(2014)Kingma and Ba]{kingma2014adam}
Kingma, D.~P. and Ba, J.
\newblock Adam: A method for stochastic optimization.
\newblock \emph{arXiv preprint arXiv:1412.6980}, 2014.

\bibitem[Kochkov et~al.(2021)Kochkov, Smith, Alieva, Wang, Brenner, and
  Hoyer]{kochkov_machine_2021}
Kochkov, D., Smith, J.~A., Alieva, A., Wang, Q., Brenner, M.~P., and Hoyer, S.
\newblock Machine learning--accelerated computational fluid dynamics.
\newblock \emph{Proc. Natl. Acad. Sci. U. S. A.}, 118\penalty0 (21), May 2021.
\newblock URL \url{https://www.pnas.org/content/118/21/e2101784118}.

\bibitem[Kochkov et~al.(2023)Kochkov, Yuval, Langmore, Norgaard, Smith, Mooers,
  Lottes, Rasp, D{\"u}ben, Kl{\"o}wer, et~al.]{kochkov2023neural}
Kochkov, D., Yuval, J., Langmore, I., Norgaard, P., Smith, J., Mooers, G.,
  Lottes, J., Rasp, S., D{\"u}ben, P., Kl{\"o}wer, M., et~al.
\newblock Neural general circulation models.
\newblock \emph{arXiv preprint arXiv:2311.07222}, 2023.

\bibitem[Koopman(1931)]{koopman1931hamiltonian}
Koopman, B.~O.
\newblock Hamiltonian systems and transformation in hilbert space.
\newblock \emph{Proceedings of the National Academy of Sciences}, 17\penalty0
  (5):\penalty0 315--318, 1931.

\bibitem[Krishnapriyan et~al.(2021)Krishnapriyan, Gholami, Zhe, Kirby, and
  Mahoney]{krishnapriyan2021characterizing}
Krishnapriyan, A., Gholami, A., Zhe, S., Kirby, R., and Mahoney, M.~W.
\newblock Characterizing possible failure modes in physics-informed neural
  networks.
\newblock \emph{Advances in Neural Information Processing Systems},
  34:\penalty0 26548--26560, 2021.

\bibitem[Kuramoto(1978)]{kuramoto1978diffusion}
Kuramoto, Y.
\newblock Diffusion-induced chaos in reaction systems.
\newblock \emph{Progress of Theoretical Physics Supplement}, 64:\penalty0
  346--367, 1978.

\bibitem[Kutta(1901)]{kutta1901beitrag}
Kutta, W.
\newblock \emph{Beitrag zur n{\"a}herungsweisen Integration totaler
  Differentialgleichungen}.
\newblock Teubner, 1901.

\bibitem[Lam et~al.(2022)Lam, Sanchez-Gonzalez, Willson, Wirnsberger,
  Fortunato, Alet, Ravuri, Ewalds, Eaton-Rosen, Hu, et~al.]{lam2022graphcast}
Lam, R., Sanchez-Gonzalez, A., Willson, M., Wirnsberger, P., Fortunato, M.,
  Alet, F., Ravuri, S., Ewalds, T., Eaton-Rosen, Z., Hu, W., et~al.
\newblock Graphcast: Learning skillful medium-range global weather forecasting.
\newblock \emph{arXiv preprint arXiv:2212.12794}, 2022.

\bibitem[Lamb et~al.(2016)Lamb, ALIAS PARTH~GOYAL, Zhang, Zhang, Courville, and
  Bengio]{lamb2016professor}
Lamb, A.~M., ALIAS PARTH~GOYAL, A.~G., Zhang, Y., Zhang, S., Courville, A.~C.,
  and Bengio, Y.
\newblock Professor forcing: A new algorithm for training recurrent networks.
\newblock \emph{Advances in neural information processing systems}, 29, 2016.

\bibitem[Li et~al.(2017)Li, Chang, Cheng, Yang, and P{\'o}czos]{li2017mmd}
Li, C.-L., Chang, W.-C., Cheng, Y., Yang, Y., and P{\'o}czos, B.
\newblock {MMD GAN}: Towards deeper understanding of moment matching network.
\newblock \emph{Advances in neural information processing systems}, 30, 2017.

\bibitem[Li et~al.(2015)Li, Swersky, and Zemel]{li2015generative}
Li, Y., Swersky, K., and Zemel, R.
\newblock Generative moment matching networks.
\newblock In \emph{International conference on machine learning}, pp.\
  1718--1727. PMLR, 2015.

\bibitem[Li et~al.(2020)Li, Kovachki, Azizzadenesheli, Liu, Bhattacharya,
  Stuart, and Anandkumar]{graph_fmm:2020}
Li, Z., Kovachki, N., Azizzadenesheli, K., Liu, B., Bhattacharya, K., Stuart,
  A., and Anandkumar, A.
\newblock Multipole graph neural operator for parametric partial differential
  equations.
\newblock In \emph{Proceedings of the 34th International Conference on Neural
  Information Processing Systems}, NIPS'20, Red Hook, NY, USA, 2020. Curran
  Associates Inc.
\newblock ISBN 9781713829546.

\bibitem[Li et~al.(2021)Li, Kovachki, Azizzadenesheli, Liu, Bhattacharya,
  Stuart, and Anandkumar]{li_fourier_2021}
Li, Z., Kovachki, N., Azizzadenesheli, K., Liu, B., Bhattacharya, K., Stuart,
  A., and Anandkumar, A.
\newblock Fourier {Neural} {Operator} for {Parametric} {Partial} {Differential}
  {Equations}.
\newblock \emph{arXiv:2010.08895 [cs, math]}, May 2021.
\newblock URL \url{http://arxiv.org/abs/2010.08895}.
\newblock arXiv: 2010.08895.

\bibitem[Li et~al.(2022)Li, Liu-Schiaffini, Kovachki, Azizzadenesheli, Liu,
  Bhattacharya, Stuart, and Anandkumar]{li2022learning}
Li, Z., Liu-Schiaffini, M., Kovachki, N., Azizzadenesheli, K., Liu, B.,
  Bhattacharya, K., Stuart, A., and Anandkumar, A.
\newblock Learning chaotic dynamics in dissipative systems.
\newblock \emph{Advances in Neural Information Processing Systems},
  35:\penalty0 16768--16781, 2022.

\bibitem[List et~al.(2022)List, Chen, and Thuerey]{list_learned_2022}
List, B., Chen, L.-W., and Thuerey, N.
\newblock Learned {Turbulence} {Modelling} with {Differentiable} {Fluid}
  {Solvers}.
\newblock \emph{arXiv:2202.06988 [physics]}, February 2022.
\newblock URL \url{http://arxiv.org/abs/2202.06988}.
\newblock arXiv: 2202.06988.

\bibitem[Liu et~al.(2020)Liu, Xu, Lu, Zhang, Gretton, and
  Sutherland]{Feng2020:pmlr-v119-liu20m}
Liu, F., Xu, W., Lu, J., Zhang, G., Gretton, A., and Sutherland, D.~J.
\newblock Learning deep kernels for non-parametric two-sample tests.
\newblock In \emph{Proceedings of the 37th International Conference on Machine
  Learning}, volume 119, pp.\  6316--6326. PMLR, 2020.

\bibitem[Lorenz(1963)]{lorenz1963deterministic}
Lorenz, E.~N.
\newblock Deterministic nonperiodic flow.
\newblock \emph{Journal of atmospheric sciences}, 20\penalty0 (2):\penalty0
  130--141, 1963.

\bibitem[Lu et~al.(2017)Lu, Lin, and Chorin]{lu2017data}
Lu, F., Lin, K.~K., and Chorin, A.~J.
\newblock Data-based stochastic model reduction for the
  {K}uramoto--{S}ivashinsky equation.
\newblock \emph{Physica D: Nonlinear Phenomena}, 340:\penalty0 46--57, 2017.

\bibitem[Lu et~al.(2021)Lu, Jin, Pang, Zhang, and Karniadakis]{deeponet:2021}
Lu, L., Jin, P., Pang, G., Zhang, Z., and Karniadakis, G.~E.
\newblock Learning nonlinear operators via {DeepONet} based on the universal
  approximation theorem of operators.
\newblock \emph{Nature Machine Intelligence}, 3\penalty0 (3):\penalty0
  218--229, 2021.

\bibitem[Luzzatto et~al.(2005)Luzzatto, Melbourne, and
  Paccaut]{luzzatto2005lorenz}
Luzzatto, S., Melbourne, I., and Paccaut, F.
\newblock The lorenz attractor is mixing.
\newblock \emph{Communications in Mathematical Physics}, 260:\penalty0
  393--401, 2005.

\bibitem[Mathews et~al.(2021)Mathews, Francisquez, Hughes, Hatch, Zhu, and
  Rogers]{mathews2021uncovering}
Mathews, A., Francisquez, M., Hughes, J.~W., Hatch, D.~R., Zhu, B., and Rogers,
  B.~N.
\newblock Uncovering turbulent plasma dynamics via deep learning from partial
  observations.
\newblock \emph{Physical Review E}, 104\penalty0 (2):\penalty0 025205, 2021.

\bibitem[Medio \& Lines(2001)Medio and Lines]{medio2001nonlinear}
Medio, A. and Lines, M.
\newblock \emph{Nonlinear dynamics: A primer}.
\newblock Cambridge University Press, 2001.

\bibitem[Merchant et~al.(2023)Merchant, Batzner, Schoenholz, Aykol, Cheon, and
  Cubuk]{merchant2023scaling}
Merchant, A., Batzner, S., Schoenholz, S.~S., Aykol, M., Cheon, G., and Cubuk,
  E.~D.
\newblock Scaling deep learning for materials discovery.
\newblock \emph{Nature}, pp.\  1--6, 2023.

\bibitem[Mikhaeil et~al.(2022)Mikhaeil, Monfared, and
  Durstewitz]{mikhaeil2022difficulty}
Mikhaeil, J., Monfared, Z., and Durstewitz, D.
\newblock On the difficulty of learning chaotic dynamics with {RNNs}.
\newblock \emph{Advances in Neural Information Processing Systems},
  35:\penalty0 11297--11312, 2022.

\bibitem[Mishra(2018)]{mishra_machine_2019}
Mishra, S.
\newblock A machine learning framework for data driven acceleration of
  computations of differential equations.
\newblock \emph{Mathematics in Engineering}, 1\penalty0 (1):\penalty0 118--146,
  2018.
\newblock ISSN 2640-3501.
\newblock \doi{10.3934/Mine.2018.1.118}.
\newblock URL
  \url{https://www.aimspress.com/article/doi/10.3934/Mine.2018.1.118}.

\bibitem[M{\"u}ller(1997)]{muller1997integral}
M{\"u}ller, A.
\newblock Integral probability metrics and their generating classes of
  functions.
\newblock \emph{Advances in applied probability}, 29\penalty0 (2):\penalty0
  429--443, 1997.

\bibitem[Nghiem et~al.(2023)Nghiem, Drgo{\v{n}}a, Jones, Nagy, Schwan, Dey,
  Chakrabarty, Di~Cairano, Paulson, Carron, et~al.]{nghiem2023physics}
Nghiem, T.~X., Drgo{\v{n}}a, J., Jones, C., Nagy, Z., Schwan, R., Dey, B.,
  Chakrabarty, A., Di~Cairano, S., Paulson, J.~A., Carron, A., et~al.
\newblock Physics-informed machine learning for modeling and control of
  dynamical systems.
\newblock \emph{arXiv preprint arXiv:2306.13867}, 2023.

\bibitem[Obukhov(1983)]{obukhov1983kolmogorov}
Obukhov, A.
\newblock Kolmogorov flow and laboratory simulation of it.
\newblock \emph{Russ. Math. Surv}, 38\penalty0 (4):\penalty0 113--126, 1983.

\bibitem[pandas~development team(2020)]{reback2020pandas}
pandas~development team, T.
\newblock pandas-dev/pandas: Pandas, February 2020.
\newblock URL \url{https://doi.org/10.5281/zenodo.3509134}.

\bibitem[Papageorgiou \& Smyrlis(1991)Papageorgiou and
  Smyrlis]{papageorgiou1991route}
Papageorgiou, D.~T. and Smyrlis, Y.~S.
\newblock The route to chaos for the kuramoto-sivashinsky equation.
\newblock \emph{Theoretical and Computational Fluid Dynamics}, 3\penalty0
  (1):\penalty0 15--42, 1991.

\bibitem[Papamakarios et~al.(2021)Papamakarios, Nalisnick, Rezende, Mohamed,
  and Lakshminarayanan]{papamakarios2021normalizing}
Papamakarios, G., Nalisnick, E., Rezende, D.~J., Mohamed, S., and
  Lakshminarayanan, B.
\newblock Normalizing flows for probabilistic modeling and inference.
\newblock \emph{The Journal of Machine Learning Research}, 22\penalty0
  (1):\penalty0 2617--2680, 2021.

\bibitem[Pascanu et~al.(2013)Pascanu, Mikolov, and
  Bengio]{pascanu2013difficulty}
Pascanu, R., Mikolov, T., and Bengio, Y.
\newblock On the difficulty of training recurrent neural networks.
\newblock In \emph{International conference on machine learning}, pp.\
  1310--1318. Pmlr, 2013.

\bibitem[Pathak et~al.(2017)Pathak, Lu, Hunt, Girvan, and Ott]{pathak2017using}
Pathak, J., Lu, Z., Hunt, B.~R., Girvan, M., and Ott, E.
\newblock Using machine learning to replicate chaotic attractors and calculate
  {L}yapunov exponents from data.
\newblock \emph{Chaos: An Interdisciplinary Journal of Nonlinear Science},
  27\penalty0 (12), 2017.

\bibitem[Pathak et~al.(2022)Pathak, Subramanian, Harrington, Raja,
  Chattopadhyay, Mardani, Kurth, Hall, Li, Azizzadenesheli,
  et~al.]{pathak2022fourcastnet}
Pathak, J., Subramanian, S., Harrington, P., Raja, S., Chattopadhyay, A.,
  Mardani, M., Kurth, T., Hall, D., Li, Z., Azizzadenesheli, K., et~al.
\newblock Fourcastnet: A global data-driven high-resolution weather model using
  adaptive fourier neural operators.
\newblock \emph{arXiv preprint arXiv:2202.11214}, 2022.

\bibitem[Pearson(1901)]{PCA_Pearson_1901}
Pearson, K.
\newblock Liii. on lines and planes of closest fit to systems of points in
  space.
\newblock \emph{The London, Edinburgh, and Dublin Philosophical Magazine and
  Journal of Science}, 2\penalty0 (11):\penalty0 559--572, 1901.
\newblock \doi{10.1080/14786440109462720}.
\newblock URL \url{https://doi.org/10.1080/14786440109462720}.

\bibitem[Peherstorfer(2022)]{peherstorfer2022breaking}
Peherstorfer, B.
\newblock Breaking the kolmogorov barrier with nonlinear model reduction.
\newblock \emph{Notices of the American Mathematical Society}, 69\penalty0
  (5):\penalty0 725--733, 2022.

\bibitem[Peyr{\'e} et~al.(2019)Peyr{\'e}, Cuturi,
  et~al.]{peyre2019computational}
Peyr{\'e}, G., Cuturi, M., et~al.
\newblock Computational optimal transport: With applications to data science.
\newblock \emph{Foundations and Trends{\textregistered} in Machine Learning},
  11\penalty0 (5-6):\penalty0 355--607, 2019.

\bibitem[Platt et~al.(2021)Platt, Wong, Clark, Penny, and
  Abarbanel]{platt2021robust}
Platt, J.~A., Wong, A., Clark, R., Penny, S.~G., and Abarbanel, H.~D.
\newblock Robust forecasting using predictive generalized synchronization in
  reservoir computing.
\newblock \emph{Chaos: An Interdisciplinary Journal of Nonlinear Science},
  31\penalty0 (12), 2021.

\bibitem[Platt et~al.(2023)Platt, Penny, Smith, Chen, and
  Abarbanel]{platt2023constraining}
Platt, J.~A., Penny, S.~G., Smith, T.~A., Chen, T.-C., and Abarbanel, H.~D.
\newblock Constraining chaos: Enforcing dynamical invariants in the training of
  recurrent neural networks.
\newblock \emph{arXiv preprint arXiv:2304.12865}, 2023.

\bibitem[Rajendra \& Brahmajirao(2020)Rajendra and
  Brahmajirao]{rajendra2020modeling}
Rajendra, P. and Brahmajirao, V.
\newblock Modeling of dynamical systems through deep learning.
\newblock \emph{Biophysical Reviews}, 12\penalty0 (6):\penalty0 1311--1320,
  2020.

\bibitem[Ramachandran et~al.(2018)Ramachandran, Zoph, and Le]{ramachandran18}
Ramachandran, P., Zoph, B., and Le, Q.~V.
\newblock Searching for activation functions.
\newblock In \emph{6th International Conference on Learning Representations,
  {ICLR} 2018, Vancouver, BC, Canada, April 30 - May 3, 2018, Workshop Track
  Proceedings}. OpenReview.net, 2018.
\newblock URL \url{https://openreview.net/forum?id=Hkuq2EkPf}.

\bibitem[Ramdas et~al.(2017)Ramdas, Garc{\'\i}a~Trillos, and
  Cuturi]{ramdas2017wasserstein}
Ramdas, A., Garc{\'\i}a~Trillos, N., and Cuturi, M.
\newblock On wasserstein two-sample testing and related families of
  nonparametric tests.
\newblock \emph{Entropy}, 19\penalty0 (2):\penalty0 47, 2017.

\bibitem[Rasmussen et~al.(2006)Rasmussen, Williams,
  et~al.]{rasmussen2006gaussian}
Rasmussen, C.~E., Williams, C.~K., et~al.
\newblock \emph{Gaussian processes for machine learning}, volume~1.
\newblock Springer, 2006.

\bibitem[Rezende \& Mohamed(2015)Rezende and Mohamed]{rezende2015variational}
Rezende, D. and Mohamed, S.
\newblock Variational inference with normalizing flows.
\newblock In \emph{International conference on machine learning}, pp.\
  1530--1538. PMLR, 2015.

\bibitem[Roy \& Rana(2021)Roy and Rana]{roy2021machine}
Roy, S. and Rana, D.
\newblock Machine learning in nonlinear dynamical systems.
\newblock \emph{Resonance}, 26\penalty0 (7):\penalty0 953--970, 2021.

\bibitem[Runge(1895)]{runge1895numerische}
Runge, C.
\newblock {\"U}ber die numerische aufl{\"o}sung von differentialgleichungen.
\newblock \emph{Mathematische Annalen}, 46\penalty0 (2):\penalty0 167--178,
  1895.

\bibitem[Sanchez-Gonzalez et~al.(2020)Sanchez-Gonzalez, Godwin, Pfaff, Ying,
  Leskovec, and Battaglia]{sanchez2020learning}
Sanchez-Gonzalez, A., Godwin, J., Pfaff, T., Ying, R., Leskovec, J., and
  Battaglia, P.
\newblock Learning to simulate complex physics with graph networks.
\newblock In \emph{International conference on machine learning}, pp.\
  8459--8468. PMLR, 2020.

\bibitem[Santambrogio(2015)]{santambrogio2015optimal}
Santambrogio, F.
\newblock Optimal transport for applied mathematicians.
\newblock \emph{Birk{\"a}user, NY}, 55\penalty0 (58-63):\penalty0 94, 2015.

\bibitem[Schmid(2010)]{schmid_dmd_2010}
Schmid, P.~J.
\newblock Dynamic mode decomposition of numerical and experimental data.
\newblock \emph{Journal of Fluid Mechanics}, 656:\penalty0 5–28, 2010.
\newblock \doi{10.1017/S0022112010001217}.

\bibitem[Schmid(2022)]{schmid2022dynamic}
Schmid, P.~J.
\newblock Dynamic mode decomposition and its variants.
\newblock \emph{Annual Review of Fluid Mechanics}, 54:\penalty0 225--254, 2022.

\bibitem[Schrab et~al.(2023)Schrab, Kim, Albert, Laurent, Guedj, and
  Gretton]{Schrab2023:JMLR:v24:21-1289}
Schrab, A., Kim, I., Albert, M., Laurent, B., Guedj, B., and Gretton, A.
\newblock {MMD} aggregated two-sample test.
\newblock \emph{Journal of Machine Learning Research}, 24\penalty0
  (194):\penalty0 1--81, 2023.

\bibitem[Sejdinovic et~al.(2013)Sejdinovic, Sriperumbudur, Gretton, and
  Fukumizu]{Sejdinovic:2013_energy_MMD}
Sejdinovic, D., Sriperumbudur, B., Gretton, A., and Fukumizu, K.
\newblock {Equivalence of distance-based and {RKHS}-based statistics in
  hypothesis testing}.
\newblock \emph{The Annals of Statistics}, 41\penalty0 (5):\penalty0 2263 --
  2291, 2013.
\newblock \doi{10.1214/13-AOS1140}.
\newblock URL \url{https://doi.org/10.1214/13-AOS1140}.

\bibitem[Serrano et~al.(2023)Serrano, Boudec, Koupa{\"\i}, Wang, Yin, Vittaut,
  and Gallinari]{serrano2023operator}
Serrano, L., Boudec, L.~L., Koupa{\"\i}, A.~K., Wang, T.~X., Yin, Y., Vittaut,
  J.-N., and Gallinari, P.
\newblock Operator learning with neural fields: Tackling {PDE}s on general
  geometries.
\newblock \emph{arXiv preprint arXiv:2306.07266}, 2023.

\bibitem[Si et~al.(2021)Si, Bishop, and Kuleshov]{si2021autoregressive}
Si, P., Bishop, A., and Kuleshov, V.
\newblock Autoregressive quantile flows for predictive uncertainty estimation.
\newblock \emph{arXiv preprint arXiv:2112.04643}, 2021.

\bibitem[Si et~al.(2023)Si, Chen, Sahoo, Schiff, and Kuleshov]{si2023semi}
Si, P., Chen, Z., Sahoo, S.~S., Schiff, Y., and Kuleshov, V.
\newblock Semi-autoregressive energy flows: exploring likelihood-free training
  of normalizing flows.
\newblock In \emph{International Conference on Machine Learning}, pp.\
  31732--31753. PMLR, 2023.

\bibitem[Simon-Gabriel et~al.(2023)Simon-Gabriel, Barp, Sch{\"o}lkopf, and
  Mackey]{simon2023metrizing}
Simon-Gabriel, C.-J., Barp, A., Sch{\"o}lkopf, B., and Mackey, L.
\newblock Metrizing weak convergence with maximum mean discrepancies.
\newblock \emph{Journal of Machine Learning Research}, 24\penalty0
  (184):\penalty0 1--20, 2023.

\bibitem[Sivashinsky(1988)]{ashinsky1988nonlinear}
Sivashinsky, G.~I.
\newblock Nonlinear analysis of hydrodynamic instability in laminar flames—i.
  derivation of basic equations.
\newblock In \emph{Dynamics of Curved Fronts}, pp.\  459--488. Elsevier, 1988.

\bibitem[Sriperumbudur et~al.(2009)Sriperumbudur, Fukumizu, Gretton,
  Sch{\"o}lkopf, and Lanckriet]{sriperumbudur2009integral}
Sriperumbudur, B.~K., Fukumizu, K., Gretton, A., Sch{\"o}lkopf, B., and
  Lanckriet, G.~R.
\newblock On integral probability metrics,$\backslash$phi-divergences and
  binary classification.
\newblock \emph{arXiv preprint arXiv:0901.2698}, 2009.

\bibitem[Sriperumbudur et~al.(2010)Sriperumbudur, Gretton, Fukumizu,
  Sch{\"o}lkopf, and Lanckriet]{sriperumbudur2010hilbert}
Sriperumbudur, B.~K., Gretton, A., Fukumizu, K., Sch{\"o}lkopf, B., and
  Lanckriet, G.~R.
\newblock Hilbert space embeddings and metrics on probability measures.
\newblock \emph{The Journal of Machine Learning Research}, 11:\penalty0
  1517--1561, 2010.

\bibitem[Stachenfeld et~al.(2022)Stachenfeld, Fielding, Kochkov, Cranmer,
  Pfaff, Godwin, Cui, Ho, Battaglia, and
  Sanchez-Gonzalez]{stachenfeld_learned_2022}
Stachenfeld, K., Fielding, D.~B., Kochkov, D., Cranmer, M., Pfaff, T., Godwin,
  J., Cui, C., Ho, S., Battaglia, P., and Sanchez-Gonzalez, A.
\newblock Learned {Coarse} {Models} for {Efficient} {Turbulence} {Simulation}.
\newblock \emph{arXiv:2112.15275 [physics]}, January 2022.
\newblock URL \url{http://arxiv.org/abs/2112.15275}.
\newblock arXiv: 2112.15275.

\bibitem[Strogatz(2018)]{strogatz2018nonlinear}
Strogatz, S.~H.
\newblock \emph{Nonlinear dynamics and chaos with student solutions manual:
  With applications to physics, biology, chemistry, and engineering}.
\newblock CRC press, 2018.

\bibitem[Stuart \& Humphries(1998)Stuart and Humphries]{stuart1998dynamical}
Stuart, A. and Humphries, A.~R.
\newblock \emph{Dynamical systems and numerical analysis}, volume~2.
\newblock Cambridge University Press, 1998.

\bibitem[Stuart(1994)]{stuart1994numerical}
Stuart, A.~M.
\newblock Numerical analysis of dynamical systems.
\newblock \emph{Acta numerica}, 3:\penalty0 467--572, 1994.

\bibitem[Sz{\'e}kely et~al.(2004)Sz{\'e}kely, Rizzo,
  et~al.]{szekely2004testing}
Sz{\'e}kely, G.~J., Rizzo, M.~L., et~al.
\newblock Testing for equal distributions in high dimension.
\newblock \emph{InterStat}, 5\penalty0 (16.10):\penalty0 1249--1272, 2004.

\bibitem[Temam(2012)]{temam2012infinite}
Temam, R.
\newblock \emph{Infinite-dimensional dynamical systems in mechanics and
  physics}, volume~68.
\newblock Springer Science \& Business Media, 2012.

\bibitem[Tolstikhin et~al.(2016)Tolstikhin, Sriperumbudur, and
  Sch\"{o}lkopf]{NIPS2016_Tolstikhin}
Tolstikhin, I.~O., Sriperumbudur, B.~K., and Sch\"{o}lkopf, B.
\newblock Minimax estimation of maximum mean discrepancy with radial kernels.
\newblock In Lee, D., Sugiyama, M., Luxburg, U., Guyon, I., and Garnett, R.
  (eds.), \emph{Advances in Neural Information Processing Systems}, volume~29,
  2016.
\newblock URL
  \url{https://proceedings.neurips.cc/paper_files/paper/2016/file/5055cbf43fac3f7e2336b27310f0b9ef-Paper.pdf}.

\bibitem[Tran et~al.(2021)Tran, Mathews, Xie, and Ong]{tran_factorized_2021}
Tran, A., Mathews, A., Xie, L., and Ong, C.~S.
\newblock \href{http://arxiv.org/abs/2111.13802}{Factorized {Fourier} {Neural}
  {Operators}}.
\newblock \emph{arXiv:2111.13802 [cs]}, November 2021.
\newblock arXiv: 2111.13802.

\bibitem[Trefethen(2000)]{trefethen_spectral_2000}
Trefethen, L.~N.
\newblock \emph{\href{https://doi.org/10.1137/1.9780898719598}{Spectral
  {Methods} in {MATLAB}}}.
\newblock Society for industrial and applied mathematics (SIAM), 2000.

\bibitem[Tucker(1999)]{tucker1999lorenz}
Tucker, W.
\newblock The {L}orenz attractor exists.
\newblock \emph{Comptes Rendus de l'Acad{\'e}mie des Sciences-Series
  I-Mathematics}, 328\penalty0 (12):\penalty0 1197--1202, 1999.

\bibitem[Tucker(2002)]{tucker2002rigorous}
Tucker, W.
\newblock A rigorous {ODE} solver and {S}male’s 14th problem.
\newblock \emph{Foundations of Computational Mathematics}, 2:\penalty0 53--117,
  2002.

\bibitem[Um et~al.(2020)Um, Brand, Fei, Holl, and Thuerey]{Kiwon_2020:NIPS}
Um, K., Brand, R., Fei, Y.~R., Holl, P., and Thuerey, N.
\newblock Solver-in-the-loop: Learning from differentiable physics to interact
  with iterative pde-solvers.
\newblock In Larochelle, H., Ranzato, M., Hadsell, R., Balcan, M., and Lin, H.
  (eds.), \emph{Advances in Neural Information Processing Systems}, volume~33,
  pp.\  6111--6122. Curran Associates, Inc., 2020.
\newblock URL
  \url{https://proceedings.neurips.cc/paper_files/paper/2020/file/43e4e6a6f341e00671e123714de019a8-Paper.pdf}.

\bibitem[Virtanen et~al.(2020)Virtanen, Gommers, Oliphant, Haberland, Reddy,
  Cournapeau, Burovski, Peterson, Weckesser, Bright, {van der Walt}, Brett,
  Wilson, Millman, Mayorov, Nelson, Jones, Kern, Larson, Carey, Polat, Feng,
  Moore, {VanderPlas}, Laxalde, Perktold, Cimrman, Henriksen, Quintero, Harris,
  Archibald, Ribeiro, Pedregosa, {van Mulbregt}, and {SciPy 1.0
  Contributors}]{2020SciPy-NMeth}
Virtanen, P., Gommers, R., Oliphant, T.~E., Haberland, M., Reddy, T.,
  Cournapeau, D., Burovski, E., Peterson, P., Weckesser, W., Bright, J., {van
  der Walt}, S.~J., Brett, M., Wilson, J., Millman, K.~J., Mayorov, N., Nelson,
  A. R.~J., Jones, E., Kern, R., Larson, E., Carey, C.~J., Polat, {\.I}., Feng,
  Y., Moore, E.~W., {VanderPlas}, J., Laxalde, D., Perktold, J., Cimrman, R.,
  Henriksen, I., Quintero, E.~A., Harris, C.~R., Archibald, A.~M., Ribeiro,
  A.~H., Pedregosa, F., {van Mulbregt}, P., and {SciPy 1.0 Contributors}.
\newblock {{SciPy} 1.0: Fundamental Algorithms for Scientific Computing in
  Python}.
\newblock \emph{Nature Methods}, 17:\penalty0 261--272, 2020.
\newblock \doi{10.1038/s41592-019-0686-2}.

\bibitem[Vlachas et~al.(2018)Vlachas, Byeon, Wan, Sapsis, and
  Koumoutsakos]{vlachas2018data}
Vlachas, P.~R., Byeon, W., Wan, Z.~Y., Sapsis, T.~P., and Koumoutsakos, P.
\newblock Data-driven forecasting of high-dimensional chaotic systems with long
  short-term memory networks.
\newblock \emph{Proceedings of the Royal Society A: Mathematical, Physical and
  Engineering Sciences}, 474\penalty0 (2213):\penalty0 20170844, 2018.

\bibitem[Vlachas et~al.(2020)Vlachas, Pathak, Hunt, Sapsis, Girvan, Ott, and
  Koumoutsakos]{vlachas2020backpropagation}
Vlachas, P.~R., Pathak, J., Hunt, B.~R., Sapsis, T.~P., Girvan, M., Ott, E.,
  and Koumoutsakos, P.
\newblock Backpropagation algorithms and reservoir computing in recurrent
  neural networks for the forecasting of complex spatiotemporal dynamics.
\newblock \emph{Neural Networks}, 126:\penalty0 191--217, 2020.

\bibitem[Wan et~al.(2023{\natexlab{a}})Wan, Baptista, Boral, Chen, Anderson,
  Sha, and Zepeda-N{\'u}{\~n}ez]{wan2023debias}
Wan, Z.~Y., Baptista, R., Boral, A., Chen, Y.-F., Anderson, J., Sha, F., and
  Zepeda-N{\'u}{\~n}ez, L.
\newblock Debias coarsely, sample conditionally: Statistical downscaling
  through optimal transport and probabilistic diffusion models.
\newblock In \emph{Thirty-seventh Conference on Neural Information Processing
  Systems}, 2023{\natexlab{a}}.
\newblock URL \url{https://openreview.net/forum?id=5NxJuc0T1P}.

\bibitem[Wan et~al.(2023{\natexlab{b}})Wan, Zepeda-N{\'u}{\~n}ez, Boral, and
  Sha]{wan2023evolve}
Wan, Z.~Y., Zepeda-N{\'u}{\~n}ez, L., Boral, A., and Sha, F.
\newblock Evolve smoothly, fit consistently: Learning smooth latent dynamics
  for advection-dominated systems.
\newblock In \emph{The Eleventh International Conference on Learning
  Representations}, 2023{\natexlab{b}}.

\bibitem[Wang et~al.(2014)Wang, Hu, and Blonigan]{wang2014least}
Wang, Q., Hu, R., and Blonigan, P.
\newblock Least squares shadowing sensitivity analysis of chaotic limit cycle
  oscillations.
\newblock \emph{Journal of Computational Physics}, 267:\penalty0 210--224,
  2014.

\bibitem[Waskom(2021)]{Waskom2021}
Waskom, M.~L.
\newblock seaborn: statistical data visualization.
\newblock \emph{Journal of Open Source Software}, 6\penalty0 (60):\penalty0
  3021, 2021.
\newblock \doi{10.21105/joss.03021}.
\newblock URL \url{https://doi.org/10.21105/joss.03021}.

\bibitem[Watt-Meyer et~al.(2023)Watt-Meyer, Dresdner, McGibbon, Clark, Henn,
  Duncan, Brenowitz, Kashinath, Pritchard, Bonev, et~al.]{watt2023ace}
Watt-Meyer, O., Dresdner, G., McGibbon, J., Clark, S.~K., Henn, B., Duncan, J.,
  Brenowitz, N.~D., Kashinath, K., Pritchard, M.~S., Bonev, B., et~al.
\newblock {ACE}: A fast, skillful learned global atmospheric model for climate
  prediction.
\newblock \emph{arXiv preprint arXiv:2310.02074}, 2023.

\bibitem[Weinan \& Liu(2002)Weinan and Liu]{weinan2002gibbsian}
Weinan, E. and Liu, D.
\newblock Gibbsian dynamics and invariant measures for stochastic dissipative
  {PDE}s.
\newblock \emph{Journal of Statistical Physics}, 108:\penalty0 1125--1156,
  2002.

\bibitem[Willcox(2006)]{GappyPOD}
Willcox, K.
\newblock Unsteady flow sensing and estimation via the gappy proper orthogonal
  decomposition.
\newblock \emph{Computers \& Fluids}, 35\penalty0 (2):\penalty0 208--226, 2006.
\newblock ISSN 0045-7930.
\newblock \doi{https://doi.org/10.1016/j.compfluid.2004.11.006}.
\newblock URL
  \url{https://www.sciencedirect.com/science/article/pii/S0045793005000113}.

\bibitem[Zepeda-N{\'u}{\~n}ez et~al.(2021)Zepeda-N{\'u}{\~n}ez, Chen, Zhang,
  Jia, Zhang, and Lin]{zepeda2021deep}
Zepeda-N{\'u}{\~n}ez, L., Chen, Y., Zhang, J., Jia, W., Zhang, L., and Lin, L.
\newblock Deep density: circumventing the {K}ohn-{S}ham equations via symmetry
  preserving neural networks.
\newblock \emph{Journal of Computational Physics}, 443:\penalty0 110523, 2021.

\end{thebibliography}
\bibliographystyle{icml2024}

\newpage
\appendix
\onecolumn
\section{Relation to the Pushforward trick} \label{sec:app_pfwd_trick}
We note that the pushforward trick \cite{brandstetter2022message} can be reformulated using our framework as a weak measure fitting loss.
A finite-sample approximation of
$\E_{\vu_0 \sim \mu^*}[||\gS_\theta(\sg(\gS^{k-1}_{\theta}(\vu_0))) - \vu_{k}||^2]$ is an upper bound of the discrete Wasserstein-2 distance between $\mu^{\star}$ and the approximation of $\mu_{\theta}^{\star}$. Formally, we have that
\begin{equation}
\mathcal{L}^{\text{Pfwd}}(\theta) = \E_{\vu_0 \sim \mu^*}[||\gS_\theta(\sg(\gS^{k-1}_{\theta}(\vu_0))) - \vu_{k}||^2] \gtrsim \mathcal{W}_2 (\mu, \mu_{\theta}^*) := \inf_{\gamma \in \Gamma(\mu^*, \mu_{\theta}^*)} \int \| \vu - \vv \|^2 \, d \gamma(\vu, \vv), 
\end{equation}
where $\Gamma(\mu^*, \mu_{\theta}^*)$ is the set of all couplings between $\mu^*$ and $ \mu_{\theta}^*$.

By relying on an estimate of the loss, we have that for a given set of initial conditions $\{\vu^{(i)}\}_{i=1}^n \sim \mu^*$,
\begin{equation}
    \widehat{\mathcal{L}}^{\text{Pfwd}}(\theta) :=
    n^{-1}\Sigma_{i=1}^n \|\gS_\theta(\sg(\gS^{k-1}_{\theta}(\vu^{(i)}))) -  \gS^k(\vu^{(i)}) \|^2 = 
    n^{-1}\Sigma_{i=1}^n \| \gS_{\theta}^k(\vu^{(i)}) -  \gS^k(\vu^{(i)}) \|^2,
\end{equation}
which can be lower bounded by the following
\begin{align*}
n^{-1}\Sigma_{i=1}^n \| \gS_{\theta}^k(\vu^{(i)}) -  \gS^k(\vu^{(i)}) \|^2 \geq
n^{-1}\min_{\pi} \Sigma_{i=1}^n \| \gS^k(\vu^{(i)}) - \gS_{\theta}^k(\vu^{(\pi(i))}) \|^2, 
\end{align*}
where $\pi$ is a permutation operator. Given that we are in the discrete setting where the Monge and Kantorovich problems are equivalent \citep{brezis2018remarks}, we have that
\begin{equation}
    n^{-1}\min_{\pi} \Sigma_{i=1}^n \| \gS^k(\vu^{(i)}) - \gS_{\theta}^k(\vu^{(\pi(i))}) \|^2 = 
\inf_{T \in \Pi} \Sigma_{i, j} T_{i,j} C_{i, j} := \widehat{W}_2((\gS_{\theta}^k)_{\sharp} \mu^*, (\gS^k)_{\sharp} \mu^*),
\end{equation}
where $\Pi$ is the set of all valid discrete transport maps (i.e., matrices that satisfy $T_{i,j}\geq 0$, $\sum_j T_{i, j} = \sum_i T_{i,j}  = 1$), $C$ is the quadratic cost function ($C_{i,j} = \| \gS^k(\vu^{(i)}) - \gS_{\theta}^k(\vu^{(j)}) \|^2$), and $\widehat{W}_2$ is a discrete estimate of the Wasserstein-2 metric.

We can further refine this expression using the same approximation as in \Eqref{eq:sampling_approx}, i.e., $\gS_{\theta}^k(\vu^{(i)}) \sim \mu_{\theta}^{*}$ for large $k$ and $\gS^k(\vu^{(i)}) \sim \mu^*$, we have that 
\begin{align} \label{eq:app_discrete_ot}
\widehat{W}_2((\gS_{\theta}^k)_{\sharp} \mu^*, (\gS^k)_{\sharp} \mu^*) \approx \widehat{W}_2(\mu_{\theta}^*, \mu^*).
\end{align}

Therefore, in summary we have that
\begin{equation}
 \hat{\mathcal{L}}^{\text{Pfwd}}(\theta) \gtrsim \widehat{W}_2(\mu_{\theta}^*, \mu^*) \approx 
\inf_{\gamma \in \Gamma(\mu^*, \mu_{\theta}^*)} \int \| \vu - \vv \|^2 \, d \gamma(\vu, \vv) = \mathcal{W}_2 (\mu, \mu_{\theta}^*).
\end{equation}
Thus one can argue that minimizing the Pfwd objective also induces a minimization of the discrete Wasserstein-2 metric between the two invariant measures.

\section{Maximum Mean Discrepancy}\label{appsec:mmd}
In this section, we provide additional information and context about the Maximum Mean Discrepancy (MMD).
The MMD is an instance of an integral probability metric (IPM; \citet{muller1997integral}, which is a useful construction that allows us to measure distance between distributions.
For any two distributions $\mu$ and $\nu,$ IPMs are defined with a function class $\gG$ as:
\begin{align}\label{eq:ipm}
    \ipm(\mu, \nu) = \sup_{g \in \gG}~ \Bigl\lvert\E_{\vu \sim \mu}[g(\vu)] - \E_{\vu \sim \nu}[g(\vu)]\Bigr\lvert.
\end{align}
Given that we seek our model $\gS_\theta$ to preserve $\mu^*,$ we can use an IPM as the distance $\D$ in \Eqref{eq:obj_dist_reg}, since, for a rich enough function class $\gG,$ $\ipm(\mu^*, \gS_{\theta\#}\mu^*) \rightarrow 0$ implies $\gS_{\theta\#}\mu^* \rightarrow \mu^*$.

One instance of an IPM is when $\gG$ is the space of functions with bounded norm in a reproducing kernel Hilbert space $\gH_\kappa$, i.e., $\gG = \{g: ||g||_{\gH_\kappa} \leq 1\},$ in which case, \Eqref{eq:ipm} coincides with the Maximum Mean Discrepancy (MMD) \citep{gretton2012kernel, sriperumbudur2009integral}, where $\kappa$ is the reproduced kernel.
Using the reproducing property of $\gH_\kappa$ and the Riesz representation theorem, we have that the MMD can be expressed as follows:
\begin{align}\label{eq:mmd_rkhs}
    \mmd^2 = ||\E_{\mu^*}[\kappa(\vu, \cdot)] - \E_{\nu}[\kappa(\vv, \cdot)]||^2_{\gH_\kappa},
\end{align}
where $\E_\mu[\kappa(\vu, \cdot)]$ is the mean embedding of $\mu$ \citep{gretton2012kernel}.
Applying the reproducing property of $\gH_\kappa$ again allows us to equivalently write \Eqref{eq:mmd_rkhs} as in \Eqref{eq:mmd} \citep{gretton2012kernel}.

As described in \Cref{sec:exp}, we use a rational quadratic kernel.
While other works that use MMD for distribution matching, such as \citet{li2015generative} and \citet{dziugaite2015training}, also explored the squared exponential kernel, $\kappa_\sigma(\vu, \vv) = \exp(\frac{-1}{2\sigma}||\vu - \vv||_2^2)$, they found that careful tuning of the bandwidth parameter was required.
In contrast, other than the highest dimension Kolmogorov flow experiments, we found that the mixture of bandwidths used in our rational kernel was comparatively robust and did not require a comprehensive hyperparameter search.
We therefore rely on this kernel and do not explore the more sensitive squared exponential kernel.

\section{Evaluation Criteria}\label{appsec:eval}
In this section, we provide further detail about the evaluation criteria used in \Cref{sec:exp}.
\subsection{Cosine Similarity}\label{appsubsec:cosine}
Letting $\{\vu^{(i)}_{t_k}\}_{i=1}^n$ and $\{\tilde{\vu}^{(i)}_{t_k}\}_{i=1}^n$ be the ground truth and predicted states (respectively) at time $t_k$, for $k = 1,..., N$, across test set trajectories, the cosine similarity at each time step is defined as:
\begin{align*}
 \mathrm{avg.~cosine~sim}(t_k) = \frac{1}{n}\sum_{i=1}^n\frac{(\vu^{(i)}_{t_k} - \bar{\vu}_{t_k})^\top(\tilde{\vu}^{(i)}_{t_k} - \bar{\vu}_{t_k})}{||(\vu^{(i)}_{t_k} - \bar{\vu}_{t_k})||\cdot||(\tilde{\vu}^{(i)}_{t_k} - \bar{\vu}_{t_k})||},
\end{align*}
where $\bar{\vu}_{t_k} = \frac{1}{n}\sum_{i=1}^n\vu_{t_k}$ is the mean of the ground truth trajectories at each time step.
Here $t_k = k\cdot\Delta t$ refers to number of discrete time steps multiplied by the time resolution of the trajectories. Intuitively this metric provides the angle between the different trajectories, i.e., it measures if the snapshots are ``pointing'' in the same direction.

\subsection{Sinkhorn Divergence}\label{appsubsec:ot}

Popular metrics used to measure distance between distributions include Optimal Transport (OT) based metrics, such as the Sinkhorn divergence, which we describe below.
The field of OT is concerned with transforming (or transporting) one distribution into another, i.e., finding a map between them, in an optimal manner with respect to a pre-defined cost. 
The cost of the minimal (or optimal) transformation, often called the cost of the OT map, can then be used to define distances between distributions that `lifts' the underlying metric $\mathrm{d}$ defined on $\gU$ to one over the space of probability measures $\gP(\gU)$ \citep{santambrogio2015optimal}.

In this context, we define the Kantorovich formulation of the OT cost \citep{kantorovich1942translocation} as
\begin{align*}
    \gW(\mu, \nu) = \min_{\gamma \in \Gamma(\mu, \nu)}\int_{\gU \times \gU} c(\vu, \vv)d\gamma(\vu, \vv),
\end{align*}
where $c: \gU \times \gU \rightarrow \mathbb{R}^+$ is an arbitrary cost function for transporting a unit of mass from $\vu$ to $\vv,$ and $\Gamma$ is the set of joint distributions defined on $\gU \times \gU$ with correct marginals, i.e.,
\begin{align*}
    \Gamma(\mu, \nu) = \{\gamma \in \gP(\gU \times \gU) \mid P_{1\#}\gamma = \mu, P_{2\#}\gamma = \nu\},
\end{align*} 
with $P_1(\vu, \vv) = \vu$ and $P_2(\vu, \vv) = \vv$ being simple projection operators.
When $c(\vu, \vv) = \mathrm{d}(\vu, \vv)^p$ with $p \geq 1$, then $\gW^{1/p}$ is known as a Wasserstein-$p$ distance.

Practically, finding OT maps is a computationally expensive procedure. We therefore use entropic regularized versions of OT costs, which are amenable to efficient implementation on computational accelerators, by means of the Sinkhorn algorithm \citep{cuturi2013sinkhorn, peyre2019computational}:
\begin{align}\label{eq:ot_reg}
    \gW_\varepsilon(\mu, \nu) = \min_{\gamma \in \Gamma(\mu, \nu)} \gW + \mathrm{KL(\gamma || \mu \otimes \nu)},
\end{align}
where $\mathrm{KL}$ is the Kullback-Leibler divergence, and $\mu \otimes \nu$ is the product of the marginal distributions.
This gives rise to the Sinkhorn Divergence (SD):
\begin{align*}
    \sd(\mu, \nu) = 2\gW_\varepsilon(\mu, \nu) - \gW_\varepsilon(\mu, \mu) - \gW_\varepsilon(\nu, \nu),
\end{align*}
which alleviates the entropic bias present in \Eqref{eq:ot_reg}, i.e. $\gW_\varepsilon(\mu, \mu) \neq 0.$
Of note, the SD can be shown to interpolate between a pure OT cost $\gW$ (as $\varepsilon \rightarrow 0$) and a MMD (as $\varepsilon \rightarrow 
\infty$) \citep{ramachandran18, genevay2018learning, feydy2019interpolating}.

In \Cref{sec:exp}, the SD was used to compare empirical version of the ground truth and predicted distributions of trajectories.
We use the Optimal Transport Tools library \citep{cuturi2022optimal} with its default hyperparameters to perform this computation.
We also explored using the Sinkhorn Divergence as the measure distance in \Eqref{eq:obj_dist_reg_full}.
However, especially in higher dimension experiments, we found this divergence to be less informative in guiding training, likely owing to its less favorable estimation properties compared to the MMD, particularly in the high-dimensional regime, see \Cref{app:SD_reg} for more details.

\subsection{Radially Averaged Energy Spectrum}
The energy spectrum is one of the main metrics to quantitatively assess generated samples \citep{wan2023debias}.
In a nutshell, the energy spectrum measures the energy in each Fourier mode, thereby providing insights into the similarity between the generated and reference samples.  

The energy spectrum is defined\footnote{This definition is applied to each sample and averaged to obtain the metric (same for MELR).} as 
\begin{equation}
\label{eq:energy_spectrum}
    E(K) = \sum_{|\underline{K}| =  K} | \hat{\vu}(\underline{K}) | ^2 = \sum_{|\underline{K}| = K} \left | \sum_{i,j} \vu(x_{i,j}) \exp(-j 2\pi \underline{K} \cdot x_{i,j}/L) \right|^2
\end{equation}
where $\vu$ is a snapshot system state, $K$ is the magnitude of the wave-number (wave-vector in 2D) $\underline{K}$, and $x_{i,j}$ is the underlying (possibly 2D) spatial grid. 
To assess the overall consistency of the spectrum between the generated and reference samples using a single scalar measure, we consider the mean energy log ratio (MELR):
\begin{equation}
\label{eq:MELR}
    \text{MELR} = \sum_K w_K\left |\log \left (E_{\text{pred}}(K) /E_{\text{ref}}(K) \right )\right |,
\end{equation}
where $w_K$ represents the weight assigned to each $K$.
We further define $w_{K}^{\text{unweighted}} = 1/\text{card}(K)$ and $w_{K}^{\text{weighted}} = E_{\text{ref}}(K)/\sum_K E_\text{ref}(K)$.
The latter skews more towards high-energy/low-frequency modes. 

\subsection{Covariance RMSE (covRMSE)} \label{app:covariance_rmse}
The covariance root mean squared error quantifies the difference in the long-term spatial correlation structure between the prediction and the reference. It involves first computing the (empirical) covariance on a long rollout:
\begin{equation}
\label{eq:empirical_covariance}
    \text{Cov}(\vu) = \frac{1}{N\cdot n} \sum_{i=1}^n \sum_{k=1}^N (\vu^{(i)}_{t_k} - \bar{\vu})(\vu^{(i)}_{t_k} - \bar{\vu})^T, \quad \bar{\vu} = \frac{1}{N \cdot n} \sum_{i=1}^n \sum_{k=1}^N \vu^{(i)}_{t_k},
\end{equation}
where $\vu^{(i)}_{t_k}$ are realizations of the multi-dimensional random variable $U$ (in this case, they are just the snapshots of the trajectory $i$ at time steps $t_k$.)
For 2D Kolmogorov flow, we leverage the translation invariance in the system to compute the covariance on slices with fixed $x$-coordinate. The error is then given by:
\begin{equation}
\label{eq:RMSE}
    \text{covRMSE} = \frac{\|\text{Cov}_{\text{pred}} - \text{Cov}_{\text{ref}}\|}{\|\text{Cov}_{\text{ref}}\|},
\end{equation}
where $\|\cdot\|$ is taken to be the Frobenious norm.

\subsection{Time Correlation Metric (TCM)}
The quantities introduced above, such as the energy spectrum, are single-time quantities. Compared to single-time quantities, examining multiple-time statistics can provide a better view of more complex temporal behavior. 


We leverage the spatial homogeneity and compute pointwise statistics for a scalar time series $u$, then average over space.  Assuming stationarity, one definition of the autocorrelation function is $\rho(t) = C(t) / C(0)$, where
    \begin{equation}
        C(t_i) = \frac{1}{N} \sum_{k=1}^N (u_{t_k} - \bar{u}) (u_{t_{k-i}} - \bar{u}), \quad \bar{u} = \frac{1}{N} \sum_{k=1}^N u_{t_k}
    \end{equation}
    
The \emph{autocorrelation time} $\tau$, which is defined as
    \begin{equation}
        \tau = \Delta t \left( 1 + 2 \sum_{i=1}^\infty \rho(t_i) \right),
    \end{equation}
can be interpreted as the time for the signal to forget its past. We compute the average pixel-wise $\tau$ for ground truth as well as prediction rollouts and take their absolute difference to form a metric.

\section{Regularization}\label{appsec:regularization}
For better reproducibility of our work, we provide explicit formulas for the regularized objective functions.
We reproduce \Eqref{eq:obj_dist_reg_full} from \Cref{sec:exp}
\begin{align*}
    \widehat{\gL}_\reg^{\D}(\theta) = \widehat{\gL}^{\text{obj}}(\theta) +  \reg_1 \widehat{\D}(\mu^*, (\gS_\theta^\ell)_{\sharp}\mu^{*}) + \reg_2 \widehat{\D}((\gS^\ell)_{\sharp}\mu^*, (\gS_\theta^\ell)_{\sharp}\mu^{*}). 
\end{align*}
For each type of objective the training schedule is slightly different, namely:
\begin{itemize}
    \item When $\widehat{\gL}^{\text{obj}}(\theta)$ corresponds to the 1-step objective, $\widehat{\gL}^{\text{1-step}}(\theta),$ then $\ell=1.$
    \item When $\widehat{\gL}^{\text{obj}}(\theta)$ corresponds to the Curr objective, $\widehat{\gL}^{\text{Curr}}(\theta),$ then we gradually increase $\ell$ from 1 to some maximum rollout value according to a schedule determined by the number of training steps, as described in \Cref{appsubsec:lorenz63}, \Cref{appsubsec:ks}, and \Cref{appsubsec:ns}, below.
    \item When $\widehat{\gL}^{\text{obj}}(\theta)$ corresponds to the Pfwd objective, we use the same schedule as in Curriculum training, but randomly sample the rollout length up to $\ell$ for each batch, following the implementation provided by \citet{brandstetter2022message}\footnote{See \href{https://github.com/brandstetter-johannes/MP-Neural-PDE-Solvers}{\texttt{https://github.com/brandstetter-johannes/MP-Neural-PDE-Solvers}} for more details.}.
\end{itemize}

For the Curr objectives, we have the following formulas for the regularization terms.
Suppose that $\{\vu^{(i)}\}_{i=1}^n \sim \mu^*$ is a mini-batch of size $n$ sampled from the invariant measure, then using the sample-based MMD estimator, the estimate of the  term $\widehat{\D}(\mu^*, (\gS_\theta^\ell)_{\sharp}\mu^{*})$ in \Eqref{eq:obj_dist_reg_full}, i.e., the unconditional regularization term, can be written as 
\begin{align} \label{eq:app_mmd_estimate_conditional_curr}
    \widehat{\mmd}^2( \mu^*, ( \gS_{\theta}^k)_{\sharp} \mu^*) =
    \frac{1}{n^2}\sum_{i, j} \kappa(\vu^{(i)}, \vu^{(j)}) 
    + \frac{1}{n^2}\sum_{i, j} \kappa(\gS_{\theta}^k(\vu^{(i)}), \gS_{\theta}^k(\vu^{(j)})) 
    - \frac{2}{n^2}\sum_{i, j} \kappa(\vu^{(i)}, \gS_{\theta}^k(\vu^{(j)})).
\end{align}
The last term in \Eqref{eq:obj_dist_reg_full}, i.e., the conditional regularization term given by $\widehat{\D}((\gS^\ell)_{\sharp}\mu^*, (\gS_\theta^\ell)_{\sharp}\mu^{*})$, can be written as 
\begin{align} \label{eq:mmd_estimate_conditional_curr}
\begin{split}
    \widehat{\mmd}^2((\gS^k)_{\sharp} \mu^*, ( \gS_{\theta}^k)_{\sharp} \mu^*) &=
    \frac{1}{n^2}\sum_{i, j} \kappa(\gS^k(\vu^{(i)}), \gS^k(\vu^{(j)}))
    + \frac{1}{n^2}\sum_{i, j} \kappa(\gS_{\theta}^k(\vu^{(i)}), \gS_{\theta}^k(\vu^{(j)})) \\
    &- \frac{2}{n^2}\sum_{i, j} \kappa(\gS^k(\vu^{(i)}), \gS_{\theta}^k(\vu^{(j)})).
\end{split}
\end{align}

Similar formulas are also presented for the Pfwd objectives, although they introduce a stop gradient in the second to last unrolling step, namely
\begin{align} \label{eq:app_mmd_estimate_conditional_pfwd}
\begin{split}
    \widehat{\mmd}^2( \mu^*, ( \gS_{\theta}^k)_{\sharp} \mu^*)&=
    \frac{1}{n^2}\sum_{i, j} \kappa(\vu^{(i)}, \vu^{(j)}) 
    + \frac{1}{n^2}\sum_{i, j} \kappa(\gS_{\theta}(\sg (\gS_{\theta}^{k-1}(\vu^{(i)})), \gS_{\theta}( \sg (\gS_{\theta}^{k-1}(\vu^{(j)}))) \\
    & - \frac{2}{n^2}\sum_{i, j} \kappa(\vu^{(i)}, \gS_{\theta}( \sg (\gS_{\theta}^{k-1}(\vu^{(j)}))), 
\end{split}
\end{align}
and
\begin{align} \label{eq:mmd_estimate_conditional_pfwd}
\begin{split}
    \widehat{\mmd}^2((\gS^k)_{\sharp} \mu^*, ( \gS_{\theta}^k)_{\sharp} \mu^*) &=
    \frac{1}{n^2}\sum_{i, j} \kappa(\gS^k(\vu^{(i)}), \gS^k(\vu^{(j)}))
    + \frac{1}{n^2}\sum_{i, j} \kappa(\gS_{\theta}(\sg (\gS_{\theta}^{k-1}(\vu^{(i)}))),\gS_{\theta}( \sg (\gS_{\theta}^{k-1}(\vu^{(j)})))) \\
    &- \frac{2}{n^2}\sum_{i, j} \kappa(\gS^k(\vu^{(i)}), \gS_{\theta}( \sg (\gS_{\theta}^{k-1}(\vu^{(j)})))).
\end{split}
\end{align}

\section{Experimental Setup}\label{appsec:exp_setup}
Below, we provide information about each dynamical system from \Secref{sec:exp} and their corresponding experimental setup.
In Table \ref{tab:exp_hyperparam}, we give an overview of the model, learning rate, and number of training steps used in each experiment.
\begin{table}[h!]
    \centering
    \caption{Model, learning rate, and number of training steps for each experiment in \Secref{sec:exp}.}
    \begin{tabular}{lccc}
    \toprule
        System  & $\gS_\theta$ & LR & Training steps\\
        \midrule
         Lorenz 63 & MLP w/residual connection to input & $1\mathrm{e}^{-4}$ & 500k\\
         Kuramoto–Sivashinsky & Dilated convolutional network \cite{stachenfeld_learned_2022} &$5\mathrm{e}^{-4}$ & 300k \\
         Kolmogorov Flow & Dilated convolutional network \cite{stachenfeld_learned_2022} & $5\mathrm{e}^{-4}$ & 720k\\
         \bottomrule
    \end{tabular}
    \label{tab:exp_hyperparam}
\end{table}

\subsection{Lorenz 63}\label{appsubsec:lorenz63}
The Lorenz 63 model \citep{lorenz1963deterministic} is defined on a 3-dimensional state space by the following non-linear ordinary differential equation $\dot{\vu} = f(\vu)$:
\begin{align}\label{eq:lorenz}
    \begin{split}
        \dot{x} &= \sigma(y - x)\\
        \dot{y} &= \rho x - y - xz \\
        \dot{z} &= xy - \beta z
    \end{split}
\end{align}
The Lorenz 63 system is typically associated to parameter values of $\sigma = 10, \rho = 28,$ and $\beta = 8/3$ and is known to be chaotic with an attractor that supports an ergodic measure \citep{tucker2002rigorous, luzzatto2005lorenz}.

Training and evaluation data were generated using a 4$^{\text{th}}$ order Runge-Kutta numerical integrator \citep{runge1895numerische,kutta1901beitrag} with time scale $\Delta t = 0.001.$
We first selected random initial conditions. Trajectories were then rolled out for 100,000 warm-up steps to ensure that points were sampled from the the invariant measure supported on the Lorenz attractor.
These warm-up steps were subsequently discarded.
Starting from initial conditions sampled from $\mu^*$, we generate 5,000 training trajectories each of length 100,000 steps and 20,000 test trajectories of length 1,000,000 steps.
At training and evaluation time these trajectories are down-sampled along the temporal dimension by a factor of 400, so that the effective time scale was $\Delta t = 0.4$.
Data were normalized to have roughly zero mean and unit variance based on statistics of the training set
During training we randomly sample batches of size 2,048 that consist of 10 step windows in the training trajectories.

We define $\gS_\theta$ as a one-step finite difference model: $\tilde{\vu}_{k+1} = \gS_\theta(\vu_{k}) = \vu_{k} + \Delta t f_\theta(\vu_k),$ where $f_\theta$ is a parametric model of the continuous time dynamics.
We parameterize $f_\theta$ by a multi-layer perceptron (MLP) with two hidden layers each of dimension 32 and use the ReLU activation function.
We trained with an \textsc{ADAM} optimizer \citep{kingma2014adam} with learning rate $1\mathrm{e}^{-4}.$

Models were trained for 500,000 steps.
For the curriculum training (and its regularized counterpart), we increase $\ell$ by one every 50,000 training steps, and hence by the end of training $\gS_\theta$ is predicting trajectories of length 10.
For curriculum training, we weight rollout loss using a geometric weighting $\omega(k) = \max(0.1^{k-1}, 1\mathrm{e}^{-7}).$ 
For pushforward training we use the same rollout schedule as in curriculum training, but the rollout loss weight is $\omega(k) = \max(0.1^{k-1}, 1\mathrm{e}^{-4}).$
These weighting schemes were chosen empirically to ensure that training loss was of the same order of magnitude throughout training, even as rollout length increased.
The MMD bandwidth values used were $\boldsymbol{\sigma} = \{0.2, 0.5, 0.9, 1.3\}$.

\subsection{Kuramoto–Sivashinsky}\label{appsubsec:ks}
The non-linear PDE known as the Kuramoto–Sivashinsky equation (KS) \citep{kuramoto1978diffusion, ashinsky1988nonlinear}, has the following form:
\begin{equation} \label{eq:app_ks_pde}
    \partial _t \vu + u \partial_x \vu + \nu \partial_{xx} \vu  -  \nu \partial_{xxxx} \vu = 0 \qquad  \text{in } [0, L] \times \mathbb{R}^+, 
\end{equation}
with periodic boundary conditions, and $L=64$.
Here the domain is re-scaled in order to balance the diffusion and anti-diffusion components so the solutions are chaotic \citep{dresdner2022learning}.

The KS system is known to be chaotic \citep{papageorgiou1991route} and, when stochastically forced, ergodic with an invariant measure \citep{weinan2002gibbsian,ferrario2008invariant}.
We generate data for this system using a spectral solver \citep{dresdner2022learning} on a spatial grid $[0, 64]$ with $512$ equally-spaced points and
a 4th-order implicit-explicit Crack-Nicolson Runge-Kutta scheme~\citep{Canuto:2007_spectral_methods}, with a time resolution of $\Delta t = 0.001$.
For each trajectory, we start with a randomly generated initial condition given by
\begin{equation}
\label{eq:ic_distribution}
    \vu_0(x) =  \sum_{j = 1}^{n_c} a_j \sin(\omega_j * x + \phi_j),
\end{equation}
where $\omega_j$ is chosen randomly from $\{2\pi/L, 4\pi/L, 6\pi/L\}$, $a_j$ is sampled from a uniform distribution on $[-0.5, 0.5]$, and phase $\phi_j$ follows a uniform distribution on $[0, 2\pi]$.
We use $n_c = 30$.
We let the system ``warm up'' for 20 units of time, before recording the trajectories.
The training dataset consists of 800 trajectories of 1,200 steps with a time sampling rate $\Delta t = 0.2$ time units, from which we randomly sample batches of size 128 and trajectory length of 10 steps.
Our evaluation set consists of 100 trajectories of length 1,000 steps.

We parameterize $\gS_\theta$ as a dilated convolution neural network with residual connections, as described in \citep{stachenfeld_learned_2022}. In contrast to the Lorenz 63 model, we do not involve the time step $\Delta t$ directly in the computation of the update, instead we use $\tilde{\vu}_{k+1} = \gS_\theta(\vu_k)$. 
The architecture consists of an encoder convolutional module, four dilated convolutional blocks, and a decoder convolutional module.
There exists a residual connection from the encoder to the output of the first dilated convolution block and from the input to the decoder output.
The intermediate representations have $48$ channels.
The encoder, decoder, and intermediate dilated convolutions use kernels of width 5.
Within each dilated convolution block, there are four dilated convolutional layers followed by ReLU non-linear activations, with dilation factor increasing by a multiple of 2 for each layer.
Each block has a residual connection to the previous one.
The model has a total of 324,433 parameters.
The model was trained using the \textsc{ADAM} optimizer and an initial learning rate of $5\mathrm{e}^{-4}$.
A staircase exponential decay learning rate scheduler was used with a decay factor of 0.5 and decay transitions every 60,000 steps.

Models were trained for 300,000 steps with the rollout increased every 60,000 steps, and hence by the end of training $\gS_\theta$ is predicting trajectories of length 5.
For both the Curr and Pfwd objectives we use the same rollout schedule and rollout loss weight: $\omega(k) = \max(0.9^{k-1}, 1\mathrm{e}^{-3})$.
The MMD bandwidth values used were $\boldsymbol{\sigma} = \{0.2, 0.5, 0.9, 1.3\}$.

\subsection{Kolmogorov Flow}\label{appsubsec:ns}

We also consider the Navier-Stokes equation with Kolmogorov forcing given by 
\begin{gather}
\label{eq:NS_equation}
    \frac{\partial \vu}{ \partial t} = - \nabla \cdot (\vu \otimes \vu) + \nu \nabla^2 - \frac{1}{\rho} \nabla p + \mathbf{f} \qquad \text{in } \Omega, \\ 
    \nabla \cdot \vu = 0  \qquad \text{in } \Omega,
\end{gather}
where $\Omega = [0, 2\pi]^2$, $\vu(x,y) = (\vu_x, \vu_y)$ is the field, $\rho$ is the density, $p$ is the pressure, and $\mathbf{f}$ is the forcing term given by 
\begin{equation}
    \mathbf{f} = \left( \begin{array}{c}
        0 \\
        \sin(k_0 y)
        \end{array} 
    \right) + 0.1 \vu,
\end{equation}
where $k_0 = 4$. The forcing only acts in the $y$ coordinate. Following \citet{kochkov_machine_2021}, we add a small drag term to dissipate energy. 
An equivalent problem is given by its vorticity formulation 
\begin{equation}\label{eq:vorticity_NS}
\partial_t\omega = - \vu\cdot\nabla \omega + \nu \nabla^2 \omega  - \alpha\,\omega + f,
\end{equation}
where \mbox{$\omega := \partial_x \vu_y - \partial_y \vu_x$}, which we use for spectral method which avoids the need to separately enforce the incompressibility condition~\mbox{$\nabla \cdot \vv = 0$}.
The initial conditions are the same as the ones proposed in \citet{kochkov_machine_2021}.

\paragraph{Pseudo-Spectral Discretization}
Equations \ref{eq:app_ks_pde} and \ref{eq:vorticity_NS} were discretized using a pseudo-spectral discretization, to avoid issues stemming from dispersion errors.
Pseudo-spectral methods are known to be dispersion free, due to the \textit{exact} evaluation of the derivatives in Fourier space, while possessing excellent approximation guarantees \citep{trefethen_spectral_2000}. We use the \texttt{jax-cfd} spectral elements tool box \citep{dresdner2022learning}.
Learning to
correct spectral methods for simulating turbulent flows, which leverages the Fast Fourier Transform \citep{Cooley_Tukey:1965} to compute the Fourier transform in space of the field $\vu(x, t)$, denoted by $\hat{\vu}(t)$, allows for a very efficient computation of spatial derivatives by diagonal rescaling following $\partial_x \hat{\vu}_K = i K \hat{\vu}_K$, where $K$ is the wave number.
This renders the application and inversion of linear differential operators trivial, since they are simply element-wise operations~\citep{trefethen_spectral_2000}.

This procedure transforms \Eqref{eq:app_ks_pde} and \Eqref{eq:NS_equation} to a system in Fourier domain of the form
\begin{equation}\label{eq:spectral_time_evol}
\partial_t \hat{\vu}(t) = \mathbf{D} \hat{\vu}(t) + \mathbf{N}(\hat{\vu}(t)),
\end{equation}
where $\mathbf{D}$ denotes the linear differential operators in the Fourier domain and is often a diagonal matrix whose entries only depend on the wave number $K$ and $\mathbf{N}$ denotes the nonlinear part.
These non-linear terms are computed in real space.

\Eqref{eq:vorticity_NS} was discretized with spatial discretization $n_x = n_y = 256$ and a 4th order implicit-explicit Crack-Nicolson Runge-Kutta scheme~\citep{Canuto:2007_spectral_methods}, where we treat the linear part implicitly and the nonlinear one explicitly with $\Delta t = 0.001$ using \texttt{jax-cfd}. 

For each trajectory, we let the solver ``warm up'' for 50 units of time, in order for the trajectory to reach the attractor. We further evolve the equation for 120 units of time, and we sample the trajectories at a rate of $\Delta t = 0.1$.
Finally, we downsample the trajectories by a factor 4 in each spatial direction. 
We repeated the process 128 times to obtain the training data, and 32 times for both the validation and test data.

\subsubsection{Kolmogorov flow experiments for each objective.}
We set the learning rate to be $5\mathrm{e}^{-4}$, and we vary the batch size from 32 to 512 (depending of the experiment) in increments of power to two.
We use an exponential learning rate scheduler, which halves the learning rate every 72,000 iterations.
We trained the models for up to 720,000 iterations.
Unless otherwise stated the MMD bandwidth used was $\boldsymbol{\sigma} = \{2, 5, 9, 13, 20, 50, 90, 120\}$.
This value was found after a quick hyperparameter tuning on a small dataset.

We parametrize $\gS_{\theta}$ using a two-dimensional dilated convolutional neural network with residual connections and periodic boundary conditions following \citep{stachenfeld_learned_2022} the total number of parameters is 6,458,785.
We follow the same unrolling scheme as in the KS system, i.e., $\tilde{\vu}_{k+1} = \gS(\vu_k)$.

\paragraph{One-step} 
For this objective, we use the same set up as experiments above. We halved the learning rate every 72,000 iterations and the models were trained for 720,000 iterations while keeping $\ell$ constant and equal to one.

\paragraph{Pushforward}
For Pfwd training, we consider a rollout schedule that follows the learning rate schedule: every 72,000 iterations we increase the number of unrolling steps $\ell$ by one, where $\ell$ increases from $1$ to $10$. The effective number of unrolling step at each training step is sampled uniformly from $1$ to $\ell$.

\paragraph{Curriculum} \label{sec:app_curr_experiments}
Given the higher memory requirement, we decrease the number of maximum unrolling steps from $10$ to just $5$.
Also, depending on the batch size, we further decrease the maximum number of unrolling steps.
In particular, for large batch sizes, we cannot afford unrolling more than 2 time steps.
All the other parameters were kept constant relative to Pushfoward experiments.

\begin{figure*}[t]
    \centering
    \includegraphics[scale=0.3]{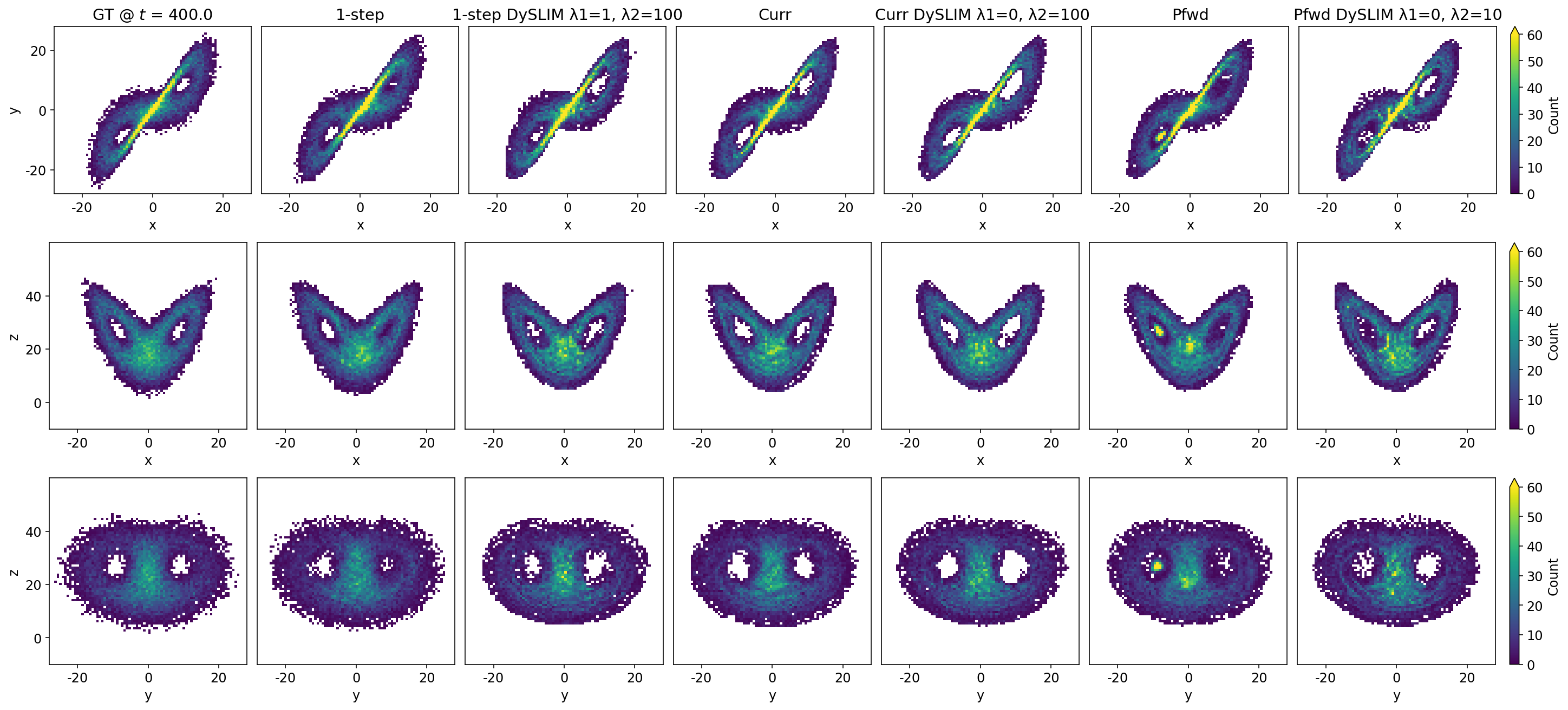}
    \caption{Histograms of trajectories at rollout time $t=400$ for one of the random training seeds. We showcase the well known ``butterfly'' attractor.
    }
    \label{fig:app_l63}
\end{figure*}

\begin{figure*}[t]
    \centering
    \includegraphics[width=0.75\textwidth, trim={0mm 0mm 0mm 0mm}, clip]{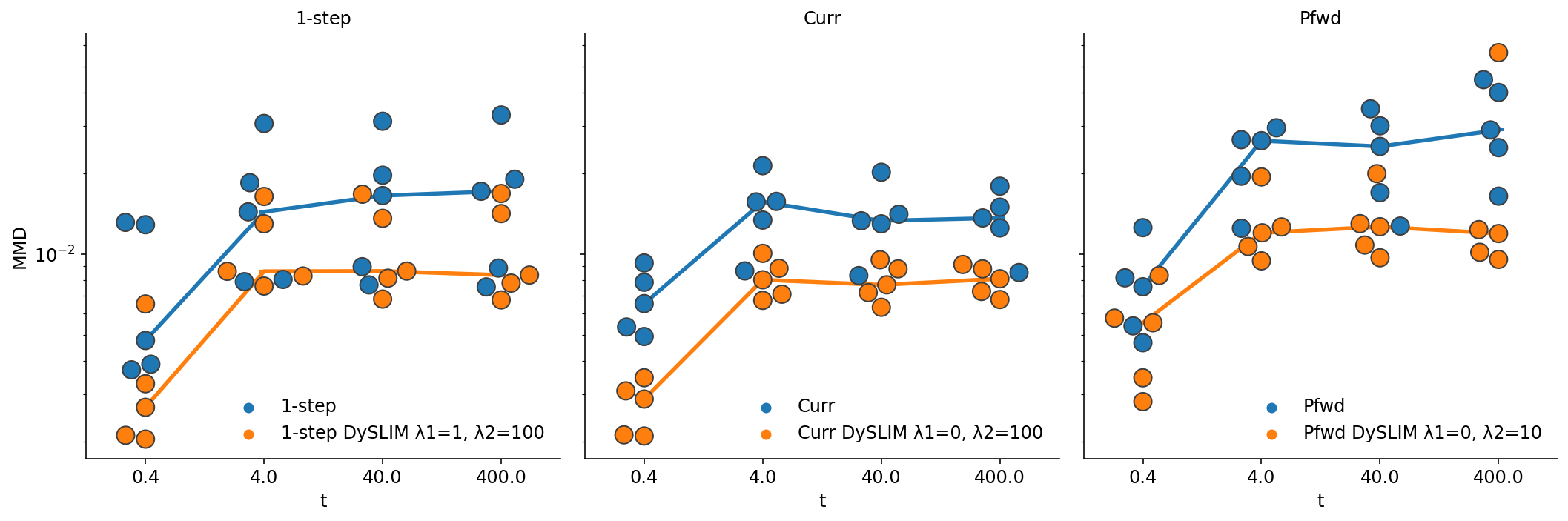}
    \caption{Values of the MMD metric for the baselines and the DySLIM regularization.
    Each point represents a random training seed that remains stable, with the solid line indicating median values.
    }
    \label{fig:lz63_mmd}
\end{figure*}

\begin{figure*}[ht]
    \centering
    \includegraphics[width=0.75\textwidth, trim={0mm 0mm 0mm 0mm}, clip]{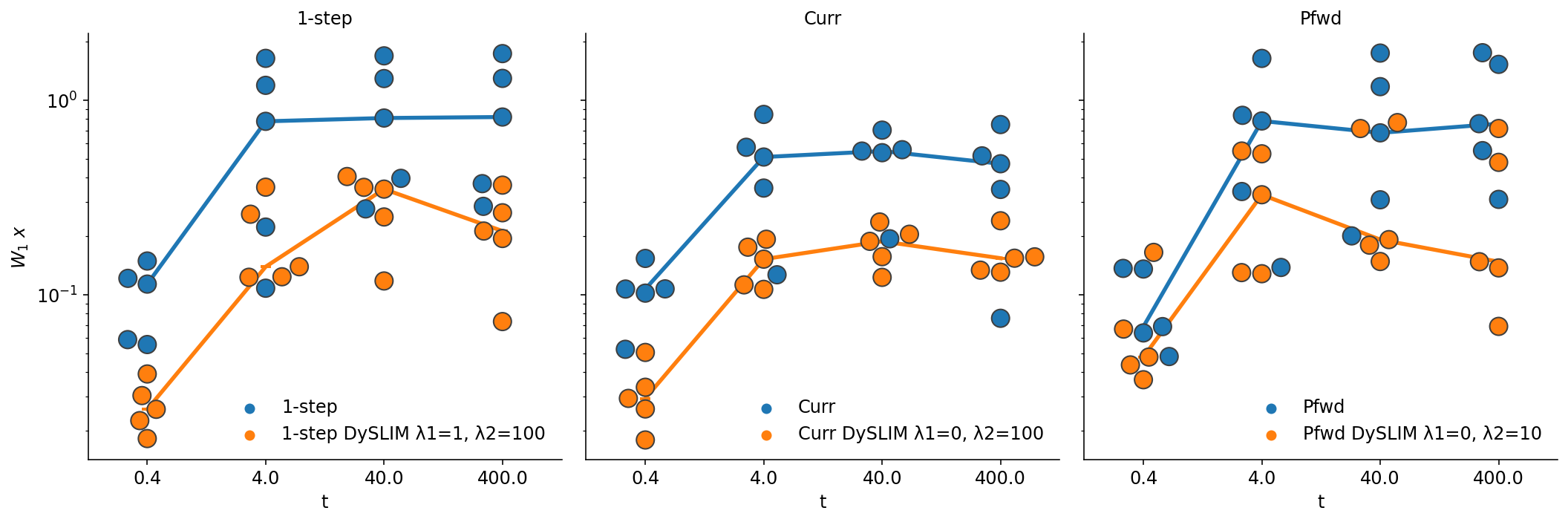}

    \includegraphics[width=0.75\textwidth, trim={0mm 0mm 0mm 0mm}, clip]{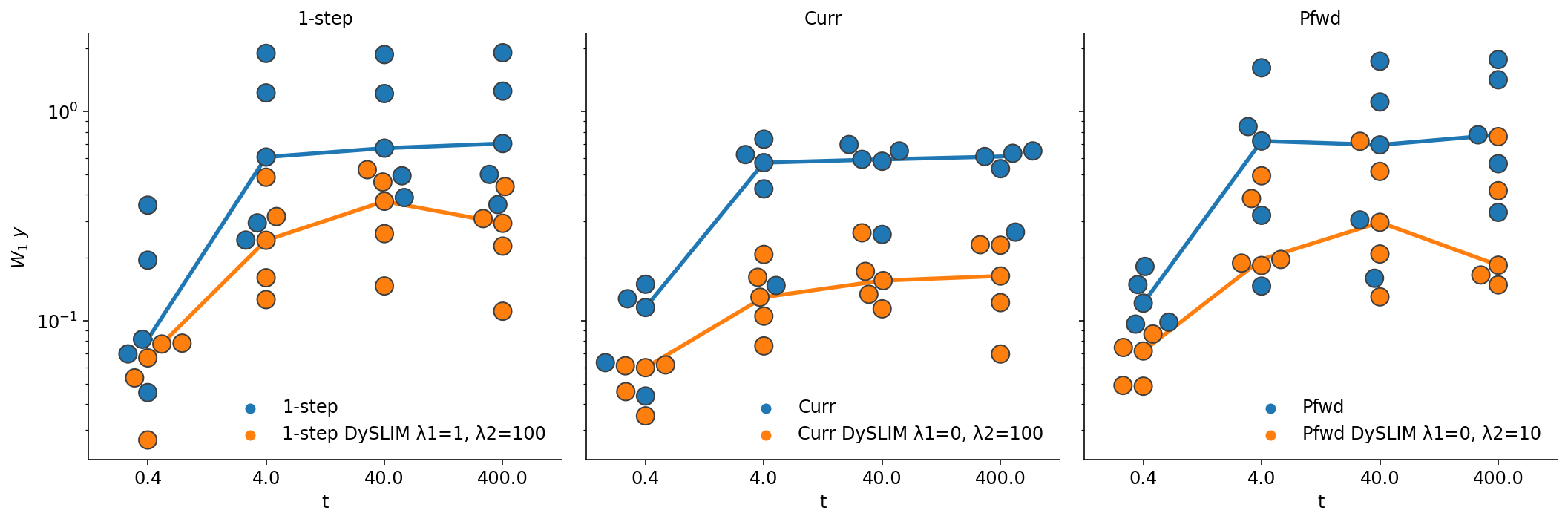}
    
    \includegraphics[width=0.75\textwidth, trim={0mm 0mm 0mm 0mm}, clip]{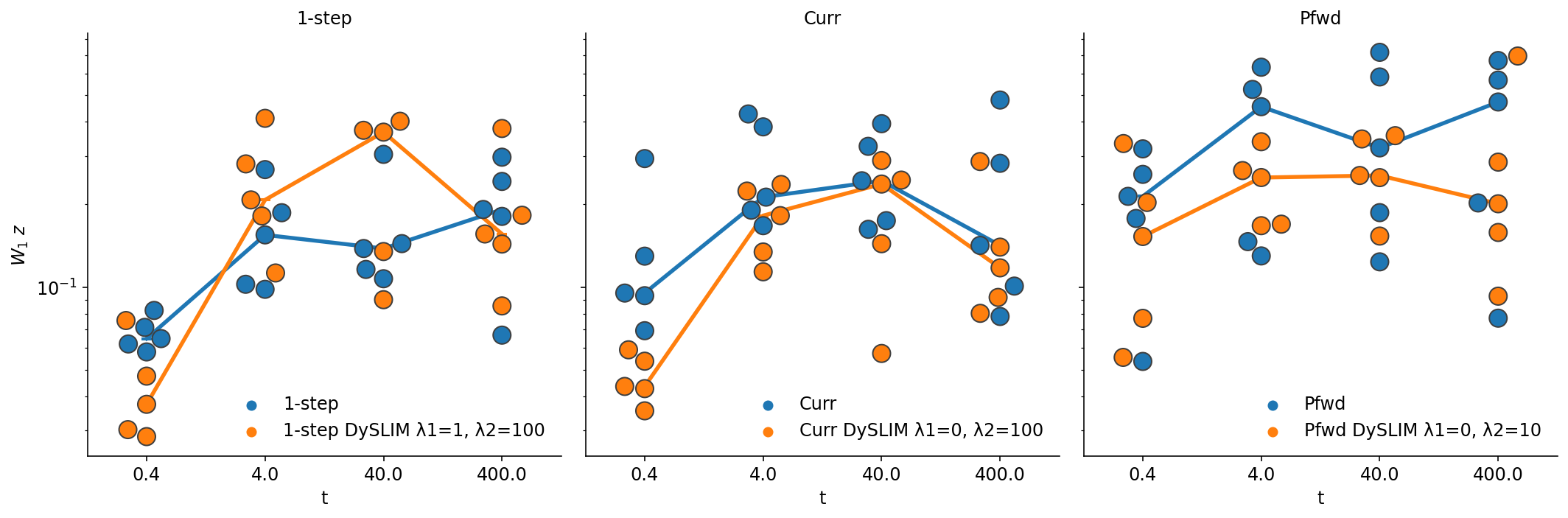}
    
    \includegraphics[width=0.75\textwidth, trim={0mm 0mm 0mm 0mm}, clip]{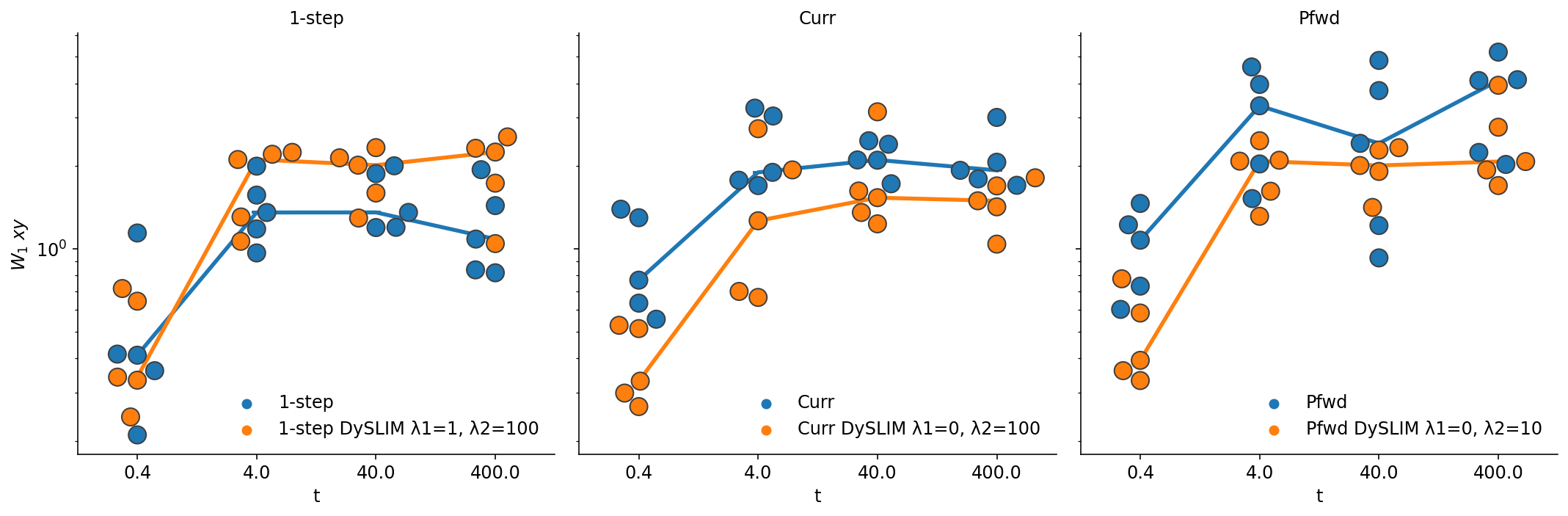}
    
    \includegraphics[width=0.75\textwidth, trim={0mm 0mm 0mm 0mm}, clip]{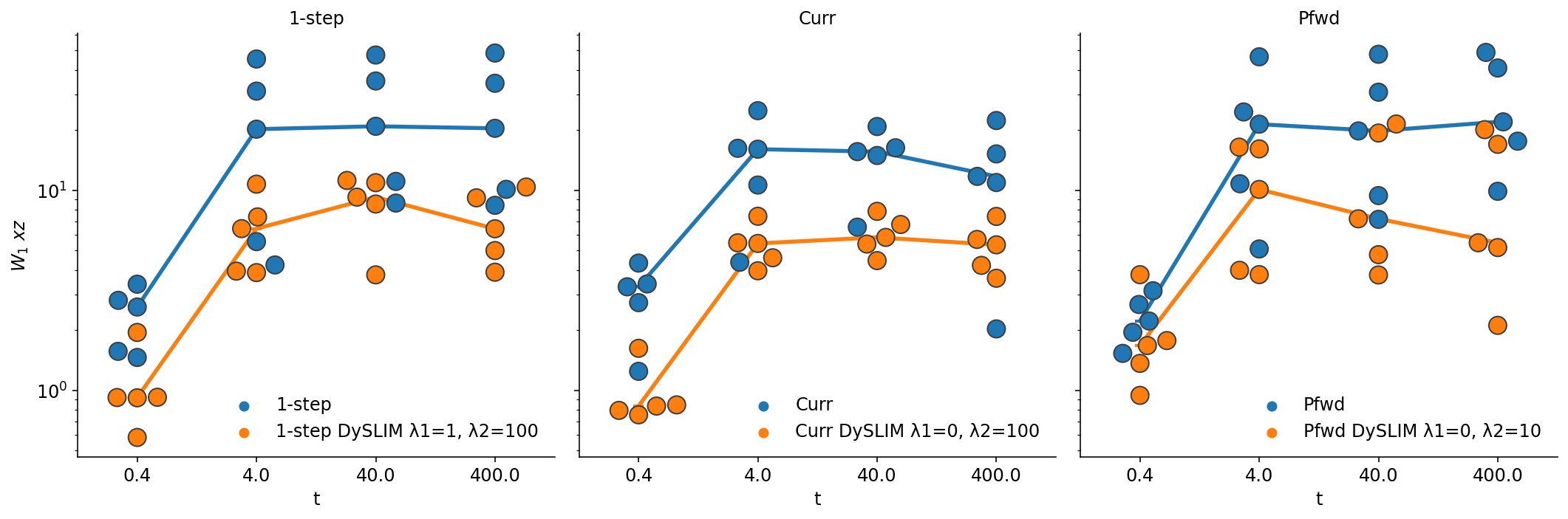}

    \caption{Wasserstein-1 distance ($\downarrow$) between the different components of the forcing in \Eqref{eq:lorenz} at different unrolling times. From top to bottom, $x$, $y$, $z$, and the crossed products $xy$ and $xz$.
    Each point represents a random training seed that remains stable, with the solid line indicating median values.
    }
    \label{fig:lz63_W1_metrics}
\end{figure*}

\section{Further Results}
\subsection{Lorenz 63} \label{sec:app_l64_further_results}
\Figref{fig:app_l63} depicts the attractor for the Lorenz 63 system for an ensemble of trajectories (with randomly chosen initial conditions close to the attractor) at time $t=400$.
We observe that DySLIM regularization provides some extra symmetry to the attractor, as evidenced by a better defined right-wing.

In addition, we provide several other metrics to showcase the advantage of our methodology.
\Figref{fig:lz63_mmd} shows the behavior of the MMD metric for much longer horizons to the ones used for training.
We can observe that DySLIM outperforms all the baselines. 
We also, studied the Wasserstein-1 distance of different features involved in the ground truth dynamics shown in \Eqref{eq:lorenz}.
In this case, \Figref{fig:lz63_W1_metrics} shows that the distribution of each of the components is better captured in the models trained with DySLIM.

\subsection{Kuramoto-Sivashinsky} \label{sec:app_ks_further_resuls}
\begin{figure*}[t]
    \centering
    \includegraphics[width=0.75\textwidth]{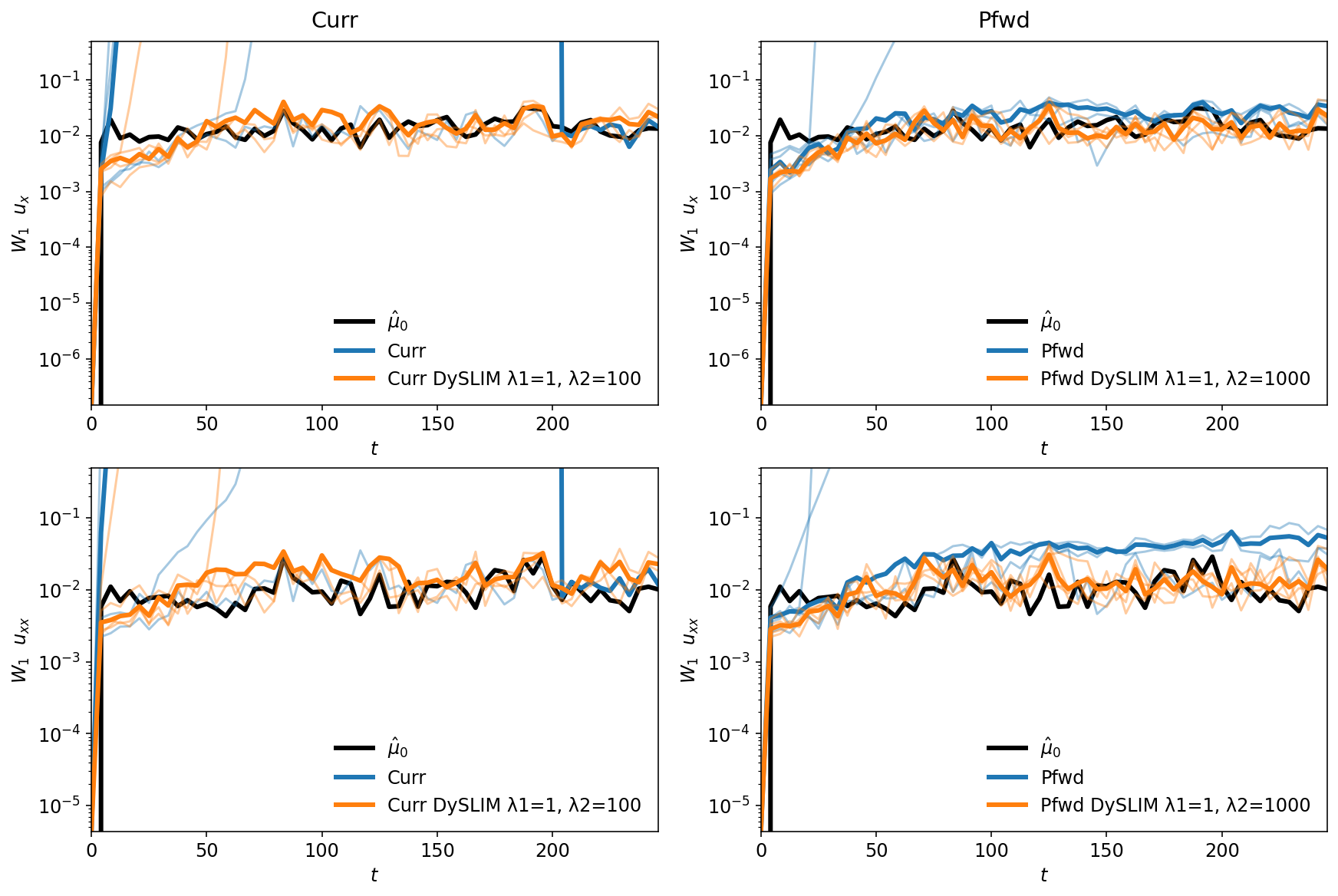}
    \caption{Wasserstein-1 distance ($\downarrow$) on distribution of first order (\textit{Top}) and second order (\textit{Bottom}) spatial derivatives across time.
    Values greater than 3 are not shown on the plot.
    Each line corresponds to one of five random training seeds with bolded lines indicating median values (excluding trajectories that produce \texttt{NaN} values).
    The solid black line corresponds to statistics calculated from the distribution of initial conditions.
    }
    \label{fig:app_ks_W1}
\end{figure*}

In addition to the results shown in the main text, \Figref{fig:app_ks_W1} shows the improved stability from DySLIM by comparing the distribution of first order ($\vu_x$) and second order ($\vu_{xx}$) spatial derivatives for ground truth and predicted trajectories, which are relevant quantities that appear in the PDE that defines this system in \Eqref{eq:app_ks_pde}.
We use finite difference methods to compute $\vu_x$ and $\vu_{xx}$ at each point in time for each trajectory in the test set.
At each point in time, we thus have a distribution for these derivatives across test set trajectories and spatial grid.
We use the Wasserstein-1 distance to compare ground truth and predicted distributions and find that models trained with regularized objectives better match the distribution of ground truth spatial derivatives.
In summary, \Figref{fig:app_ks_W1} shows that by regularizing the loss, we obtain a closer distribution on the derivatives than when using the unregularized loss.
In fact, for some cases of the curriculum training, the Wasserstein-1 matrix explodes as the trajectories are highly unstable.

\subsection{Kolmogorov Flow}
In this section, we provide ablation results and additional trajectory samples.

Table \ref{tab:ns_stats_full} compiles additional results for the model trained with and without regularization following the description in Appendix \ref{sec:app_curr_experiments}.
This table shows that using DySLIM either improves or roughly maintains values for all metrics.
We point out that for curriculum training, the error tends to increase with batch size due to the lower number of rollout steps during training, which is a direct consequence of the higher memory footprint required for curriculum training. 

\begin{table*}[t]
  \centering
  \small
\caption{Metrics for pushforward, curriculum and 1-step baselines with ($\lambda_1 = 0$, $\lambda_2 = 100$) and without regularization under various batch sizes and learning rates for the Kolmogorov flow. Boldface numbers indicate that the metric is improved by our regularizations. The best-performing run shown in Table~\ref{tab:ns_stats} is highlighted in green.}
\label{tab:ns_stats_full}
{
\setlength\tabcolsep{4pt} 
{
\setlength{\extrarowheight}{2.5pt}
\vspace{2pt}
\begin{tabular}{cccccccccccc}
\toprule
\multicolumn{1}{c|}{}                                                                       & \multicolumn{1}{c|}{}                                                                          & \multicolumn{2}{c|}{\begin{tabular}[c]{@{}c@{}}MELR ($\downarrow$)\\ ($\times10^{-2}$)\end{tabular}}  & \multicolumn{2}{c|}{\begin{tabular}[c]{@{}c@{}}MELRw ($\downarrow$)\\ ($\times10^{-2}$)\end{tabular}} & \multicolumn{2}{c|}{\begin{tabular}[c]{@{}c@{}}covRMSE ($\downarrow$)\\ ($\times10^{-2}$)\end{tabular}} & \multicolumn{2}{c|}{\begin{tabular}[c]{@{}c@{}}Wass1 ($\downarrow$)\\ ($\times10^{-2}$)\end{tabular}} & \multicolumn{2}{c}{\begin{tabular}[c]{@{}c@{}}TCM ($\downarrow$)\\ ($\times10^{-2}$)\end{tabular}} \\
\multicolumn{1}{c|}{\multirow{-2}{*}{\begin{tabular}[c]{@{}c@{}}Batch\\ size\end{tabular}}} & \multicolumn{1}{c|}{\multirow{-2}{*}{\begin{tabular}[c]{@{}c@{}}Learning\\ rate\end{tabular}}} & Base                        & \multicolumn{1}{c|}{DySLIM}                               & Base                        & \multicolumn{1}{c|}{DySLIM}                               & Base                         & \multicolumn{1}{c|}{DySLIM}                                & Base                        & \multicolumn{1}{c|}{DySLIM}                               & Base                                 & DySLIM                                        \\ \midrule
\multicolumn{2}{l}{\textbf{Pushforward}}                                                                                                                                                     &                             &                                                           &                             &                                                           &                              &                                                            &                             &                                                           &                                      &                                               \\ \midrule
\multicolumn{1}{c|}{}                                                                       & \multicolumn{1}{c|}{5e-4}                                                                      & 17.2                        & \multicolumn{1}{c|}{\textbf{2.65}}                        & 9.87                        & \multicolumn{1}{c|}{\textbf{0.66}}                        & 23.6                         & \multicolumn{1}{c|}{\textbf{7.41}}                         & 69.3                        & \multicolumn{1}{c|}{\textbf{5.20}}                        & 83.5                                 & \textbf{3.90}                                 \\
\multicolumn{1}{c|}{}                                                                       & \multicolumn{1}{c|}{1e-4}                                                                      & 3.16                        & \multicolumn{1}{c|}{\textbf{3.13}}                        & 0.49                        & \multicolumn{1}{c|}{0.77}                                 & 6.27                         & \multicolumn{1}{c|}{7.28}                                  & 5.96                        & \multicolumn{1}{c|}{\textbf{5.04}}                        & 1.17                                 & 3.15                                          \\
\multicolumn{1}{c|}{}                                                                       & \multicolumn{1}{c|}{5e-5}                                                                      & 4.37                        & \multicolumn{1}{c|}{\textbf{4.05}}                        & 0.61                        & \multicolumn{1}{c|}{0.82}                                 & 6.74                         & \multicolumn{1}{c|}{\textbf{6.57}}                         & 5.89                        & \multicolumn{1}{c|}{\textbf{4.99}}                        & 2.75                                 & \textbf{2.62}                                 \\
\multicolumn{1}{c|}{\multirow{-4}{*}{32}}                                                   & \multicolumn{1}{c|}{1e-5}                                                                      & 11.2                        & \multicolumn{1}{c|}{\textbf{9.11}}                        & 1.29                        & \multicolumn{1}{c|}{\textbf{0.99}}                        & 9.00                         & \multicolumn{1}{c|}{\textbf{8.32}}                         & 8.33                        & \multicolumn{1}{c|}{\textbf{7.86}}                        & 4.80                                 & \textbf{3.98}                                 \\ \midrule
\multicolumn{1}{c|}{}                                                                       & \multicolumn{1}{c|}{5e-4}                                                                      & 18.3                        & \multicolumn{1}{c|}{\textbf{2.40}}                        & 9.81                        & \multicolumn{1}{c|}{\textbf{0.66}}                        & 22.6                         & \multicolumn{1}{c|}{\textbf{7.33}}                         & 68.6                        & \multicolumn{1}{c|}{\textbf{4.49}}                        & 85.6                                 & \textbf{2.07}                                 \\
\multicolumn{1}{c|}{}                                                                       & \multicolumn{1}{c|}{1e-4}                                                                      & 3.19                        & \multicolumn{1}{c|}{\textbf{2.95}}                        & 0.54                        & \multicolumn{1}{c|}{0.70}                                 & 6.90                         & \multicolumn{1}{c|}{6.91}                                  & 5.09                        & \multicolumn{1}{c|}{5.15}                                 & 1.62                                 & 3.30                                          \\
\multicolumn{1}{c|}{}                                                                       & \multicolumn{1}{c|}{5e-5}                                                                      & 4.52                        & \multicolumn{1}{c|}{\textbf{3.71}}                        & 0.71                        & \multicolumn{1}{c|}{0.82}                                 & 7.02                         & \multicolumn{1}{c|}{\textbf{6.66}}                         & 5.81                        & \multicolumn{1}{c|}{\textbf{5.21}}                        & 4.12                                 & 4.59                                          \\
\multicolumn{1}{c|}{\multirow{-4}{*}{64}}                                                   & \multicolumn{1}{c|}{1e-5}                                                                      & 10.6                        & \multicolumn{1}{c|}{\textbf{8.27}}                        & 1.22                        & \multicolumn{1}{c|}{\textbf{0.91}}                        & 7.94                         & \multicolumn{1}{c|}{\textbf{7.52}}                         & 9.44                        & \multicolumn{1}{c|}{\textbf{6.61}}                        & 4.58                                 & \textbf{3.75}                                 \\ \midrule
\multicolumn{1}{c|}{}                                                                       & \multicolumn{1}{c|}{5e-4}                                                                      & 73.2                        & \multicolumn{1}{c|}{\textbf{2.35}}                        & 61.3                        & \multicolumn{1}{c|}{\textbf{0.60}}                        & 80.2                         & \multicolumn{1}{c|}{\textbf{7.27}}                         & 26.3                        & \multicolumn{1}{c|}{\textbf{5.30}}                        & 33.4                                 & \textbf{2.28}                                 \\
\multicolumn{1}{c|}{}                                                                       & \multicolumn{1}{c|}{{\color[HTML]{036400} 1e-4}}                                               & {\color[HTML]{036400} 3.19} & \multicolumn{1}{c|}{{\color[HTML]{036400} \textbf{2.46}}} & {\color[HTML]{036400} 0.53} & \multicolumn{1}{c|}{{\color[HTML]{036400} 0.53}}          & {\color[HTML]{036400} 6.81}  & \multicolumn{1}{c|}{{\color[HTML]{036400} \textbf{6.69}}}  & {\color[HTML]{036400} 4.64} & \multicolumn{1}{c|}{{\color[HTML]{036400} \textbf{4.51}}} & {\color[HTML]{036400} 3.68}          & {\color[HTML]{036400} \textbf{0.72}}          \\
\multicolumn{1}{c|}{}                                                                       & \multicolumn{1}{c|}{5e-5}                                                                      & 4.18                        & \multicolumn{1}{c|}{\textbf{3.53}}                        & 0.55                        & \multicolumn{1}{c|}{0.71}                                 & 6.73                         & \multicolumn{1}{c|}{7.55}                                  & 5.20                        & \multicolumn{1}{c|}{5.33}                                 & 1.69                                 & 3.55                                          \\
\multicolumn{1}{c|}{\multirow{-4}{*}{128}}                                                  & \multicolumn{1}{c|}{1e-5}                                                                      & 9.66                        & \multicolumn{1}{c|}{\textbf{7.54}}                        & 1.09                        & \multicolumn{1}{c|}{\textbf{0.75}}                        & 8.71                         & \multicolumn{1}{c|}{\textbf{8.10}}                         & 8.59                        & \multicolumn{1}{c|}{\textbf{6.06}}                        & 4.59                                 & \textbf{3.00}                                 \\ \midrule
\multicolumn{2}{l}{\textbf{Curriculum}}                                                                                                                                                      &                             &                                                           &                             &                                                           &                              &                                                            &                             &                                                           &                                      &                                               \\ \midrule
\multicolumn{1}{c|}{32}                                                                     & \multicolumn{1}{c|}{5e-4}                                                                      & 5.14                        & \multicolumn{1}{c|}{\textbf{2.21}}                        & 0.74                        & \multicolumn{1}{c|}{0.74}                                 & 8.12                         & \multicolumn{1}{c|}{\textbf{7.41}}                         & 13.6                        & \multicolumn{1}{c|}{\textbf{4.42}}                        & 4.60                                 & \textbf{2.45}                                 \\
\multicolumn{1}{c|}{{\color[HTML]{036400} 64}}                                              & \multicolumn{1}{c|}{{\color[HTML]{036400} 5e-4}}                                               & {\color[HTML]{036400} 5.35} & \multicolumn{1}{c|}{{\color[HTML]{036400} \textbf{1.64}}} & {\color[HTML]{036400} 0.95} & \multicolumn{1}{c|}{{\color[HTML]{036400} \textbf{0.45}}} & {\color[HTML]{036400} 8.13}  & \multicolumn{1}{c|}{{\color[HTML]{036400} \textbf{6.95}}}  & {\color[HTML]{036400} 9.66} & \multicolumn{1}{c|}{{\color[HTML]{036400} \textbf{4.76}}} & {\color[HTML]{036400} 3.50}          & {\color[HTML]{036400} \textbf{2.83}}          \\
\multicolumn{1}{c|}{128}                                                                    & \multicolumn{1}{c|}{5e-4}                                                                      & 6.80                        & \multicolumn{1}{c|}{\textbf{3.06}}                        & 1.19                        & \multicolumn{1}{c|}{1.47}                                 & 8.60                         & \multicolumn{1}{c|}{\textbf{8.13}}                         & 70.5                        & \multicolumn{1}{c|}{\textbf{8.77}}                        & 27.4                                 & \textbf{3.50}                                 \\
\multicolumn{1}{c|}{256}                                                                    & \multicolumn{1}{c|}{5e-4}                                                                      & 42.8                        & \multicolumn{1}{c|}{\textbf{3.35}}                        & 31.3                        & \multicolumn{1}{c|}{\textbf{1.53}}                        & diverge                      & \multicolumn{1}{c|}{\textbf{8.91}}                         & 89.1                        & \multicolumn{1}{c|}{\textbf{19.7}}                        & 817                                  & \textbf{65.0}                                 \\
\multicolumn{1}{c|}{512}                                                                    & \multicolumn{1}{c|}{5e-4}                                                                      & 25.8                        & \multicolumn{1}{c|}{\textbf{4.42}}                        & 23.1                        & \multicolumn{1}{c|}{\textbf{1.78}}                        & diverge                      & \multicolumn{1}{c|}{\textbf{9.31}}                         & 166                         & \multicolumn{1}{c|}{\textbf{21.4}}                        & 819                                  & \textbf{7.53}                                 \\ \midrule
\multicolumn{2}{l}{\textbf{1-step}}                                                                                                                                                          &                             &                                                           &                             &                                                           &                              &                                                            &                             &                                                           &                                      &                                               \\ \midrule
\multicolumn{1}{c|}{32}                                                                     & \multicolumn{1}{c|}{5e-4}                                                                      & 3.11                        & \multicolumn{1}{c|}{\textbf{2.05}}                        & 0.45                        & \multicolumn{1}{c|}{0.98}                                 & 7.30                         & \multicolumn{1}{c|}{8.20}                                  & 8.17                        & \multicolumn{1}{c|}{\textbf{4.87}}                        & 2.58                                 & 3.76                                          \\
\multicolumn{1}{c|}{{\color[HTML]{036400} 64}}                                              & \multicolumn{1}{c|}{{\color[HTML]{036400} 5e-4}}                                               & {\color[HTML]{036400} 2.77} & \multicolumn{1}{c|}{{\color[HTML]{036400} \textbf{1.84}}} & {\color[HTML]{036400} 0.44} & \multicolumn{1}{c|}{{\color[HTML]{036400} 0.85}}          & {\color[HTML]{036400} 7.93}  & \multicolumn{1}{c|}{{\color[HTML]{036400} \textbf{7.30}}}  & {\color[HTML]{036400} 16.2} & \multicolumn{1}{c|}{{\color[HTML]{036400} \textbf{5.55}}} & {\color[HTML]{036400} 5.39}          & {\color[HTML]{036400} \textbf{2.45}}          \\
\multicolumn{1}{c|}{128}                                                                    & \multicolumn{1}{c|}{5e-4}                                                                      & 1.47                        & \multicolumn{1}{c|}{1.80}                                 & 0.28                        & \multicolumn{1}{c|}{0.89}                                 & 7.03                         & \multicolumn{1}{c|}{\textbf{7.02}}                         & 6.65                        & \multicolumn{1}{c|}{\textbf{5.92}}                        & 2.66                                 & 3.82                                          \\
\multicolumn{1}{c|}{256}                                                                    & \multicolumn{1}{c|}{5e-4}                                                                      & 30.5                        & \multicolumn{1}{c|}{\textbf{1.90}}                        & 24.7                        & \multicolumn{1}{c|}{\textbf{0.88}}                        & 116                          & \multicolumn{1}{c|}{\textbf{7.04}}                         & 30.4                        & \multicolumn{1}{c|}{\textbf{18.4}}                        & 347                                  & \textbf{4.10}                                 \\ \bottomrule
\end{tabular}
}
}
\end{table*}

\begin{table*}[t]
  \centering
  \small
\caption{
Kolmogorov flow: metrics for pushforward training for $40$ time steps. ($\lambda_1 = 0$, $\lambda_2 = 100$). Boldface numbers indicate that the metric is improved by our regularization. }
\label{tab:ns_stats_very_long_rollout}
{
\setlength\tabcolsep{4pt} 
{
\setlength{\extrarowheight}{2.5pt}
\vspace{2pt}
\begin{tabular}{l|c|c|cc|cc|cc|cc|cc}
\hline
& \multirow{2}{*}{\begin{tabular}[c]{@{}c@{}}Batch\\ size\end{tabular}} & \multirow{2}{*}{\begin{tabular}[c]{@{}c@{}}Learning\\ rate\end{tabular}} & \multicolumn{2}{c|}{\begin{tabular}[c]{@{}c@{}}MELR ($\downarrow$)\\ ($\times10^{-2}$)\end{tabular}} & \multicolumn{2}{c|}{\begin{tabular}[c]{@{}c@{}}MELRw ($\downarrow$)\\ ($\times10^{-2}$)\end{tabular}} & \multicolumn{2}{c|}{\begin{tabular}[c]{@{}c@{}}covRMSE ($\downarrow$)\\ ($\times10^{-2}$)\end{tabular}} & \multicolumn{2}{c|}{\begin{tabular}[c]{@{}c@{}}Wass1 ($\downarrow$)\\ ($\times10^{-2}$)\end{tabular}} & \multicolumn{2}{c}{\begin{tabular}[c]{@{}c@{}}TCM ($\downarrow$)\\ ($\times10^{-2}$)\end{tabular}} \\
                          &                                                                       &                                                                          & Base    & DySLIM    & Base      & DySLIM       & Base     & DySLIM      & Base     & DySLIM    & Base    & DySLIM      \\ \hline
Pushforward               & 128    & 5e-4           & 62.22   & \textbf{8.51}             & 24.01   & \textbf{1.65}    & 42.6      & \textbf{8.07}    & 175                                  & \textbf{11.34}          & 161        & \textbf{4.3}                                 \\\hline
\end{tabular}
}}
\end{table*}

\begin{figure*}[t]
    \centering
    \includegraphics[width=0.9\textwidth]{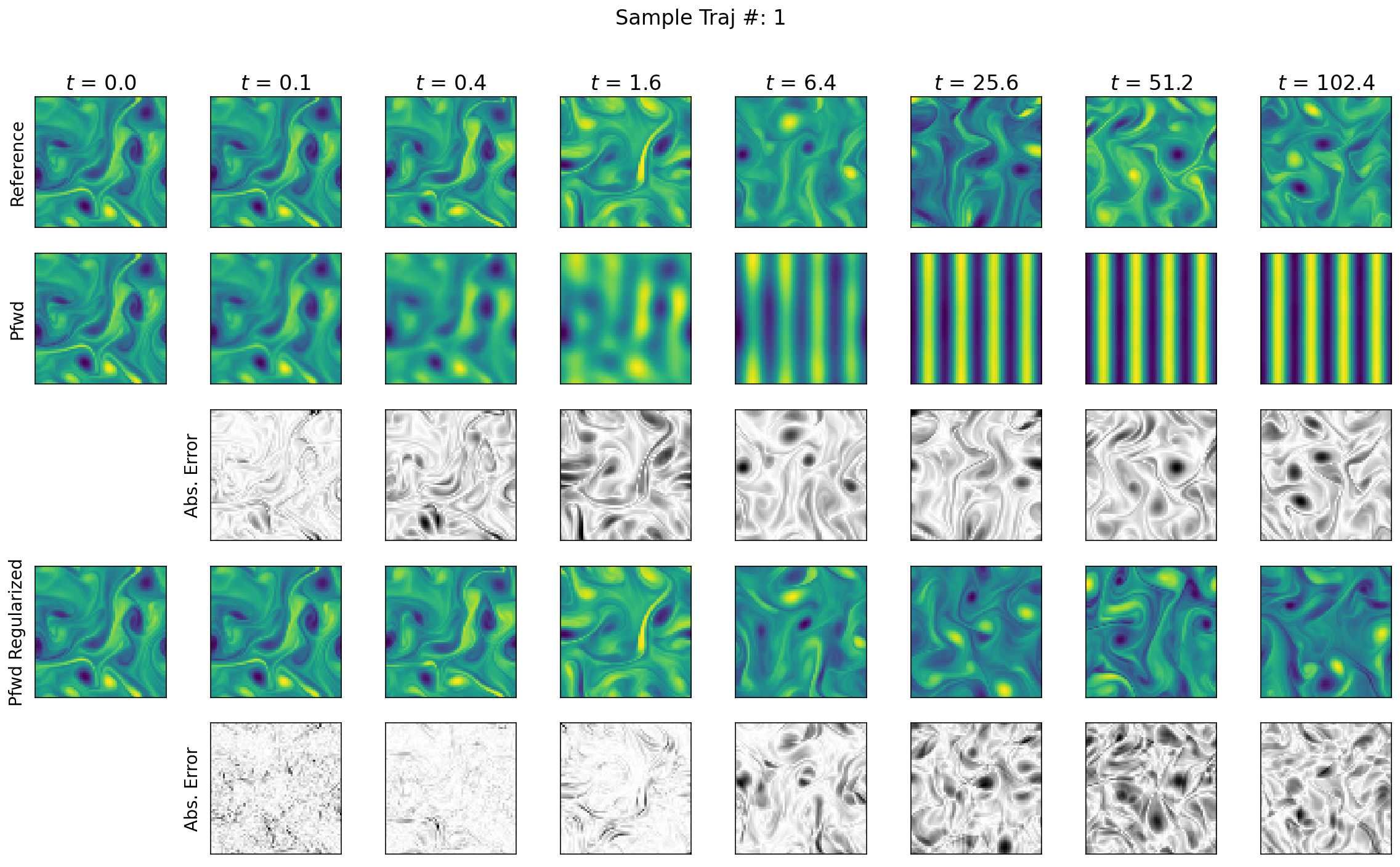}
    \includegraphics[width=0.9\textwidth]{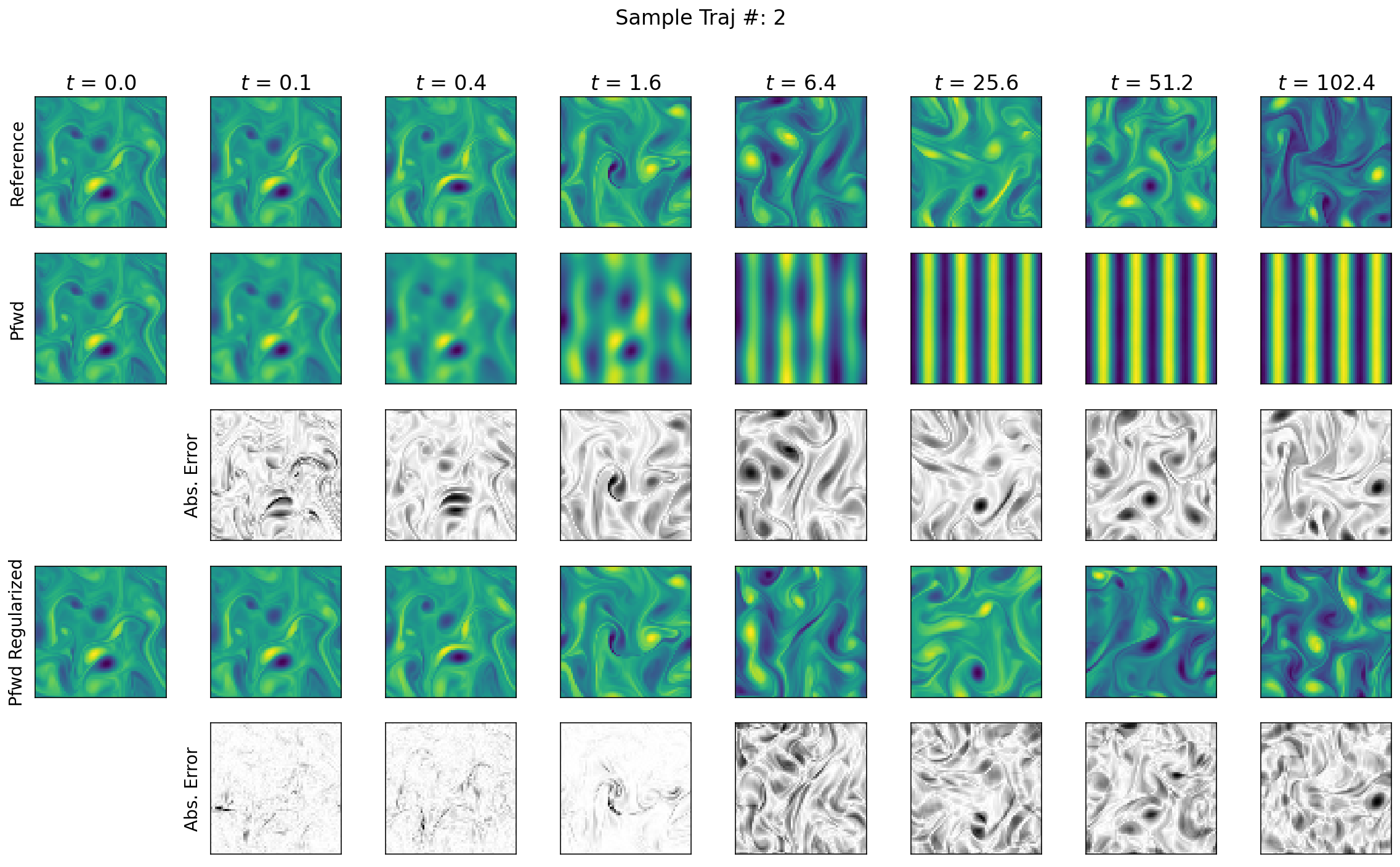}
    \caption{Additional samples of reference trajectories of Navier Stokes system with Kolmogorov forcing and predicted trajectory generated with models trained using the pushforward objective, with and without regularization. ($\lambda_1 = 0$, $\lambda_2 = 100$, batch size = $128$ and learning rate= $5\mathrm{e}^{-4}$).}
     \label{fig:app_ns_further_samples}
\end{figure*}

\begin{figure*}[t]
    \centering
    \includegraphics[width=0.9\textwidth]{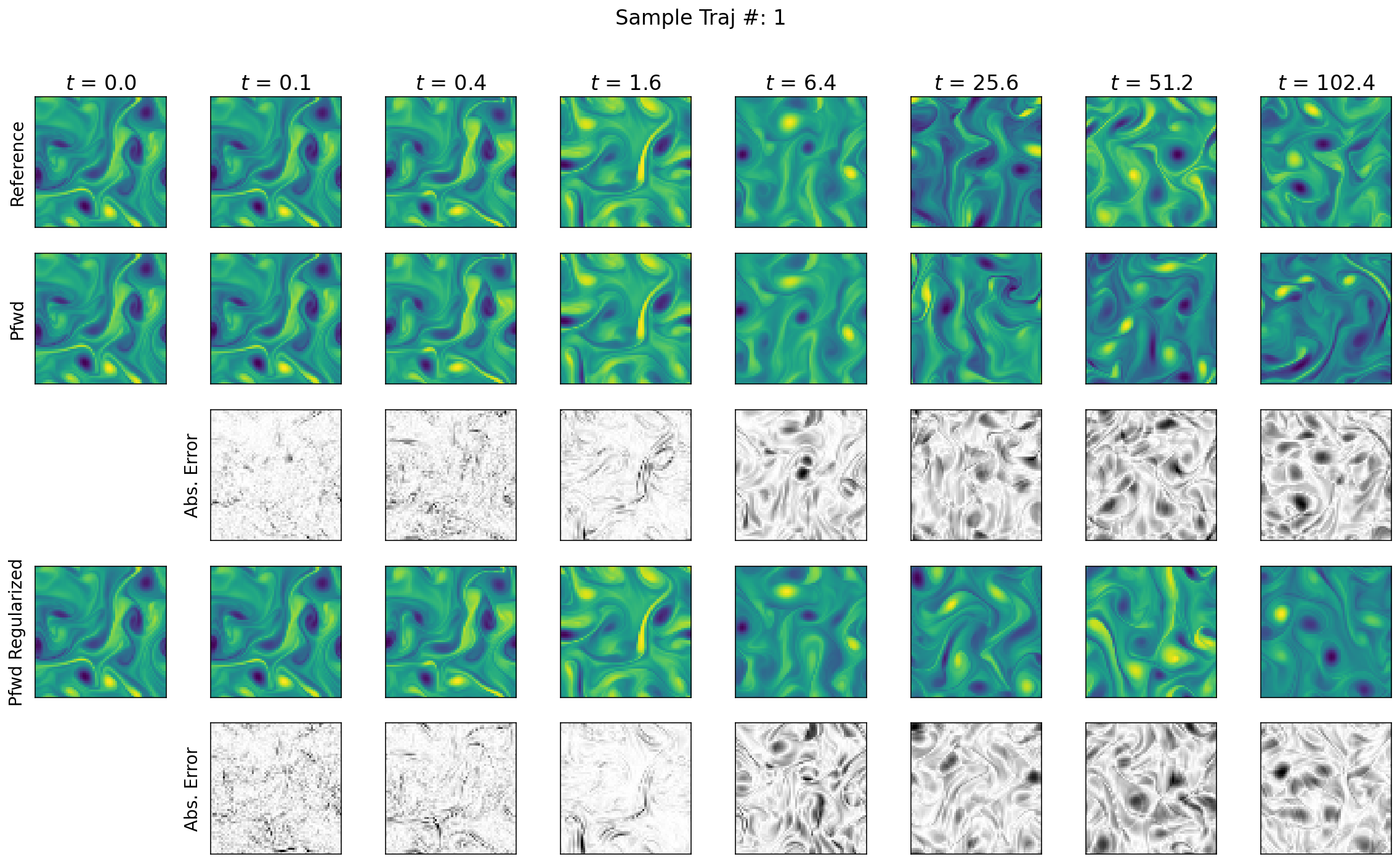}
    \includegraphics[width=0.9\textwidth]{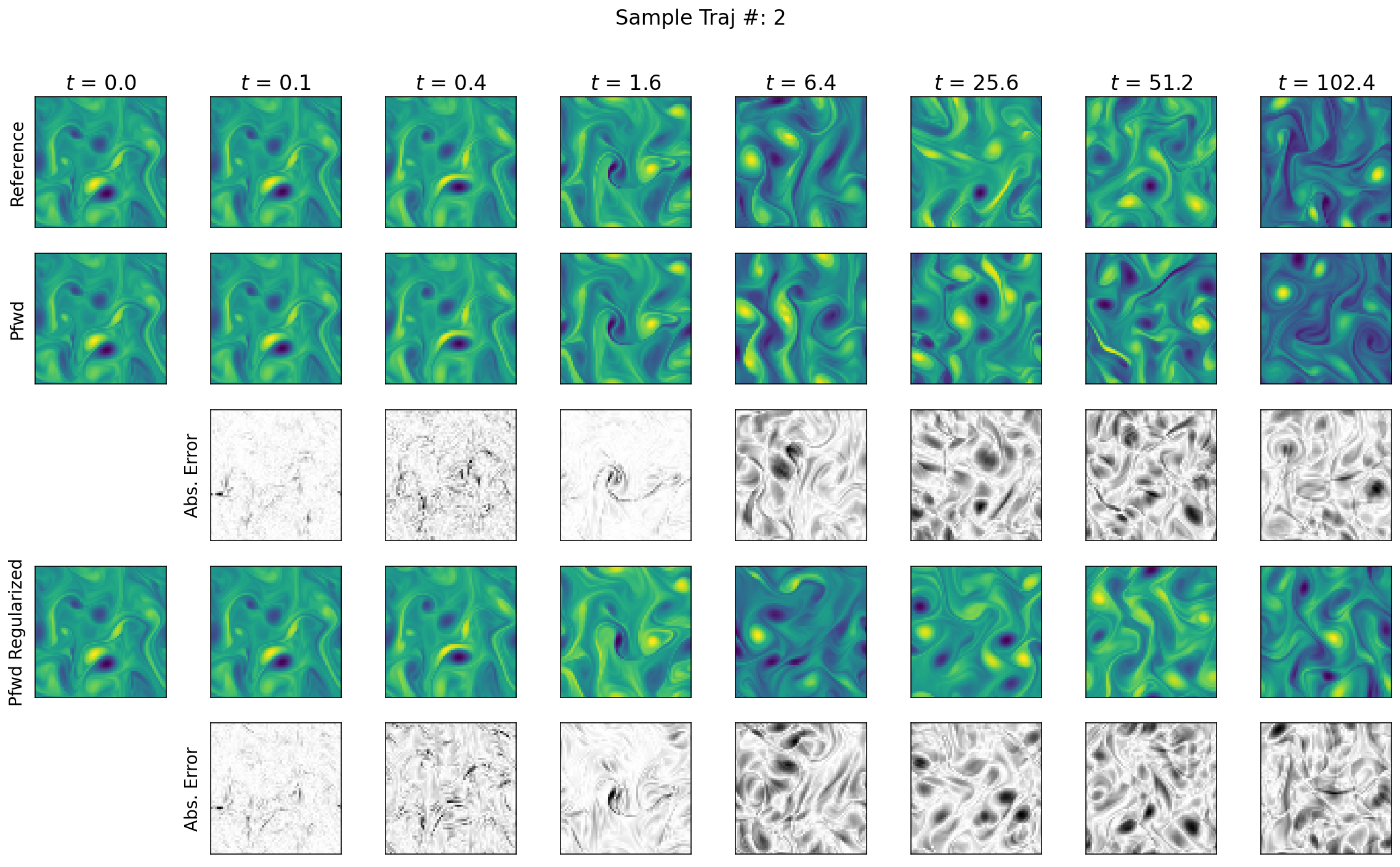}
    \caption{Samples of reference trajectories of Navier Stokes system with Kolmogorov forcing and predicted trajectory generated with models trained the pushforward loss, with and without regularization.
    In this case, we show samples of the random seed with the best results for the unregularized training.
    ($\lambda_1 = 0$, $\lambda_2 = 100$, \texttt{batch size} = 128 and \texttt{learning rate}= 5e-4).}
    \label{fig:app_ns_further_samples_best_bs_128}
\end{figure*}

\Figref{fig:app_ns_further_samples} provides additional samples of the trajectories presented in \Figref{fig:ns_evolution} for the training with the pushforward objective.
From \Figref{fig:app_ns_further_samples}, we observe that models trained with the unregularized objectives remain highly dissipative despite using different random seeds.
For this configuration of parameters, we were able obtain only one stable model trained without the regularization among the random seeds, which we present in \Figref{fig:app_ns_further_samples_best_bs_128}.
In this case, even though the trajectories are visually more realistic, if we consider the metrics used for evaluation, Table \ref{tab:ns_128_best_pfwd_versus_reg} shows that the models trained with the DySLIM regularization still provide better statistics.

\begin{figure*}[t]
    \centering
    \includegraphics[width=0.9\textwidth]{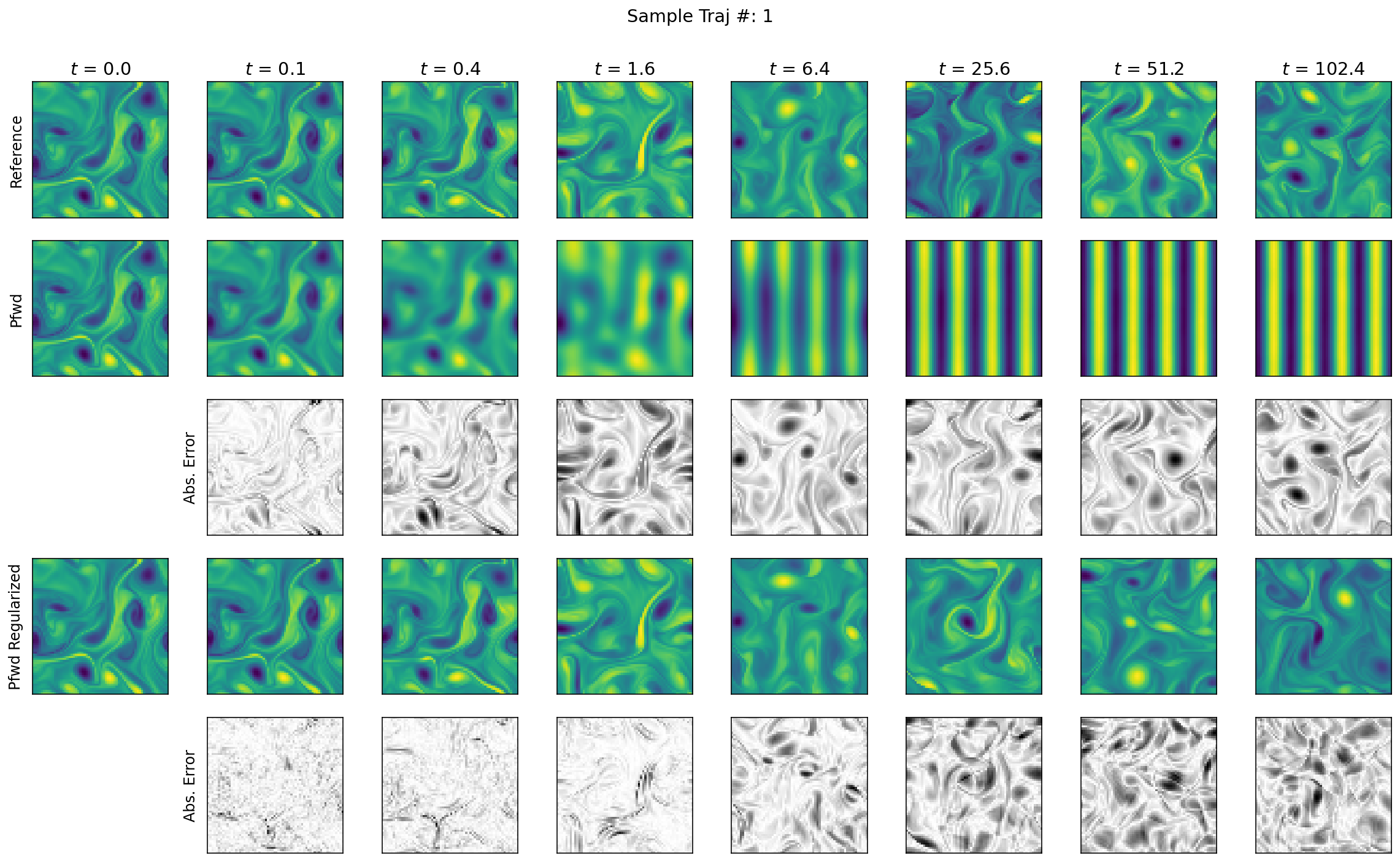}
    \includegraphics[width=0.9\textwidth]{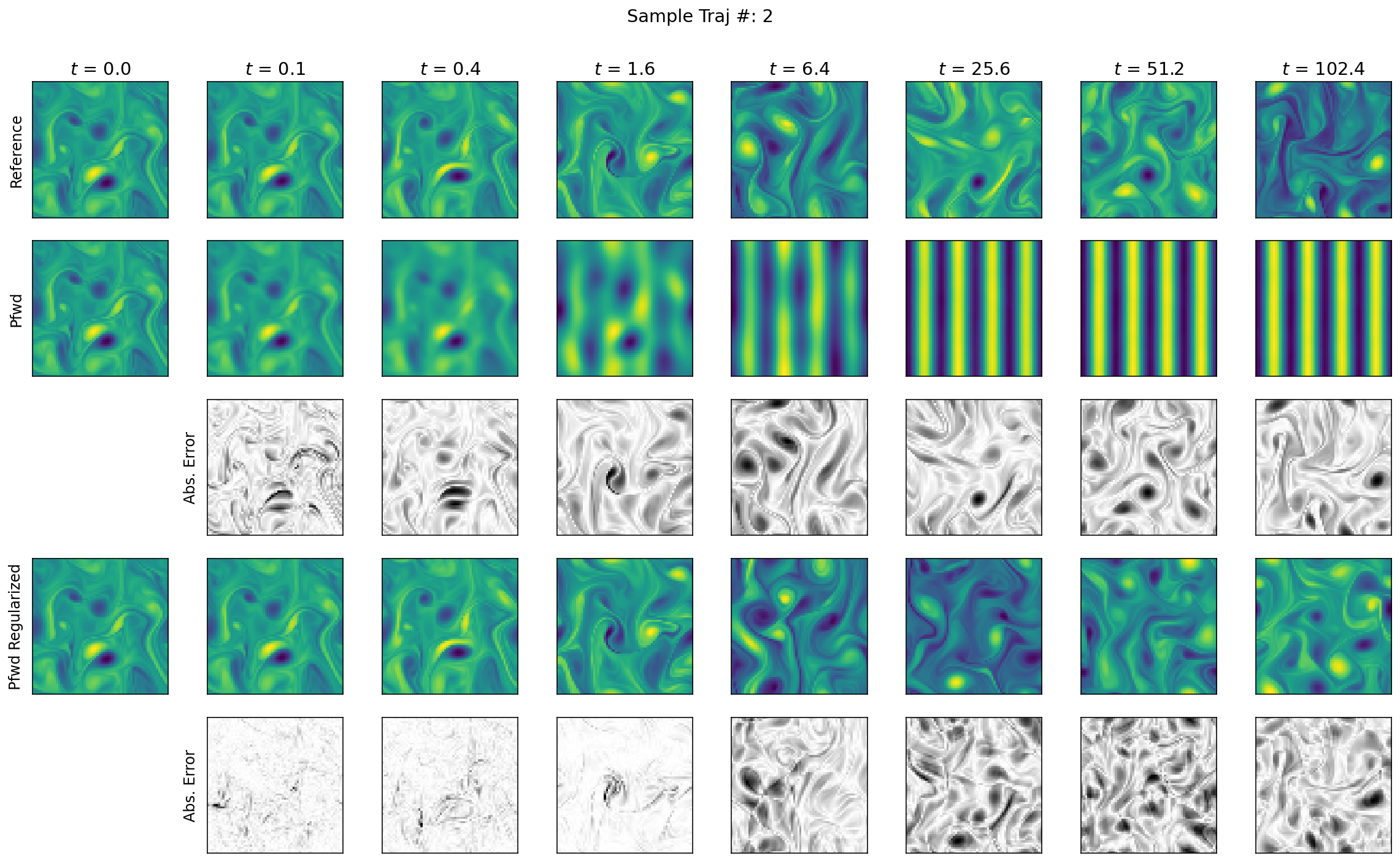}
    \caption{Samples of reference trajectories of Navier Stokes system with Kolmogorov forcing and predicted trajectory generated with models trained the pushforward loss, with and without regularization with a batch size of $256$ for the same random seed. ($\lambda_1 = 0$, $\lambda_2 = 100$, and learning rate = $5\mathrm{e}^{-4}$).}
    \label{fig:app_ns_further_samples_best_bs_256}
\end{figure*}

\begin{figure*}[t]
    \centering
    \includegraphics[width=0.9\textwidth]{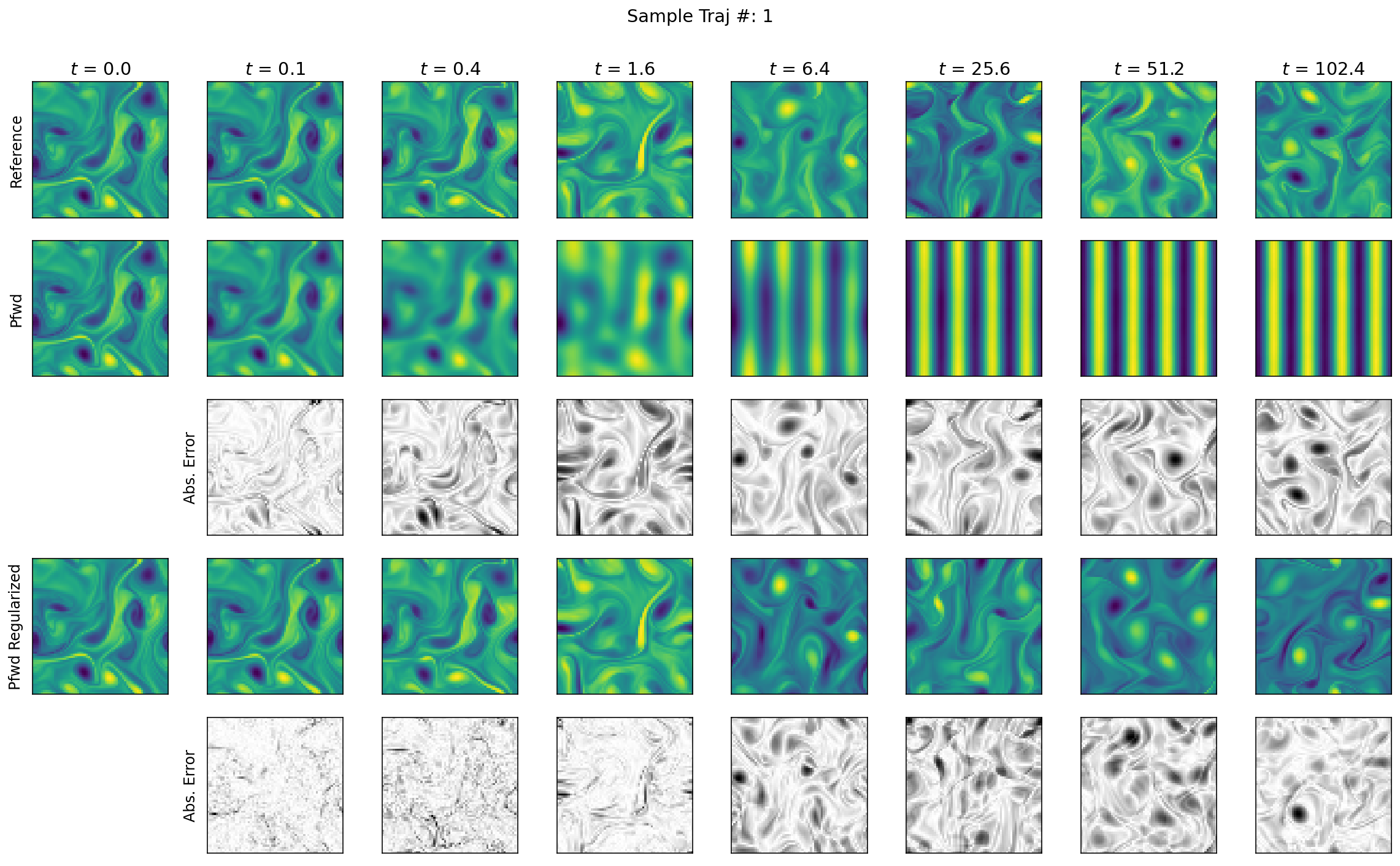}
    \includegraphics[width=0.9\textwidth]{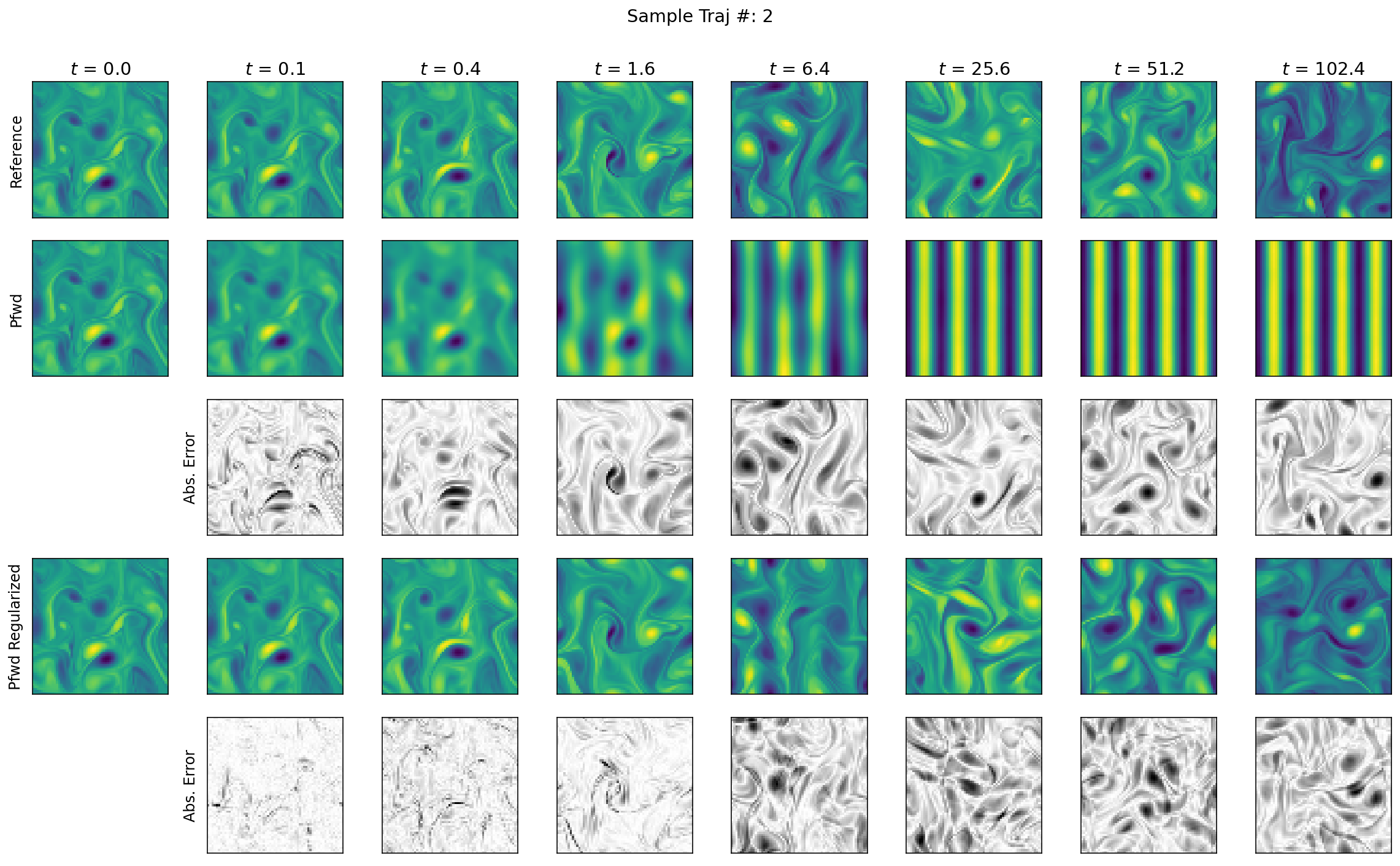}
    \caption{Samples of reference trajectories of Navier Stokes system with Kolmogorov forcing and predicted trajectory generated with models trained the pushforward loss, with and without regularization with a batch size of $512$ for the same random seed. ($\lambda_1 = 0$, $\lambda_2 = 100$, and learning rate = $5\mathrm{e}^{-4}$).}
    \label{fig:app_ns_further_samples_best_bs_512}
\end{figure*}

We point out that these highly dissipative models are also present as we increase the batch size.
For example, Figures \ref{fig:app_ns_further_samples_best_bs_256} and \ref{fig:app_ns_further_samples_best_bs_512} show the same phenomenon for trajectories learned without regularization for batch sizes of $256$ and $512$. 

In addition, Table \ref{tab:ns_stats_very_long_rollout} shows the statistics of models trained with much longer time-horizons ($40$ time steps instead of $10$) for the Kolmogorov flow.
The baseline in this case becomes completely uninformative, whereas the regularized version still provides models with reasonable statistics.

\begin{figure*}[ht]
    \centering
    \begin{minipage}[c]{0.3\textwidth}
        \centering
        \vfill
        \includegraphics[width=\textwidth, trim={0mm 0mm 0mm 0mm}, clip]{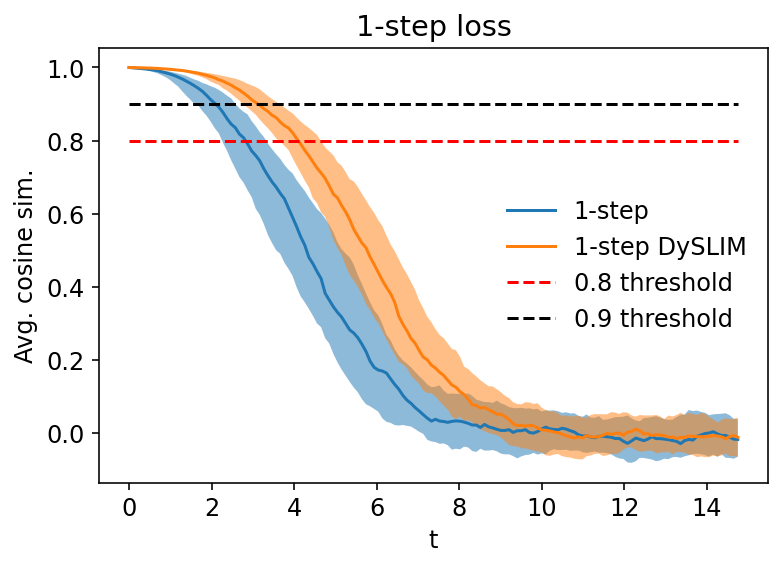}

        \includegraphics[width=\textwidth, trim={0mm 0mm 0mm 0mm}, clip]{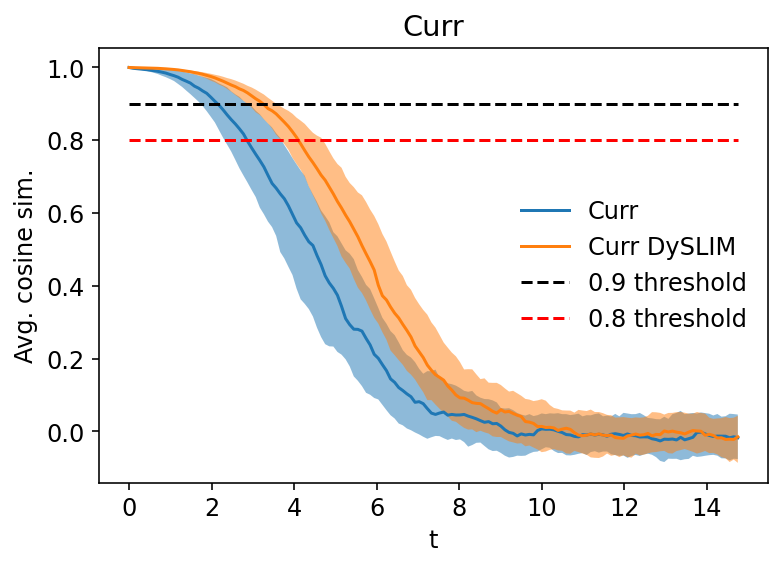}

        \subfloat[]{\includegraphics[width=\textwidth, trim={0mm 0mm 0mm 0mm}, clip]{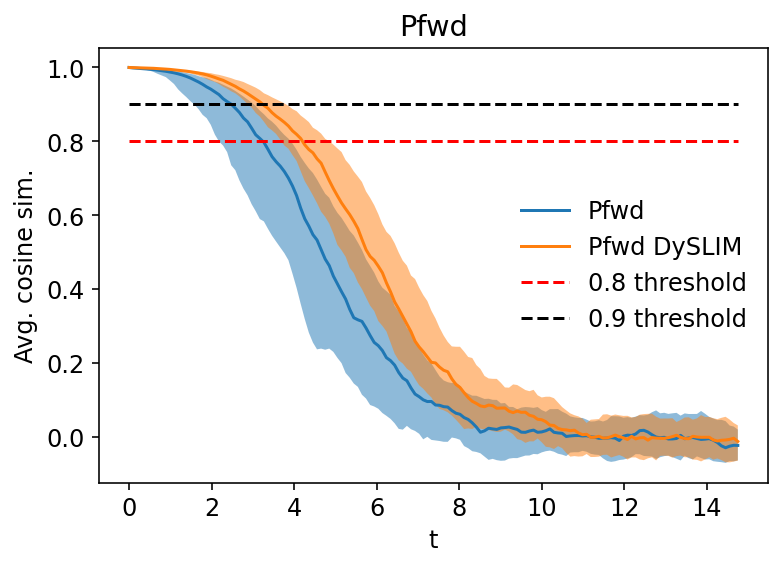}}
    \end{minipage}
        \begin{minipage}[c]{0.3\textwidth}
        \centering
        \vfill
        \includegraphics[width=\textwidth, trim={0mm 0mm 0mm 0mm}, clip]{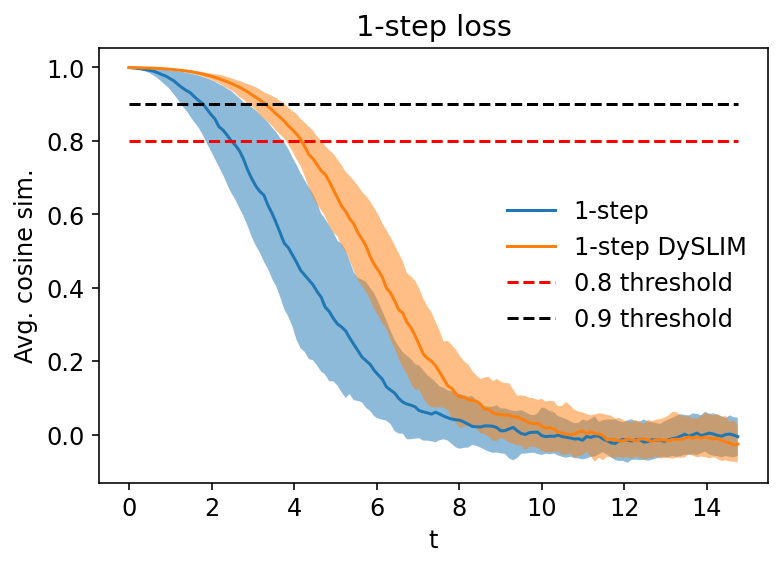}
        
        \includegraphics[width=\textwidth, trim={0mm 0mm 0mm 0mm}, clip]{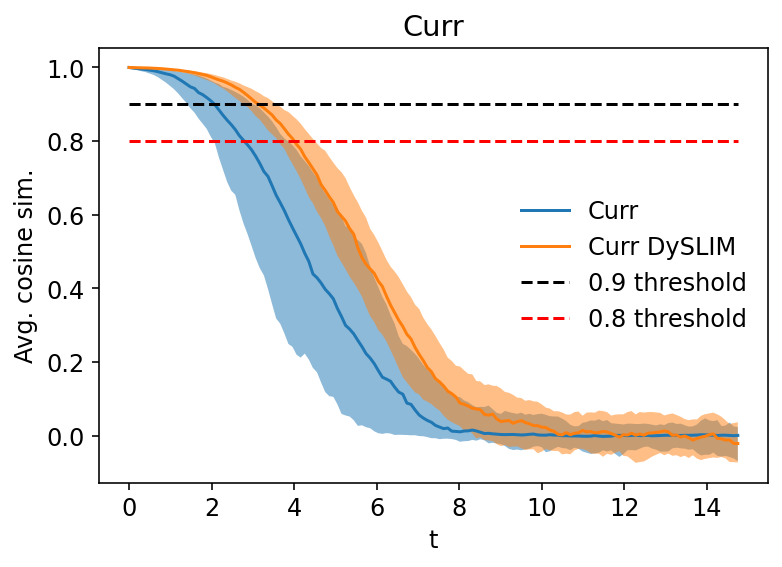}

        \subfloat[]{\includegraphics[width=\textwidth, trim={0mm 0mm 0mm 0mm}, clip]{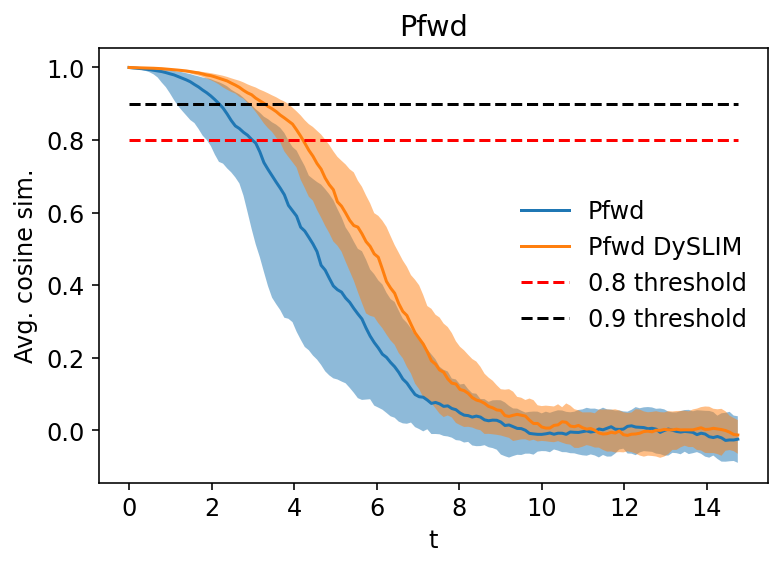}}
    \end{minipage}
        \begin{minipage}[c]{0.3\textwidth}
        \centering
        \vfill
        \includegraphics[width=\textwidth, trim={0mm 0mm 0mm 0mm}, clip]{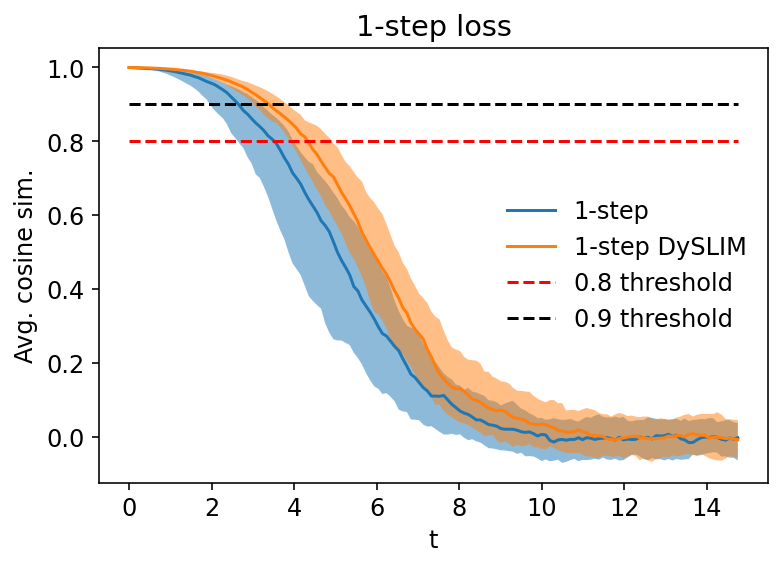}
        \includegraphics[width=\textwidth, trim={0mm 0mm 0mm 0mm}, clip]{images/cos_sim_ns_curr_128_main_text.png}

        \subfloat[]{\includegraphics[width=\textwidth, trim={0mm 0mm 0mm 0mm}, clip]{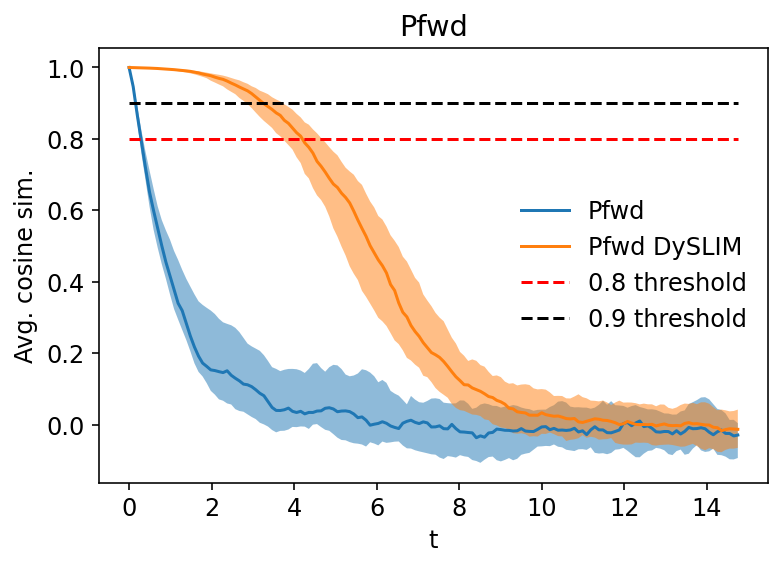}}
    \end{minipage}
    \caption{ Evolution of the cosine similarity ($\uparrow$) overt time for trajectories trained with both regularized and unregularized objectives for different batch sizes of (a) 32, (b) 64, and (c) 128. The solid line is the median among 160 runs, and the shaded regions corresponds second and third quartile. ($\lambda_1 = 0$, $\lambda_2 = 100$, batch size = 128 and learning rate= $5\mathrm{e}^{-4}$).
    }
    \label{fig:ns_ablation_cos_sim}
\end{figure*}

\begin{table}
    \centering
    \caption{Metrics of the best model using the unregularized pushforward loss versus the average of the models trained using the regularized pushforward loss, for batch size = 128 and learning rate = 5e-5.}
    \label{tab:ns_128_best_pfwd_versus_reg}
    \begin{tabular}{c|l|l|l}
    \toprule
             & MELR ($\downarrow$) ($\times10^{-2}$)  & MELRw ($\downarrow$) ($\times10^{-2}$) & covRMSE ($\downarrow$) ($\times10^{-2}$) \\
             \midrule
            Pfwd    &  4.10    &  0.670 & 8.10 \\
~+DySLIM    &  \textbf{2.34}   &  \textbf{0.588}  & \textbf{7.89} \\
       \bottomrule
    \end{tabular}
    
\end{table}

\section{Sinkhorn Divergence for Measure Matching} \label{app:SD_reg}
In this section we provide an ablation for using SD in place of MMD as the measure distance regularizer $\D.$

\begin{figure*}[h]
\includegraphics[width=0.95\textwidth, trim={2mm 0mm 5mm 0mm}, clip]{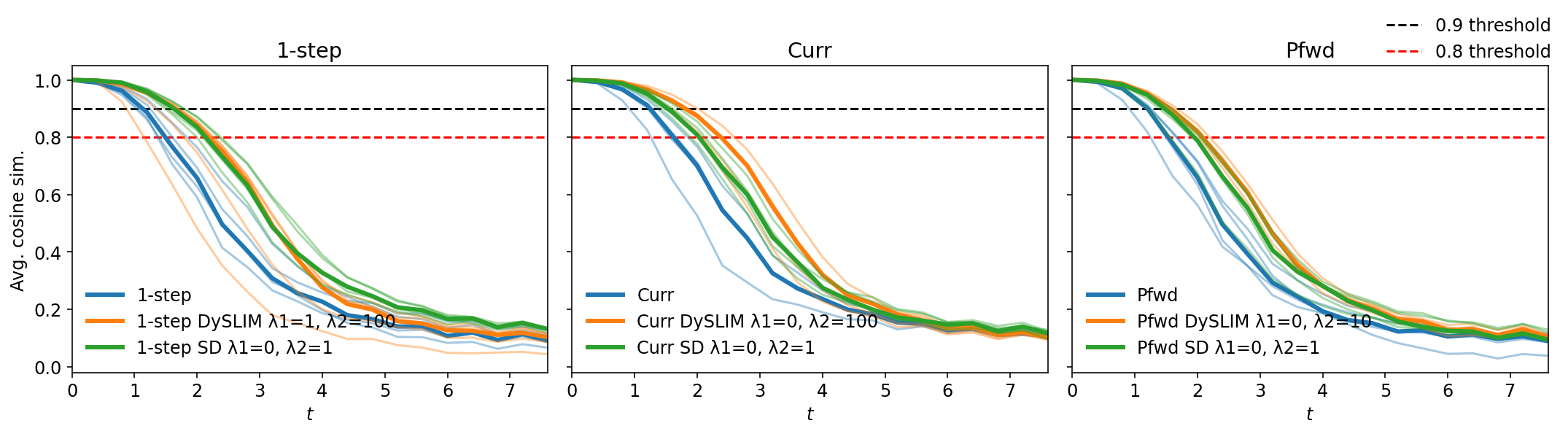}
    \caption{Cosine similarity metric ($\uparrow$) for test trajectories at various rollout times of the Lorenz 63 system using MMD and SD for measure matching during training.
    Each line corresponds to the mean over trajectories of each of five random training seeds, with bold lines indicating median values.
    }
    \label{fig:lz63_cos_sim}
\end{figure*}

\begin{figure*}[h]
    \centering
    \begin{minipage}[c]{0.67\textwidth}  
        \vfill
        \subfloat[]{\includegraphics[width=0.95\textwidth, trim={2mm 0mm 5mm 0mm}, clip]{images/l63_sd.png}}
         \vfill
        \vspace{-0.3cm}
        \subfloat[]{\includegraphics[width=0.95\textwidth, trim={2mm 0mm 5mm 0mm}, clip]{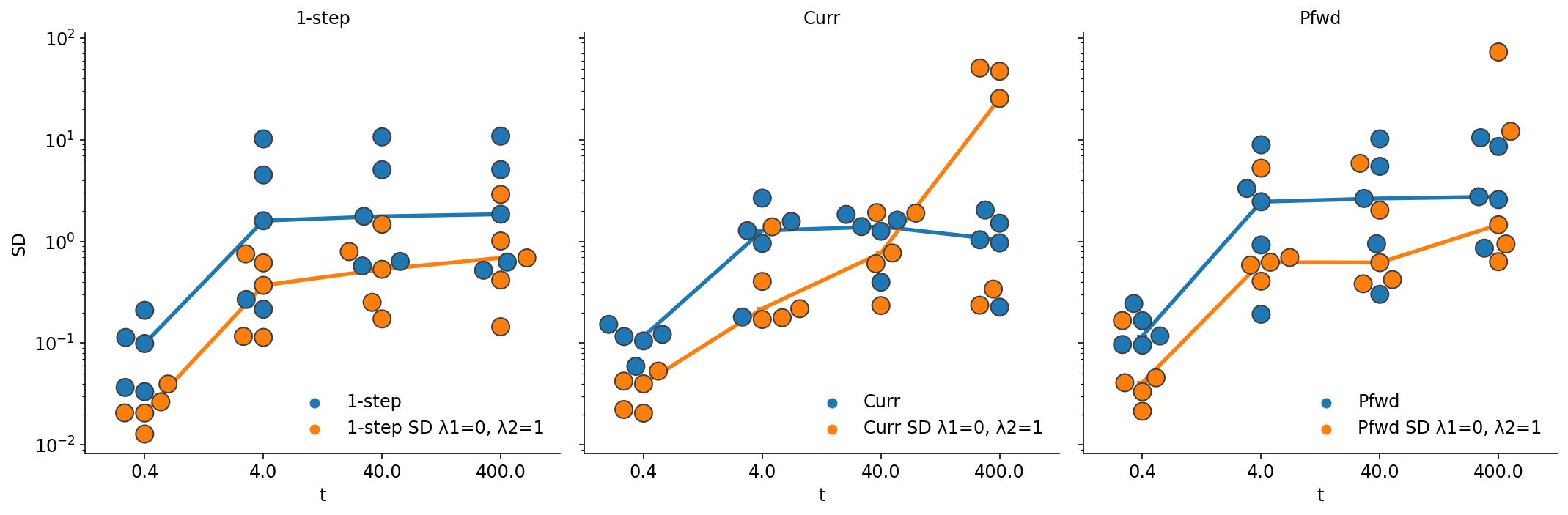}}

    \end{minipage}
    \caption{Sinkhorn Divergence (SD; $\downarrow$) between trajectories at various rollout times of the Lorenz 63 system.
    (a) Values when using MMD for measure matching at training. (b) Values when using SD for measure matching at training. 
    Each point represents a random training seed that remains stable, with the solid line indicating median values.
    }
    \label{fig:lz63_SD}
\end{figure*}

\begin{figure*}[h]
    \centering
    \begin{minipage}[c]{0.67\textwidth}  
        \vfill
        \subfloat[]{\includegraphics[width=0.95\textwidth, trim={2mm 0mm 5mm 0mm}, clip]{images/l63_mmd.png}}
         \vfill
        \vspace{-0.3cm}
        \subfloat[]{\includegraphics[width=0.95\textwidth, trim={2mm 0mm 5mm 0mm}, clip]{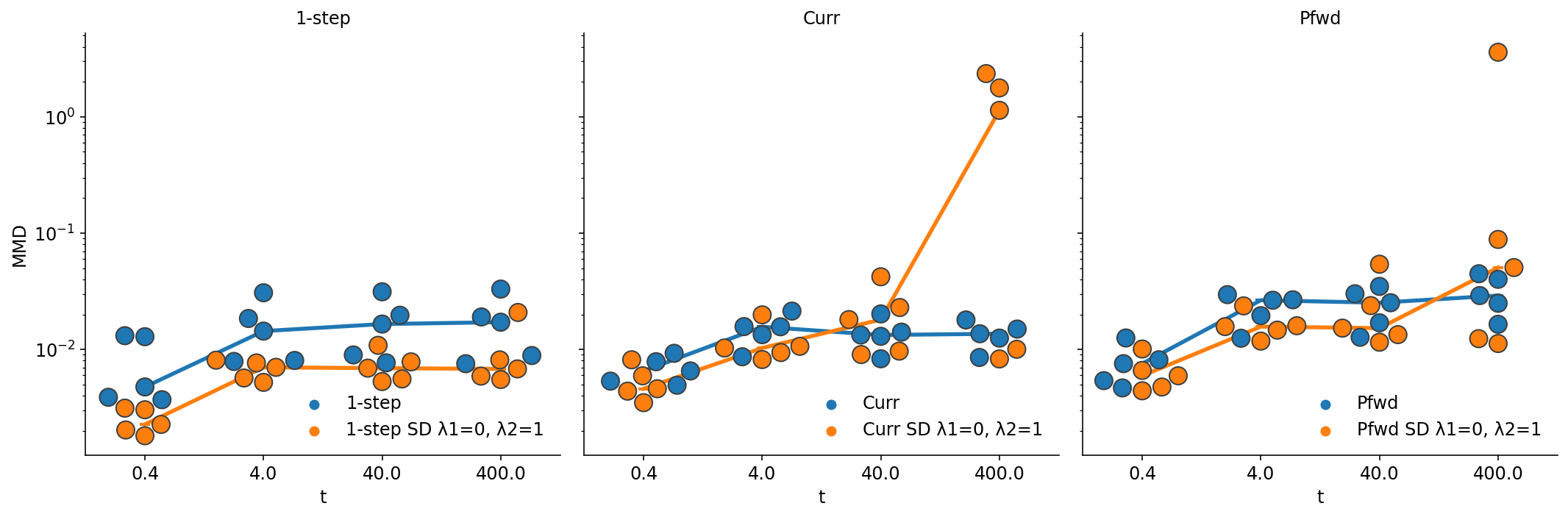}}

    \end{minipage}
    \caption{Maximum Mean Discrepancy (MMD; $\downarrow$) between trajectories at various rollout times of the Lorenz 63 system.
    (a) Values when using MMD for measure matching at training. (b) Values when using SD for measure matching at training. 
    Each point represents a random training seed that remains stable, with the solid line indicating median values. 
    }
    \label{fig:lz63_mmd_ablation}
\end{figure*}

\begin{figure*}[h]
    \centering
    \begin{minipage}[c]{0.67\textwidth}  
        \vfill
        \subfloat[]{\includegraphics[width=0.95\textwidth, trim={2mm 0mm 5mm 0mm}, clip]{images/l63_W_xz.png}}
         \vfill
        \vspace{-0.3cm}
        \subfloat[]{\includegraphics[width=0.95\textwidth, trim={2mm 0mm 5mm 0mm}, clip]{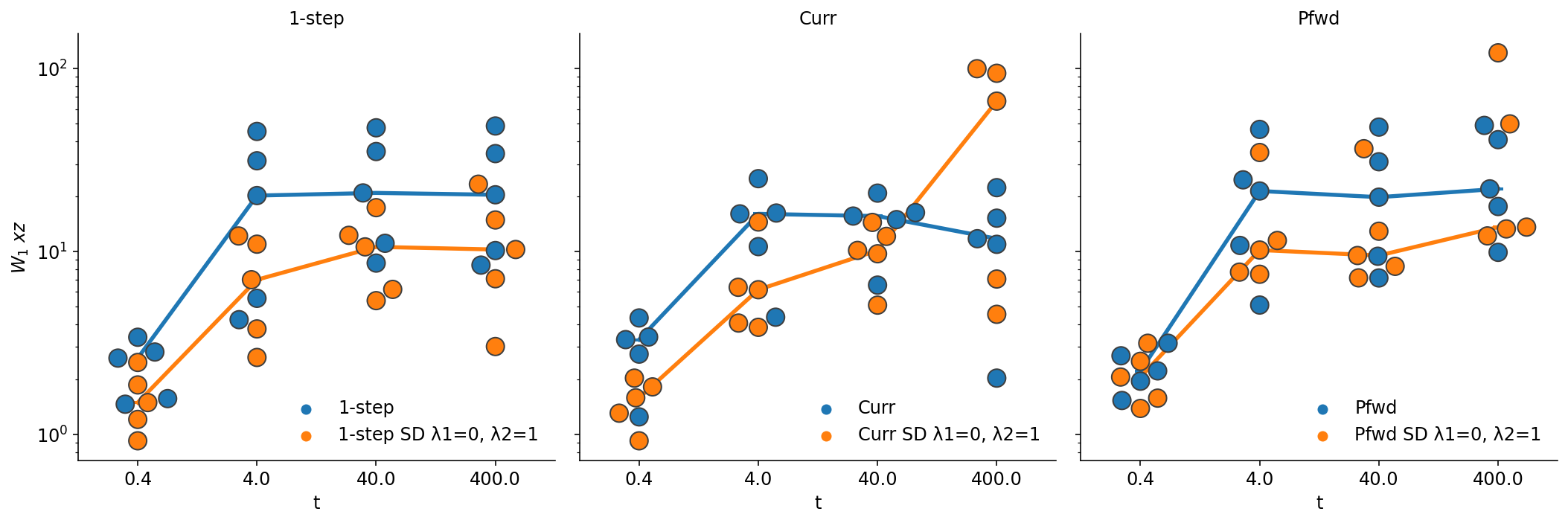}}

    \end{minipage}
    \caption{Wasserstein-1 metric ($\downarrow$) for the $xz$ values between trajectories at various rollout times of the Lorenz 63 system.
    (a) Values when using MMD for measure matching at training. (b) Values when using SD for measure matching at training. 
    Each point represents a random training seed that remains stable, with the solid line indicating median values. 
    }
    \label{fig:lz63_w_x}
\end{figure*}

\begin{figure*}[h]
    \centering
    \begin{minipage}[c]{0.67\textwidth}  
        \vfill
        \subfloat[]{\includegraphics[width=0.95\textwidth, trim={2mm 0mm 5mm 0mm}, clip]{images/l63_W_xy.png}}
         \vfill
        \vspace{-0.3cm}
        \subfloat[]{\includegraphics[width=0.95\textwidth, trim={2mm 0mm 5mm 0mm}, clip]{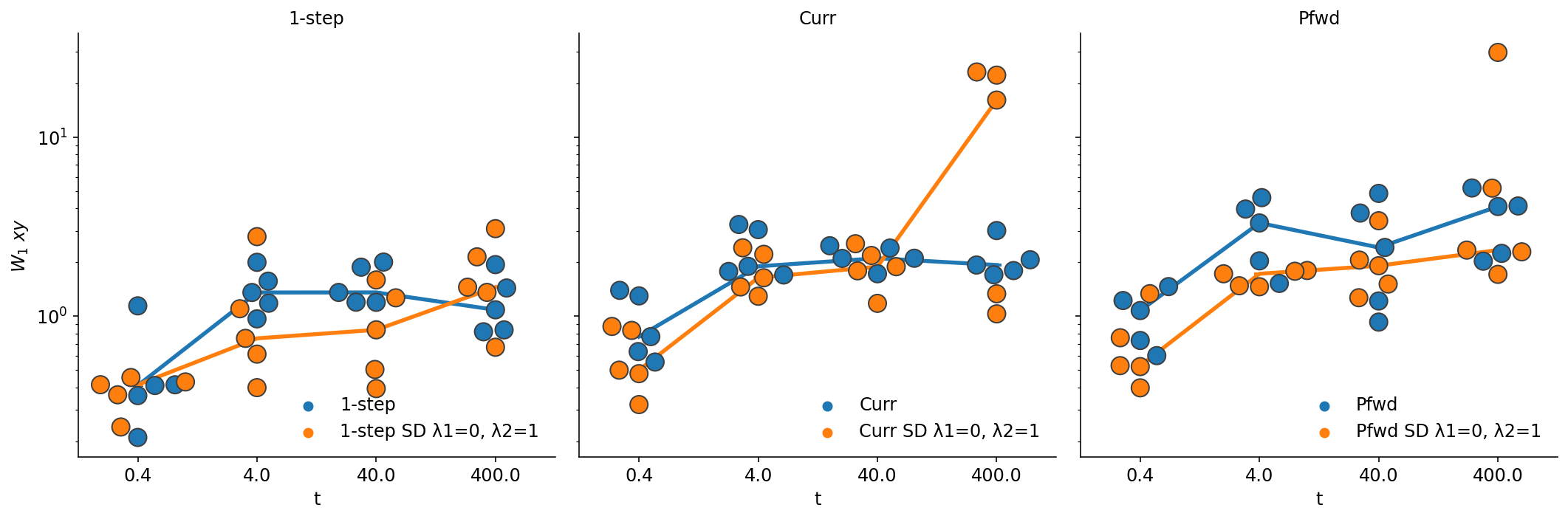}}

    \end{minipage}
    \caption{Wasserstein-1 metric ($\downarrow$) for the $xy$ coordinates between trajectories at various rollout times of the Lorenz 63 system.
    (a) Values when using MMD for measure matching at training. (b) Values when using SD for measure matching at training. 
    Each point represents a random training seed that remains stable, with the solid line indicating median values.  
    }
    \label{fig:lz63_w_z}
\end{figure*}

\paragraph{Lorenz 63 Ablation}
From Figures \ref{fig:lz63_cos_sim}, \ref{fig:lz63_SD}, and \ref{fig:lz63_mmd_ablation}, we can observe that using SD to regularize the objectives provides some benefits.
For the short term behavior in Figure \ref{fig:lz63_cos_sim}, we see that the gains are very similar between using SD or MMD in place of $\D$.
For the measure matching metrics, we see that both measure-matching metrics stabilize the trajectories. 
This can be further be seen in Figures \ref{fig:lz63_w_x} and \ref{fig:lz63_w_z}, which show that SD does stabilize some of the summary metrics. 
We point out that even though the performance is competitive, using MMD still seems to have an edge, albeit fairly small.

\begin{figure*}[h]
    \centering
    \vfill
    \subfloat[]{\includegraphics[width=0.67\textwidth, trim={2mm 0mm 0mm 0mm}, clip]{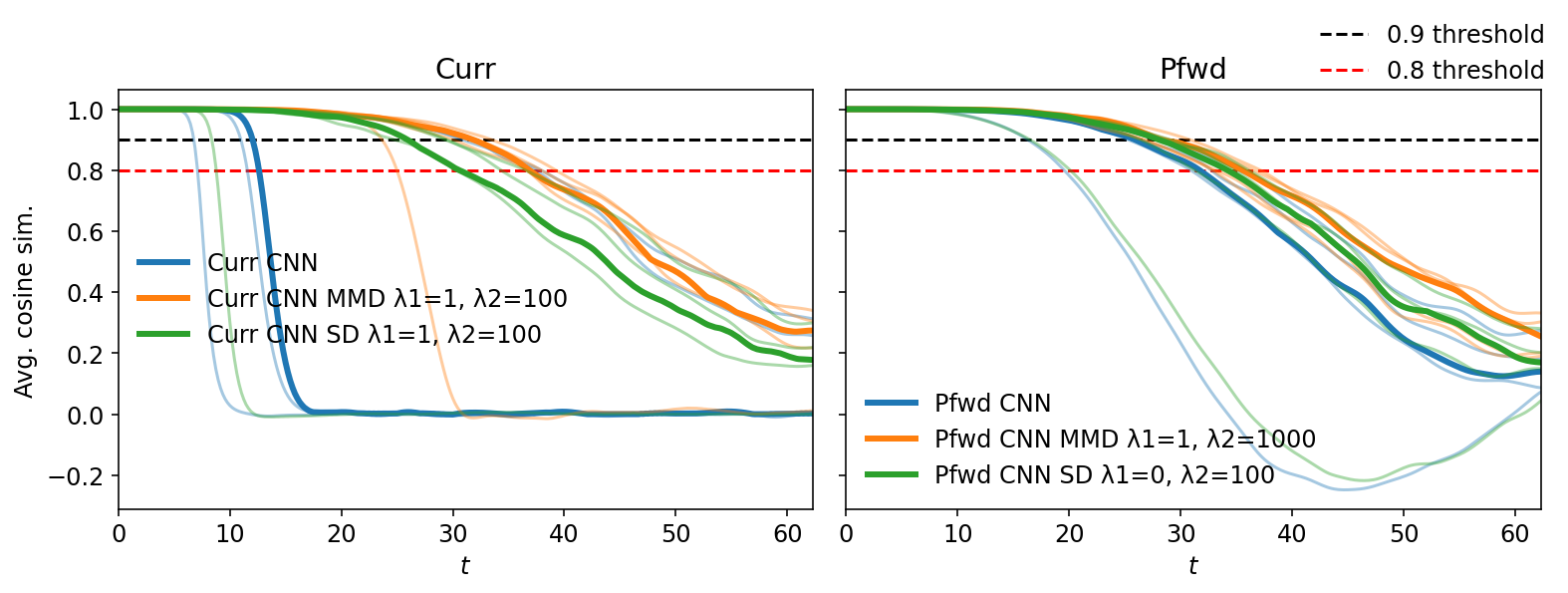}}
     \vfill
    \vspace{-0.3cm}
    \subfloat[]{\includegraphics[width=0.48\textwidth, trim={0mm 0mm 0mm 0mm}, clip]{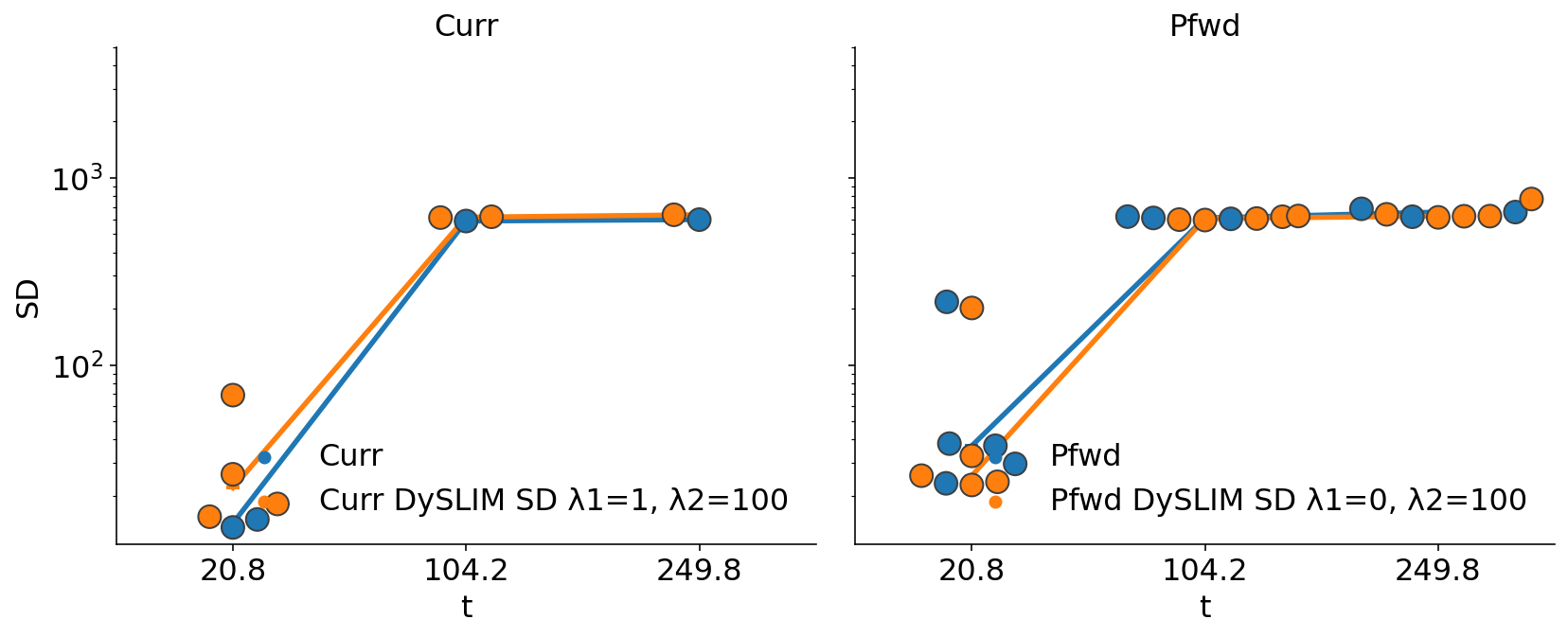}}
    \subfloat[]{\includegraphics[width=0.48\textwidth, trim={0mm 0mm 0mm 0mm}, clip]{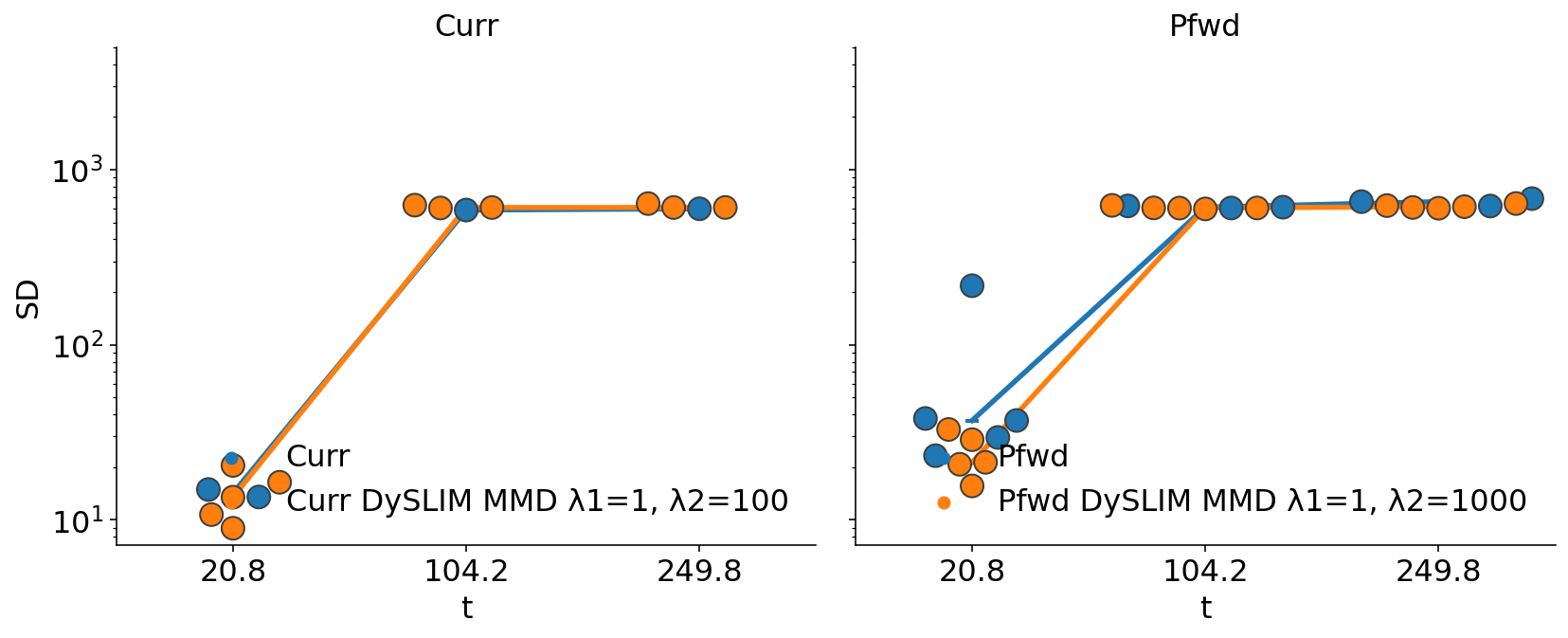}}
    
    \caption{Metrics for KS system.
    (a) Cosine Similarity ($\uparrow$) between the different trajectories.
    Each line corresponds to the mean over trajectories of each of five random training seeds, with bold lines indicating median values.
    (b) Sinkhorn Divergence (SD; $\downarrow$) between trajectories at various rollout times when using MMD for measure matching at training.
    (c) SD between trajectories at various rollout times when using SD for measure matching at training.
    Each point represents a random training seed that remains stable, with the solid line indicating median values. 
    }
    \label{fig:ks_mmd_sd}
\end{figure*}

\paragraph{Kuramoto–Sivashinsky Ablation}
In Figure \ref{fig:ks_mmd_sd}, we find evidence that using SD as the regularizer for the KS system does improve stability, but performance with respect to using MMD starts to deteriorate. 
In Figure \ref{fig:ks_mmd_sd} (a), we see that there is indeed a better performance when using MMD as opposed to SD.
In Figure \ref{fig:ks_mmd_sd} (b) we find that even though the SD regularization helps, many of the trajectories still diverge particularly for curriculum training.

The conclusion from this ablation is that while SD is compatible with our DySLIM framework and does provide increased stability, we find that using MMD in the regularizer leads to superior performance, especially for systems with larger dimension, e.g., KS.

\section{Assets}\label{appsec:assets}
\subsection{Software and Libraries}
In Table
\ref{tab:assets}, we list relevant open-source software, and corresponding licenses, used in this work:

\begin{table}[ht]
    \caption{Open source libraries used in this work, with corresponding licenses.}
    \label{tab:assets}
    \vskip 0.15in
    \begin{center}
    \begin{tabular}{ll}
        \toprule
        Library & License \\
        \midrule        
        Flax \citep{flax2020github} & Apache 2.0 \\
        Jax \citep{jax2018github} & Apache 2.0 \\
        Jax-CFD \citep{dresdner2022learning} & Apache 2.0 \\
        NumPy \citep{harris2020array} & \href{https://numpy.org/doc/stable/license.html}{NumPy license} \\
        Matplotlib \citep{Hunter:2007} & \href{https://matplotlib.org/stable/users/project/license.html}{Matplotib license} \\
        ML Collections & Apache 2.0 \\
        Optax & Apache 2.0 \\
        Orbax & Apache 2.0 \\
        OTT-Jax \citep{cuturi2022optimal} & Apache 2.0 \\
        Pandas \citep{reback2020pandas} & BSD 3-Clause ``New" or ``Revised" \\
        SciPy \citep{2020SciPy-NMeth} & \href{https://projects.scipy.org/scipylib/license.html}{SciPy license} \\
        Seaborn \citep{Waskom2021} & BSD 3-Clause ``New" or ``Revised" \\
        Swirl Dynamics \citep{wan2023evolve} & Apache 2.0 \\
        TensorFlow \citep{tensorflow2015-whitepaper} & Apache 2.0 \\
        Xarray \citep{hoyer2017xarray} & Apache 2.0 \\
        \bottomrule
        \end{tabular}
    \end{center}
\end{table}

\subsection{Computational Resources}\label{appsubsec:compute}
Experiments were submitted as resource-restricted jobs to a shared compute cluster.
Computational resources used in each dynamical system experiment are listed in Table \ref{tab:comp}.

\begin{table}[t]
    \caption{Computational resources by experiment.}
    \label{tab:comp}
    \vskip 0.15in
    \begin{center}
    \begin{tabular}{ll}
        \toprule
        Experiment & Hardware \\
        \midrule
        Lorenz 63 & 1 V100 GPU, 16 GB\\
        Kuramoto–Sivashinsky & 1 V100/A100 GPU, 16/40GB\\
        Kolmogorov Flow & 1 A100 GPU, 40GB\\
        \bottomrule
        \end{tabular}
    \end{center}
\end{table}


\end{document}